\documentclass[aos]{imsart}

\RequirePackage{amsthm,amsmath,amsfonts,amssymb}
\RequirePackage[numbers,sort&compress]{natbib}
\RequirePackage[colorlinks,citecolor=blue,urlcolor=blue]{hyperref}
\RequirePackage{graphicx}
\usepackage{tcolorbox}
\usepackage{mathrsfs}
\usepackage{bm,booktabs,float}
\usepackage{hyperref}
\usepackage{makecell}
\usepackage{multirow}
\usepackage{adjustbox}
\usepackage{algorithm}
\usepackage{algorithmic}
\usepackage{subcaption}

\startlocaldefs
 \theoremstyle{plain}

 \newtheorem{theorem}{Theorem}[section]
 \newtheorem{lemma}[theorem]{Lemma}
 \newtheorem{prop}[theorem]{Proposition}
 \newtheorem{assumption}{Assumption}
 
 \newtheorem{cor}[theorem]{Corollary}
 \newtheorem{defi}[theorem]{Definition}
 \newtheorem{remark}[theorem]{Remark}
 \newenvironment{statement}[1]
 \theoremstyle{remark}
 
 \newtheorem{exam}{Example}

\newcommand{\change}[1]{{{\color{black} #1}}}
\newcommand{\mainchange}[1]{{{\color{black} #1}}}

\DeclareMathOperator*{\argmax}{argmax}
\DeclareMathOperator*{\argmin}{argmin} 

\def\bx{\boldsymbol{x}}
\def\by{\boldsymbol{y}}
\def\bz{\boldsymbol{z}}
\def\bb{\boldsymbol{b}}
\def\bu{\boldsymbol{u}}
\def\bv{\boldsymbol{v}}
\def\bw{\boldsymbol{w}}

\def\bV{\boldsymbol{V}}
\def\bI{\boldsymbol{I}}
\def\bs{\boldsymbol{s}}

\def\bA{\boldsymbol{A}}

\def\bW{\boldsymbol{W}}

\def\bA{\boldsymbol{A}}
\def\cA{\mathcal{A}}

\def\bD{\boldsymbol{D}}
\def\cD{\mathcal{D}}
\def\cE{\mathcal{E}}
\def\cS{\mathcal{S}}
\def\cF{\mathcal{F}}
\def\cG{\mathcal{G}}
\def\cH{\mathcal{H}}

\def\cN{\mathcal{N}}
\def\cP{\mathcal{P}}

\def\cR{\mathcal{R}}
\def\cX{\mathcal{X}}

\def\cZ{\mathcal{Z}}

\def\Rbb{\mathbb{R}}

\def\Ebb{\mathbb{E}}
\def\Pbb{\mathbb{P}}
\def\Qbb{\mathbb{Q}}

\def\sign{\operatorname{sign}}
\def\Lip{\operatorname{Lip}}

\DeclareMathOperator*{\esssup}{ess\,sup}

 \endlocaldefs

\begin{document}
	
\begin{frontmatter}
	\title{Wasserstein Distributionally Robust Nonparametric Regression}
	\runtitle{Distributionally Robust Nonparametric Regression}
	
	\begin{aug}
		\author[A]{\fnms{Changyu}~\snm{Liu}
		\ead[label=e1]{changyu.liu@polyu.edu.hk}},
		\author[B]{\fnms{Yuling}~\snm{Jiao}\ead[label=e2]{yulingjiaomath@whu.edu.cn}},
		\author[C]{\fnms{Junhui}~\snm{Wang}\ead[label=e3]{junhuiwang@cuhk.edu.hk}}
		\and
		\author[D]{\fnms{Jian}~\snm{Huang}\ead[label=e4]{j.huang@polyu.edu.hk}}
		\address[A]{Institute for Math and AI, Wuhan University
			\printead[presep={,\ }]{e1}}
		\address[B]{School of Artificial Intelligence,
			Wuhan University\printead[presep={,\ }]{e2}}
		\address[C]{Department of Statistics and Data Science, The Chinese University of Hong Kong\printead[presep={,\ }]{e3}}
		\address[D]{Departments of Data Science and AI, and Applied Mathematics,
			The Hong Kong Polytechnic University\printead[presep={ ,\ }]{e4}}
\end{aug}

\begin{abstract}
\change{Wasserstein distributionally robust optimization (WDRO) strengthens statistical learning under model uncertainty by minimizing the local worst-case risk within a prescribed ambiguity set. Although WDRO has been extensively studied in parametric settings, its theoretical properties in nonparametric frameworks remain underexplored. This paper investigates WDRO for nonparametric regression. We first establish a   structural distinction based on the order $k$ of the Wasserstein distance, showing that $k=1$ induces Lipschitz-type regularization, whereas $k>1$ corresponds to gradient-norm regularization. To address model misspecification, we analyze the excess local worst-case risk, deriving non-asymptotic error bounds for estimators constructed using norm-constrained feedforward neural networks. This analysis is supported by new covering number and approximation bounds that simultaneously control both the function and its gradient.  The proposed estimator achieves a convergence rate of $n^{-2\beta/(d+2\beta)}$ up to logarithmic factors, where $\beta$ depends on the target’s smoothness and network parameters. This rate is shown to be minimax optimal under conditions commonly satisfied in high-dimensional settings. Moreover, these bounds on the excess local worst-case risk imply guarantees on the excess natural risk, ensuring robustness against any distribution within the ambiguity set. We show the framework’s generality across regression and classification problems. Simulation studies and an application to the MNIST dataset further illustrate the estimator's robustness.  }
\end{abstract}

\begin{keyword}[class=MSC]
	\kwd[Primary ]{62G05}
	\kwd{62G08}
	\kwd[; secondary ]{68T07}
\end{keyword}

\begin{keyword}
	\kwd{distributionally robust optimization}
\kwd{error analysis}
   \kwd{misspecified model}
\kwd{neural network approximation}
    \kwd{nonparametric regression}
\end{keyword}

\end{frontmatter}

\section{Introduction}
Consider a nonparametric regression model
\begin{equation}
\label{model_re}
Y=f_0(X)+\varepsilon,
\end{equation}
where $Y\in \mathbb{R}$ is a response, $X\in \mathbb{R}^d$ is a $d$-dimensional vector of covariates,
$f_0: [0, 1]^d \to \mathbb{R}$ is an unknown regression function, and  $\varepsilon$ is an unobservable error. Assumptions on $\varepsilon$ will be specified for each estimation method below.

Our objective is to estimate the unknown regression function $f_0$ from a finite sample. While extensive research has been conducted on methods for estimating $f_0$ in standard settings,  model uncertainty remains a significant concern. Such uncertainty can arise  from various factors, such as incorrect modeling assumptions, data contamination, or distributional inconsistencies between the training and target data. If model uncertainty is not properly addressed, estimators built on incorrect models may exhibit poor generalization performance. To address this problem, we consider a distributionally robust nonparametric estimator, derived by solving the following optimization problem:
\begin{equation} \label{eq_DROeq}
\inf_{f\in\cF_n} \Big\{\cR_{\Pbb,k}(f;\delta):= \sup_{\Qbb: W_k(\Qbb,\Pbb)\leq \delta}\Ebb_{ \Qbb}\left[\ell(Y-f(X))\right]\Big\},
\end{equation}
where $\ell(\cdot)$ is a loss function, $W_k(\cdot, \cdot)$ denotes the Wasserstein distance of order $k$, and $\cF_n$ represents an estimation function space that may vary with $n$. We refer to  $\cR_{\Pbb,k}(f; \delta)$ as the local worst-case risk,  where the inner supremum identifies  the worst-case distribution within a $\delta$-Wasserstein ball centered at the joint distribution $\Pbb$  of $(X,Y)$. This framework addresses model uncertainty by incorporating distributional perturbations within an ambiguity set. By evaluating and minimizing the risk associated with the worst-case distribution, the resulting estimator enhances robustness and reliability in the presence of potential model uncertainties.

The optimization problem \eqref{eq_DROeq} falls within the broader framework of  Wasserstein distributionally robust optimization (WDRO), a powerful approach that has recently emerged for addressing learning and prediction challenges under model uncertainty.  WDRO has been successfully applied across diverse domains, including mechanism design \citep{koccyiugit2020distributionally}, power systems \citep{ROALD2023108725}, and machine learning \citep{moos2022robust}.  Despite its broad adoption, the theoretical understanding of WDRO remains limited. Existing studies primarily focus on parametric settings, where WDRO is linked to classical regularized learning methods  \citep{BlanchetGrop2017, Blanchet2019,Shafieezadehmass2019,Blanchetbio2022}.

In nonparametric settings, while WDRO is known to exhibit certain regularization effects for smooth loss functions \citep{bartl2021sensitivity, gao2022wasserstein}, its generalization properties have yet to be systematically analyzed. Another critical challenge in WDRO analysis is model misspecification. This occurs when the estimation function space $\mathcal{F}_n$ does not contain the target function $f_0$ and
fails to approximate it well. This source of uncertainty can result in poor generalization performance, yet it has received limited attention in the WDRO literature.

A related line of research is adversarial risk, which focuses exclusively on uncertainty in the covariates. This targeted scope makes adversarial risk a special case of WDRO, primarily used in adversarial training to defend against adversarial examples \citep{Szegedy2014Intriguing, GoodfellowSS14, madry2018towards}. Its generalization properties have also been extensively studied \citep{Yin19b, khim2018adversarial, Awasthi20a, mustafa22a, Tu2019, LLLiu2024}.  However, the analysis developed for adversarial risk does not directly transfer to the WDRO framework. Adversarial risk considers only local perturbations, whereas WDRO accounts for distributional perturbations that induce distinct regularization effects and require different analytical strategies.

In this paper, we investigate the generalization properties of the distributionally robust nonparametric estimators. To evaluate the estimator's performance under model misspecification, we define the excess local worst-case risk as
  \begin{equation*}
  	\cE_{\Pbb,k}(f;\delta)= \cR_{\Pbb,k}(f;\delta)-\inf_{f\in\cH^{\alpha}\cup \cF_n}\cR_{\Pbb,k}(f;\delta),
  \end{equation*}
  where $\cH^{\alpha}$ denotes the H\"{o}lder  class with smoothness level $\alpha$.
We show that for any distribution $\Pbb^{\prime}$ satisfying $W_k(\Pbb^{\prime},\Pbb)\leq \delta$, the following holds.
  \begin{align*}
{\cal E}_{\Pbb^{\prime}}(f)=\cR_{\Pbb^{\prime}}(f)- \inf_{f\in\cH^{\alpha}}  \cR_{\Pbb^{\prime}}(f)\lesssim 	\cE_{\Pbb,k}(f;\delta)+   \delta\|\ell \|_{\Lip},
  \end{align*}
  where $\cR_{\Pbb^{\prime}}(f)=\Ebb_{\Pbb^{\prime}} [\ell (Y-f(X)) ]$  is the natural risk under distribution $\Pbb^{\prime}$ and $\|\ell\|_{\Lip}$ denotes the Lipschitz constant of the loss function.
  Thus, the excess local worst-case risk provides an upper bound on the excess natural risk for any distribution within the ambiguity set, illustrating that the estimator achieves reasonable performance even under model uncertainty.

Our main contributions are summarized as follows.
	\begin{itemize}
		
    	\item[(1)] We analyze the theoretical properties of the local worst-case risk $\cR_{\Pbb,k}(f; \delta)$ and show a structural distinction in its regularization behavior. Ambiguity sets based on the $W_1$ distance induce Lipschitz-type regularization, primarily influenced by points where the Lipschitz constant is attained. In contrast, for $k>1$, ambiguity sets based on $W_k$ correspond to a form of gradient-norm regularization.
		
		\item[(2)]
We develop key analytical tools for estimators built from norm-constrained feedforward neural networks. In particular, we establish covering-number and approximation bounds that jointly control both the function and its gradient. These results may be of independent interest for problems involving norm-constrained neural networks.
		
		\item[(3)] We establish non-asymptotic error bounds for the proposed estimator, achieving a convergence rate of $n^{-2\beta/(d+2\beta)}$ up to logarithmic factors. This rate applies to both the excess local worst-case risk and the excess natural risk under distributional shifts within the ambiguity set. Under conditions commonly satisfied in high-dimensional settings, the rate coincides with the minimax optimal rate for nonparametric regression. The established error bounds also reveal distinct robustness-related structures, showing that $W_1$ ($k=1$) induces a term linear in $\delta$, while $W_k$ ($k>1$) introduces more complex higher-order effects.
		
		\item[(4)] We illustrate the applicability of our theoretical results in both regression and classification problems. The framework accommodates unbounded losses in nonparametric least squares regression, provides explicit robustness guarantees for Huber and quantile regression, and extends naturally to binary classification with label noise.		
	\end{itemize}

Our results clarify how Wasserstein order shapes regularization and robustness: $W_1$ enforces Lipschitz control, while higher-order $W_k$ induces gradient-norm regularization. We provide general tools for norm-constrained neural networks via bounds that jointly control functions and gradients. The non-asymptotic guarantees attain near-minimax rates and  make explicit the
 robustness trade-offs (linear in $\delta$ for $W_1$ vs. higher-order effects for $W_k$), strengthening performance under shift. Our results' applicability to regression and classification, including least squares regression, Huber and quantile regression, and binary classification, demonstrates its practical relevance.
We also conduct simulation studies and use an application to the MNIST dataset to illustrate the empirical behavior of WDRO regression and classification.

\subsection{Related works}\label{sec_related}
In studying the statistical properties of WDRO, extensive work has focused on characterizing the generalization properties of the WDRO estimator.  This includes establishing non-asymptotic upper bounds for
\[
 \cR_{\Pbb,k}(f;\delta)-\cR_{\Pbb_n,k}(f;\delta) ,
\]
where $\Pbb$ denotes the underlying data distribution and $\Pbb_n$ denotes the empirical distribution of a random sample with size $n$ from $\Pbb,$ see for example, \cite{kwon2020principled} and \cite{lee2018minimax}.

Another important line of research  has focused on finite-sample statistical guarantees, typically formulated as
\begin{align}\label{eq_certificate}
 \cR_{\Pbb}(f)\leq \cR_{\Pbb_n,k}(f;\delta)
\quad \text{for any } f\in\cF_n ,
\end{align}
with high probability, ensuring that the empirical WDRO objective function provides an upper bound for expected loss under the true distribution $\Pbb$.  While increasing $\delta$ raises the likelihood of satisfying \eqref{eq_certificate}, excessively large values of $\delta$ may yield overly conservative estimates. Thus, determining  the appropriate scale of $\delta$ in relation to the sample size is critical.  \cite{KuhnMohajerin2018} established \eqref{eq_certificate} with $\delta=O(n^{-1/d})$ based on measure concentration results in \cite{Fournier2015}, with later works improving this rate to  $O(n^{-1/2})$ in  parametric settings  \citep{Shafieezadehmass2019}, and relating $\delta$ to the  function class complexity  in more general settings \citep{azizian2023exact, le2024universal, an2021generalization, gao2023finite}. However, these studies do not account for misspecification within the estimation function class $\cF_n$, which could prevent the right-hand side of \eqref{eq_certificate} from becoming small when there is a substantial discrepancy between $\cF_n$ and the target function $f_0$. This form of model uncertainty suggests that verifying \eqref{eq_certificate} alone may be insufficient to ensure reliable performance.

Our work addresses this issue by considering two different measures to characterize the generalization properties of the WDRO estimator. We first focus on the excess local worst-case risk
 \begin{equation*}
 	\cE_{\Pbb,k}(f;\delta)= \cR_{\Pbb,k}(f;\delta)-\inf_{f\in\cH^{\alpha}\cup \cF_n}\cR_{\Pbb,k}(f;\delta),
 \end{equation*}
which extends the framework in \cite{lee2018minimax} by incorporating model misspecification. Furthermore, we establish that for any distribution $\Pbb^{\prime}$ satisfying $W_k(\Pbb^{\prime},\Pbb)\leq \delta$, the following holds.
  \begin{align*}
 	{\cal E}_{\Pbb^{\prime}}(f)\lesssim 	\cE_{\Pbb,k}(f;\delta)+  \delta\|\ell \|_{\Lip},
 \end{align*}
which links the natural risk with the local worst-case risk while accounting for model misspecification on both sides of the inequality. This result illustrates  WDRO’s ability  to address model uncertainty by producing an estimator that performs reliably for any distribution within the ambiguity set.

The optimization problem in \eqref{eq_DROeq} has been shown to recover various regularized learning methods when applied to specific loss functions and parametric regression models. Notable examples include the square-root Lasso \citep{Blanchet2019}, support vector machines \citep{Shafieezadehmass2019},   group-square-root Lasso \citep{BlanchetGrop2017}, and other parametric forms \citep{Blanchetbio2022}. In more general nonparametric settings, recent research indicates that $\cR_{\Pbb,k}(f; \delta)$ exhibits regularization effects similar to those in variation regularization methods \citep{gao2022wasserstein}. Additionally, \cite{bartl2021sensitivity} analyzed the asymptotic first-order behavior of the optimal value of $\cR_{\Pbb,k}(f; \delta)$ with respect to $\delta$ and explored its influence on optimizer performance.
\vspace{0.1in}

{\bf Notation.}
Let the set of positive integers be denoted by $\mathbb{N}= \{1, 2, . . .\}$ and let  $\mathbb{N}_0=\mathbb{N}\cup\{0\}.$  If $a$ and $b$ are two quantities, we use $a\lesssim b $ or $b\gtrsim a$ to denote the statement that $a\leq Cb$ for some constant $C>0$. We denote $a\asymp b$ when $a\lesssim b\lesssim a$. Let $\lceil a \rceil$ denote the smallest integer larger than or equal to quantity $a$. For a vector $\bx$ and $p\in[1,\infty]$, we use $\|\bx \|_{p}$ to denote the $p$-norm of $\bx$. For a function $f$, we denote its supremum norm by $\|f \|_{\infty}$ and its gradient by $\nabla f$. For a measurable function $H$, let $\esssup(H)$ denote  its essential supremum. For any interger $q\in \mathbb{N}$, let $q^{\star}$ be the integer such that $1/q+1/q^{\star}=1$.

 The rest of the paper is organized as follows.  Section~\ref{sec_related} provides a brief review of related works.  Section~\ref{sec_setup} introduces the local worst-case risk and shows its theoretical properties.  Section~\ref{sec_estimator} presents the setup and establishes key properties for norm-constrained FNNs. The main results are presented in Section \ref{sec_main}.  Section~\ref{sec_app} applies these results to nonparametric regression and classification problems. Section~\ref{sec_num} presents the  numerical experiments.
Concluding remarks are given in Section~\ref{sec_dis}.  Proofs of the results and additional technical details are provided in the Supplementary Material.

\section{Wasserstein distributionally robust nonparametric regression}\label{sec_setup}
Let $\Pbb$ denote the joint distribution of $Z=(X, Y)$ under model \eqref{model_re} and let the support of $X$ be $\mathcal{X}=[0, 1]^d.$
A classical approach to nonparametric regression seeks to find an estimator $\hat{f}_n$ within an appropriate nonparametric function space by minimizing the expected loss,  also referred to as the natural risk:
\begin{equation}\label{eq_naturalrisk}
\cR_{\Pbb}(f)=\Ebb_{\Pbb}\big[\ell (Y-f(X))\big],
\end{equation}
where $\ell:\Rbb \to [0,\infty)$ is a loss function. Various nonparametric function spaces, such as reproducing kernel Hilbert spaces, B-splines, and deep neural networks, have been employed to estimate  $f_0$, with the choice of  estimation function class typically depending on the form of the loss function and any assumptions about the smoothness or structure of  $f_0$. Despite these advancements, model uncertainty remains a persistent challenge in nonparametric regression. To illustrate this issue, we present two simple examples.

\begin{exam}[Contaminated distribution]\label{ex1}
Suppose $Y=f_0(X)+\varepsilon$, where $\Ebb[\varepsilon|X]=0$ and $\Ebb[\varepsilon^2]<\infty$. A common approach to estimate $f_0$ is through least squares regression, which minimizes the expected squared error:
\[
\argmin_{f}\Ebb_{\Pbb}\big[\{Y-f(X)\}^2\big]=\argmin_{f}\big\{\Ebb \big[\{f_0(X)-f(X)\}^2 \big]+\Ebb[\varepsilon^2] \big\}=f_0.
\]
Now, consider a scenario in which the observed data are contaminated. In particular, the covariates are observed without error, $\tilde{X} = X$, while the observed response is $\tilde{Y}=Y+g(X)$, where $\Ebb[|g(X)|]\leq \delta$. In this case, the solution to the least squares problem becomes $\Ebb[\tilde{Y}|\tilde{X}]=f_0(X)+g(X),$ so directly applying least squares to contaminated data produces biased estimator, which may significantly degrade generalization performance.

This example illustrates a form of Wasserstein contamination \citep{Zhu2022,liu_robust_2022}, which assumes that the observations come from a corrupted distribution within a Wasserstein ball centered at the true distribution. This framework captures more general types of corruption than the classical Huber's $\delta$-contamination model \citep{Huber1964}, which allows only an $\delta$-fraction of the data to be arbitrarily corrupted.
\end{exam}

\begin{exam}[Distributional inconsistency]\label{ex3}
Suppose we observe $(X_i, Y_i)$, $i = 1, \dots, n$, independently sampled from the model $Y = f_0(X)+ \varepsilon$.
After deriving an estimator from these observations, our goal is to predict the response for new covariates. In practice, however, the target data may follow a slightly different model,  $Y^{\prime} = f_0^{\prime}(X^{\prime}) + \varepsilon^{\prime}$, where the distribution of $(X^{\prime}, Y^{\prime})$ differs from that of the observed data.

 In such cases, the estimator derived from the original observations may perform poorly, as it does not account for potential distributional shifts.  This scenario is commonly encountered in open-world settings, such as natural scene image analysis, where the distributions of training and target data are unknown and may exhibit subtle inconsistencies.
\end{exam}

These examples illustrate that estimators that ignore this uncertainty may perform poorly.
Explicitly incorporating uncertainty into the estimation process is therefore important.  The WDRO framework offers a systematic solution by defining an ambiguity set around the observed data and producing estimators that remain robust under uncertainty.

Specifically, we define the local worst-case risk to construct a distributionally robust nonparametric estimator for  $f_0$:
\begin{equation}
\label{Rp}
 \cR_{\Pbb, k}(f;\delta)=\sup_{\Qbb: W_k(\Qbb,\Pbb)\leq \delta}\Ebb_{\Qbb}\left[\ell(Y-f(X))\right],
\end{equation}
where $\delta$ is a nonnegative constant that determines the level of uncertainty, and $W_k(\cdot, \cdot)$ denotes the Wasserstein distance of order $k$, formally defined below.

\begin{defi}[Wasserstein distance]
Let the space $\cZ$ be equipped with a norm $\|\cdot\|$. For any $k\in\mathbb{N}$, let $\cP_k(\cZ)$ denote the space of Borel probability measures on $\cZ$ with a finite $k$-th moment. The Wasserstein distance of order $k$ between two probability measures $\Pbb,\Qbb\in\cP_k(\cZ)$ is defined as
\begin{align*}
W_k(\Pbb,\Qbb)&=\inf_{\pi\in\Pi(\Pbb,\Qbb)}\Ebb_{(Z,\widetilde{Z})\sim \pi}\big[\|Z-\widetilde{Z}\|^k\big]^{1/k},
\end{align*}
where the infimum is taken over the set  $\Pi(\Pbb,\Qbb)$ consisting of all probability measures on $\cZ\times\cZ$ with marginals $\Pbb$ and $\Qbb$, respectively.
\end{defi}

By the definition, $W_k(\Pbb,\Qbb) \leq  W_q(\Pbb,\Qbb) $ for any $1\leq k\leq q\leq \infty.$ This implies that as  $k$ increases,  $\cR_{\Pbb,k}(f;\delta)$ reflects progressively lower levels of uncertainty. Moreover, when $k=\infty$ and the norm $\|\cdot\|$ is specifically chosen, $\cR_{\Pbb,k}(f;\delta)$ reduces to adversarial risk, which is commonly used in the training process to defend against adversarial examples \citep{Szegedy2014Intriguing,GoodfellowSS14,madry2018towards}.

The level of uncertainty within $\cR_{\Pbb,k}(f;\delta)$ is determined by $\delta.$ When $\delta=0$, no uncertainty is considered, resulting in $\cR_{\Pbb,k}(f;0)=\cR_{\Pbb}(f)$, where $\cR_{\Pbb}(f)$ represents the natural risk \eqref{eq_naturalrisk}.  However, when $\delta>0$, analyzing $\cR_{\Pbb,k}(f;\delta)$ becomes challenging due to optimization over an infinite-dimensional space of probability measures. Fortunately, a tractable reformulation of $\cR_{\Pbb,k}(f;\delta)$ is enabled by a strong duality result  \citep{sinha2018,zhang2022simple}, as shown below. For notational simplicity, let $\bz=(\bx,y)$ and $\ell(\bz;f)=\ell(y-f(\bx))$.

\begin{lemma}\label{lem_dual}
	Suppose $\ell(\bz;f)$ is a continuous function defined over the normed space $(\cZ, \|\cdot\|)$.  Define $\varphi_{\gamma,k}(\bz;f)=\sup_{\bz^{\prime}\in\cZ} \{\ell(\bz^{\prime};f)-\gamma\|\bz^{\prime}-\bz\|^{k}\}.$ For any $\delta>0$,
	\begin{align}\label{eq_dual}
	\sup_{\Qbb: W_k(\Qbb,\Pbb)\leq \delta }\Ebb_{ \Qbb}\left[\ell(Z;f)\right]=\inf _{\gamma \geq 0}\big\{\gamma \delta^{k} +\mathbb{E}_{\Pbb}\left[\varphi_{\gamma,k}(Z;f)\right]\big\},
	\end{align}
	and for any $\gamma\geq 0$, there holds
\begin{align*}
		\sup_{\Qbb}  \big\{\Ebb_{\Qbb}\left[\ell(Z;f) \right] -\gamma
 W_k(\Qbb,\Pbb)^{k}  \big\} = \Ebb_{\Pbb}[\varphi_{\gamma,k}(Z;f)].
		\end{align*}
\end{lemma}
As demonstrated in \eqref{eq_dual}, $\cR_{\Pbb,k}(f;\delta)$ can be reformulated as a univariate minimization problem, enabling tractable convex programs for many common loss functions and estimation function spaces, thus facilitating numerical computation \citep{sinha2018,Shafieezadehmass2019}. Furthermore, when the regression function is assumed to follow specific parametric structures, $\cR_{\Pbb,k}(f; \delta)$ is directly connected to various classical regularized learning methods. For instance, consider the case of linear regression where $f(\bx)=\bx^{\top}\bm{\beta}$, $\ell(\bz;\bm{\beta})=(y-\bx^{\top}\bm{\beta})^2$, and $\|\bz-\bz^{\prime}\|=\|\bx-\bx^{\prime}\|_{q} +\infty 1{\{y\ne y^{\prime}\}}$.  Then,  minimizing $\cR_{\Pbb,2}(f;\delta)$ is equivalent to solving
\[
\min_{\bm{\beta}}\big\{(\Ebb_{\Pbb}[\ell(Z;\bm{\beta})])^{1/2}+\delta\|\bm{\beta}\|_{q^{\star}}\big\}^2,
\]
where $1/q^{\star}+1/q=1$ \citep{Blanchet2019}. This shows that the robustness achieved through the local worst-case risk is enabled by norm regularization of the parameter. Similar norm regularization effects of $\cR_{\Pbb,k}(f;\delta)$ on the model parameters can also be observed in other parametric settings \citep{Blanchetbio2022}.

However, the exact regularization form of $\cR_{\Pbb,k}(f;\delta)$ does not generally hold, especially in nonparametric settings,
making the analysis of distributionally robust nonparametric estimators more challenging. The following result illustrates a connection between $\cR_{\Pbb,1}(f;\delta)$ and regularization on the gradient of the loss function.
\begin{lemma}\label{lem_regu}
	Suppose the composed loss $\ell(\cdot;f)$ has a bounded Lipschitz constant $\|\ell_{f}\|_{\Lip}<\infty$, then
	\begin{align*}
	\cR_{\Pbb,1}(f;\delta)-\cR_{\Pbb}(f) 	\leq \delta \|\ell_{f}\|_{\Lip}.
	\end{align*}
 Furthermore, there holds
 	\begin{align*}
 \cR_{\Pbb,1}(f;\delta)-\cR_{\Pbb}(f)=\delta  \|\ell_{f}\|_{\Lip},
 \end{align*}
if one of the following conditions is satisfied:
	\begin{itemize}
	\item[(i)] There exists $\bz_0\in\cZ$ such that
	\[
	{\lim\sup}_{\|\bz-\bz_0\|\rightarrow\infty}\frac{\ell(\bz;f)-\ell(\bz_0;f)}{\|\bz-\bz_0\|}= \|\ell_{f}\|_{\Lip}.
	\]
	\item[(ii)] There exists $\bz_0\in\cZ$, a positive constant $\tau$, and a sequence $\{r_m\}_{m\geq 1}\rightarrow 0$ such that
		\[
	{\sup}_{\|\bz-\bz_0\|\geq \tau}\frac{\ell(\bz;f)-\ell(\bz_0;f)}{\|\bz-\bz_0\|}= \|\ell_{f}\|_{\Lip}
	\]
	and $P(\|Z-\bz_0\|\leq r_m)>0$ for any $m\in\mathbb{N}$.
	\end{itemize}
\end{lemma}
	This lemma shows that the maximum deviation between $\cR_{\Pbb,1}(f;\delta)$ and $\cR_{\Pbb}(f)$ is determined by  $\delta \|\ell_{f}\|_{\Lip}$. This deviation can be realized in both unbounded (condition (i)) and bounded (condition (ii)) scenarios. The unbounded case is covered in \cite[Theorem 2]{gao2022wasserstein}. However, when $\cZ$ is bounded, condition (i) no longer holds. Condition (ii) indicates that the maximum deviation can still be achieved if the Lipschitz constant is attained outside a neighborhood of $\bz_0$ and the probability of $Z$ taking values within any small neighborhood of  $\bz_0$ remains positive.

	When there is additional control over the growth rate of $\nabla \ell(\bz;f)$, a lower bound for $ \cR_{\Pbb,1}(f;\delta)-\cR_{\Pbb}(f)$ can be further derived. For the remainder of this paper, we focus on the Wasserstein distance induced by the $\|\cdot\|_{\infty}$ norm, and the Lipschitz constant defined with respect to $\|\cdot\|_{\infty}$.
	\begin{lemma}\label{lem_regu1}
		Suppose there exists a positive number $q>0$ and a measurable function $H(\bz;f)$ such that for any $\tilde{\bz},\bz\in\cZ,$ there holds
		\[
		\|\nabla \ell(\tilde{\bz};f)-\nabla \ell(\bz;f)\|_1\leq H(\bz;f)\|\tilde{\bz}-\bz\|_{\infty}^q.
		\]
		Let $\mathbb{A}=\{\bz:\|\nabla\ell(\bz;f)\|_1=\esssup(\|\nabla\ell(Z;f)\|_1)\}$. Suppose $\Ebb[H(Z;f) 1\{Z\in\mathbb{A}\}]<\infty$ and $P(Z\in \mathbb{A})>0$, then
		\begin{align*}
		\delta\esssup(\|\nabla\ell(Z;f)\|_1)-  \delta^{1+q}  \frac{\Ebb\big[H(Z;f) 1\{Z\in\mathbb{A}\}\big] }{P(Z\in \mathbb{A})^{1+q}}\leq \cR_{\Pbb,1}(f;\delta)-\cR_{\Pbb}(f) \leq \delta \|\ell_f\|_{\Lip}.
		\end{align*}
	\end{lemma}
	Lemmas \ref{lem_regu} and \ref{lem_regu1} show that the performance of $\cR_{\Pbb,1}(f;\delta)$ is largely determined by the points where the Lipschitz constant is attained. This differs from adversarial risk, which is mainly influenced by the local behavior around each input. The key distinction is that $ \cR_{\Pbb,1}(f;\delta)$ accounts for distributional perturbations that can assign adjustable weights to specific points, whereas adversarial risk focuses on local point perturbations.
	
	For the case where $k \in (1, \infty)$, the performance of $\cR_{\Pbb,k}(f;\delta)$ depends more on the overall behavior of the loss function.
	\begin{lemma}\label{lem_regu2}
		Suppose there exists a measurable function $H(\bz;f)$ such that  for any $\tilde{\bz},\bz\in\cZ,$ there holds
		\[
		\big\|\nabla \ell(\tilde{\bz};f)-\nabla \ell(\bz;f)\big\|_1\leq H(\bz;f)\big\|\tilde{\bz}-\bz\big\|_{\infty}^{\min\{1,k-1\}}.
		\]
		When $k\in(2,\infty)$ and  $H(\bz;f)\in L^{\frac{k}{k-2}}(\Pbb)$, we have
		\begin{align*}
			\big|\cR_{\Pbb,k}(f;\delta)-\cR_{\Pbb}(f) - \delta\Ebb\big[\|\nabla\ell(Z;f)\|_1^{k^{\star}}\big]^{ 1/k^{\star} }\big|\leq \delta^2 \Ebb\big[H(Z;f)^{\frac{k}{k-2}}\big]^{1-\frac{2}{k}}.
		\end{align*}
		When $k\in(1,2]$ and  $H(\bz;f)\in L^{\infty}(\Pbb)$,  we have
		\begin{align*}
			\big|	\cR_{\Pbb,k}(f;\delta)-\cR_{\Pbb}(f) - \delta\Ebb\big[\|\nabla\ell(Z;f)\|_1^{k^{\star}}\big]^{ 1/k^{\star}  }\big|\leq \delta^{k}\esssup\big(H(Z;f)\big).
		\end{align*}	
	\end{lemma}
	Lemma \ref{lem_regu2} shows that for small  $\delta$ and under controlled growth conditions for  $\nabla \ell(\bz;f)$, the deviation between $\cR_{\Pbb,k}(f;\delta)$ and $\cR_{\Pbb}(f)$  approaches $\delta\Ebb [\|\nabla\ell(Z;f)\|_1^{k^{\star}} ]^{1/k^{\star}}$. This result indicates that the extra uncertainty introduced by $\cR_{\Pbb,k}(f;\delta)$ can be intuitively interpreted as a form of gradient-norm regularization. Although a similar conclusion appears in Lemma 1 of \cite{gao2022wasserstein}, Lemma \ref{lem_regu2} employs a distinct proof approach based on the sensitivity analysis framework of \cite{bartl2021sensitivity}, which is originally developed to characterize the first-order asymptotic behavior of the WDRO optimal value.

%
\section{Distributionally robust nonparametric estimator}\label{sec_estimator}

 Our target function is defined as
 \begin{equation}\label{eq_target}
f_{k,\delta}^{\star}=\argmin_{f\in\cH^{\alpha}}\cR_{\Pbb,k}(f;\delta),
 \end{equation}
 where the local worst-case risk is minimized over the H\"{o}lder  class with   smoothness level $\alpha$.
  \begin{defi}[H\"{o}lder  class]
 	Let $d\in\mathbb{N}$ and $\alpha=r+\beta>0$, where $r\in\mathbb{N}_0$ and $\beta\in(0,1]$. The H\"{o}lder  class $\cH^{\alpha} $ is defined as
 	\begin{equation*}
 	\begin{split}
 	\mathcal{H}^{\alpha}   =\left\{f: \Rbb^{d} \rightarrow \Rbb, \max_{\|\bs\|_{1} \leq r} \sup _{\bx \in \mathbb{R}^{d}}\left|\partial^{\bs} f(\bx)\right| \leq 1, \max _{\|\bs\|_{1}=r} \sup _{\bx_1 \neq \bx_2} \frac{\left|\partial^{\bs} f(\bx_1)-\partial^{\bs} f(\bx_2)\right|}{\|\bx_1-\bx_2\|_{\infty}^{\beta}} \leq 1\right\},
 	\end{split}
 	\end{equation*}
 	where $\bs \in \mathbb{N}_{0}^{d}$. Let $\mathcal{H}^{\alpha}(\mathbb{I})= \{f:\mathbb{I} \rightarrow \mathbb{R}, f \in \mathcal{H}^{\alpha} \}$ denote the restriction of $\mathcal{H}^{\alpha}$ to a subset  $\mathbb{I}\subseteq \Rbb^d$.
 \end{defi}

Suppose $(X_1,Y_1), \dots, (X_n,Y_n)$ are independent and identically distributed (i.i.d.) samples from the model \eqref{model_re}.  Let  $\Pbb_n$ denote the empirical distribution induced by the samples. The estimator $\widehat{f}_{k,\delta}^{(n)}$ is defined as the solution to the minimization problem of the empirical counterpart of the local worst-case risk:
	\begin{equation*}
	\widehat{f}_{k,\delta}^{(n)}\in\argmin_{f\in \cF_n} \cR_{\Pbb_n,k}(f;\delta),
	\end{equation*}
where $\cR_{\Pbb_n,k}(f;\delta)=\sup_{\Qbb: W_k(\Qbb,\Pbb_n)\leq \delta}\Ebb_{\Qbb} [ \ell(Y-f(X))]$. Here,
$\cF_n$ represents an estimation function space that may vary with $n$. We require $\cF_n$ to be sufficiently large to provide a good approximation to the H\"{o}lder class, while remaining computationally efficient for numerical studies. Motivated by the efficient approximation ability and widespread use of feedforward neural networks (FNNs) in high-dimensional settings, we construct  $\cF_n$ using FNNs.

 \subsection{FNNs with norm constraints}\label{sec_FNN}
 An FNN can be represented as
\begin{equation*}
 g=g_{L}\circ g_{L-1}\circ\cdots\circ g_{0},
\end{equation*}
 where $g_{\ell}(\bx)=\sigma_{m}(\bA_{\ell}\bx+\bb_{\ell})$  for $\ell=0,\dots, L-1$, and $g_{L}(\bx)=\bA_{L}\bx$. Here  $\bA_{\ell}\in\Rbb^{d_{\ell+1}\times d_{\ell}}$ and $\bb_{\ell}\in\Rbb^{d_{\ell+1}}$  denote weight matrices and  bias vectors, respectively, with  $d_{\ell}\in\mathbb{N}$, $d_0=d$, and $d_{L+1}=1$. \change{The activation function is defined as $\sigma_m(x) = (\max\{x,0\})^m$ for $m\in\mathbb{N}$, corresponding to the rectified power unit (RePU) of order $m$, applied componentwise. The RePU is $(m-1)$-times continuously differentiable.}  The FNN is then parameterized by $\theta=(\bA_0,\dots,\bA_{L}, \bb_{0},\dots, \bb_{L-1})$. Its width is defined as $W=\max\{d_1,\dots,d_L\}$, depth as $L$, and size as $\mathcal{S}=\sum_{s=0}^{L-1}d_{s+1}(d_{s}+1)$. The class of FNNs with width at most $W$, depth at most $L$, and size at most $\mathcal{S}$ is denoted by $\cN\cN(W,L,\mathcal{S})$.

 \change{We further consider a norm-constrained subclass. For $K>0$,   define   $\cN\cN(W,L,\mathcal{S},K)\subseteq \cN\cN(W,L,\mathcal{S})$ as the class of FNNs whose parameters satisfy
 \begin{equation*}
 	\kappa(\theta)  =   m^L \|\bA_L\|_{\infty} \prod_{s=0}^{L-1} \max \{\| (\bA_{s},\bb_{s})\|_{\infty},1\}^{m^{L-s}}\leq K,
 \end{equation*}
where the matrix infinity norm is defined by $\|\bA\|_{\infty}=\max_i\sum_{j}|a_{ij}|$ for   $\bA=(a_{ij})$.
 For any $g_{\theta}\in\cN\cN(W,L,\mathcal{S},K)$, the Lipschitz constant is bounded by
 \[
 \|g_{\theta}\|_{\Lip}  =\sup_{\bx_1\ne \bx_2}\frac{|g_{\theta}(\bx_1)-g_{\theta}(\bx_2)|}{\|\bx_1-\bx_2\|_{\infty}}\leq \kappa(\theta)\leq K.
 \]
 Hence, all FNNs  in $\cN\cN(W,L,\mathcal{S}, K)$ admit a uniform Lipschitz bound.  Moreover, they also satisfy a uniform Hessian bound,
 	\[
 \sup_{\bx\in\cX}	\|\nabla^2 g_{\theta}(\bx)\|_{\infty} \leq L m^{L} K.
 	\]
 A detailed verification is provided in Lemma~A.4 of the Supplementary Material. Notably, these norm constraints are not only theoretically motivated but also practically effective for enhancing model stability and robustness in tasks such as image classification \cite{cisse2017parseval, tsuzuku2018lipschitz}.

 Next, we establish two key properties of $\cN\cN(W,L,\mathcal{S},K)$, its complexity and approximation ability, which provide the foundation for the main theoretical analysis in Section~\ref{sec_main}.

 \begin{defi}[Covering number]
 	Let $(\mathcal{H},\|\cdot\|_{\mathcal{H}})$ denote a class of functions $\cH$ equipped with the norm $\|\cdot\|_{\mathcal{H}}$.
 	For $u>0$,  the $u$-covering number $\mathcal{N}\left(u, \mathcal{H},\|\cdot\|_{\mathcal{H}}\right)$  is defined as the smallest cardinality of a $u$-cover of $\mathcal{H}$, where the set $H_{u}$ is a $u$-cover of $\mathcal{H}$ if for each $h \in \mathcal{H}$, there exists $ h^{\prime} \in H_u$ such that $\|h^{\prime}-h\|_{\mathcal{H}} \leq u$.
 \end{defi}

 \begin{theorem}\label{lem_coverFNN}
 Let $\cN\cN(W,L,\mathcal{S},K)$  be defined over $  \mathcal{X}$. For any $u > 0$, the following bounds hold.
 	 	   (i) For the supremum norm $\|\cdot\|_{\infty}$, we have
 	 	  	\begin{equation*}
 	 	  		\begin{split}
 	 	  			\log \cN(u, \cN\cN(W,L,\mathcal{S},K) ,\|\cdot\|_{\infty})\lesssim \mathcal{S}\log\left(1+ Ku^{-1} \right).
 	 	  		\end{split}
 	 	  	\end{equation*}
 	 	  	(ii) If further $m \ge 2$, then
 	 	  	\begin{equation*}
 	 	  		\log \cN(u, \cN\cN(W,L,\mathcal{S},K), \rho_\infty) \lesssim \mathcal{S}\log\left(1+ LK m^Lu^{-1} \right),
 	 	  	\end{equation*}
 	 	  	where   $\rho_\infty$ is defined as $\rho_\infty(g,\tilde{g}) = \sup_{\bx\in \mathcal{X}}   \{ |g(\bx)-\tilde{g}(\bx)| + \|\nabla g(\bx)-\nabla \tilde{g}(\bx)\|_1 \}.$
 	 \end{theorem}

 Theorem~\ref{lem_coverFNN} establishes complexity bounds for $\cN\cN(W, L,  \mathcal{S}, K)$. Classical results such as those based on the pseudo-dimension (see, for example \cite{JMLRPeter2019}) typically depend only on the network architecture $(W,L)$ and ignore parameter norms, which often leads to loose estimates for modern over-parameterized models.  In contrast, Theorem~\ref{lem_coverFNN} provides improved bounds by simultaneously incorporating the norm constraint $K$ to capture implicit regularization, utilizing  $\mathcal{S}$ to reflect the effective model size, and enforcing control over both function values and gradients through the metric $\rho_\infty$. These features lead to sharper and more practical bounds for the  norm-constrained FNNs.

 \begin{theorem}\label{lem_appFNN}
	        For any $f\in\mathcal{H}^\alpha(\mathcal{X})$, if  $\cN\cN(W,L,\mathcal{S},K)$ satisfies $W\!\ge\! N$,  $\mathcal{S}\!\ge\! N(d+1)$, and $K\!\gtrsim\!\max\{N^{\frac{d+2m+1-2\alpha}{2d}},\sqrt{\log N}\}$, then
			there exists $g\in\cN\cN(W,L,\mathcal{S},K)$ such that
			\begin{align*}
			\sup_{\bx\in\mathcal{X}}|g(\bx)-f(\bx)|&\lesssim
			N^{-\frac{\min\{d+2m+1,2\alpha\}}{2d}}\sqrt{\log N},\\
			\sup_{\bx\in\mathcal{X}}\|\nabla(g-f)(\bx)\|_\infty&\lesssim
			N^{-\frac{\min\{d+2m-1,2(\alpha-1)\}}{2d}}\sqrt{\log N}.
			\end{align*}
\end{theorem}

Theorem~\ref{lem_appFNN} establishes simultaneous approximation error bounds for both function values and gradients when approximating functions in $\mathcal{H}^\alpha(\mathcal{X})$ with norm-constrained FNNs. Whereas previous results have primarily focused on characterizing rates in terms of width and depth \cite{YAROTSKY2017103,Schmidt2020,LU2021}, Theorem~\ref{lem_appFNN}  shows the effect of the norm constraint $K$ on the achievable accuracy. Notably, the requirement that $K$ must grow with $N$ reflects an intrinsic limitation of norm-constrained FNNs, a finding consistent with the lower-bound analyses of \cite{jiao2022approximation}.

Following \cite{bach2017breaking,yang2024optimal}, the proof strategy first approximates the $\mathcal{H}^\alpha(\mathcal{X})$ function via an integral representation, which is in turn approximated by norm-constrained FNNs. The resulting piecewise convergence rate,  exhibiting a breakpoint at $\alpha = (d+2m+1)/2$, arises from balancing these two sources of approximation error.  This theorem improves upon previous results by explicitly incorporating the simultaneous gradient approximation, which is a key component of our analysis in the $k>1$ case. For clarity and completeness, we also formally introduce the integral representation class that serves as the intermediate step in the proof strategy.

	\begin{remark} \label{remark1}
		Consider functions in the form as
		$f(\bx) = \int_{\mathbb{S}^d} \sigma_m\big((\bx^\top,1)\bv\big)  d\mu(\bv),$
		where $\mu$ is a signed Radon measure on $\mathbb{S}^d =\{\bv\in\Rbb^{d+1}:\|\bv\|_2=1\}$, and define the associated norm
		\[
		\|f\|_{\mathcal{B}(\mathbb{P}_{X})}  = \inf \Big\{ \|\mu\| : f(\bx) = \int_{\mathbb{S}^d} \sigma_m\big((\bx^\top,1)\bv\big)\, d\mu(\bv) \quad \mathbb{P}_{X}\text{-a.e.} \Big\},
		\]
		with $\|\mu\|$ the total variation norm. The class of functions with finite $\mathcal{B}(\mathbb{P}_{X})$-norm arises naturally as the space of infinitely wide two-layer neural networks, known as the variation space for RePU activations \cite{siegel2023optimal}. This function space is also closely related to the Barron space, originally introduced in \cite{Barron1993} and extended in \cite{E2022}, where the above integral representation is established.
		
		This class exhibits favorable approximation properties and, in certain regimes, can circumvent the curse of dimensionality. In particular,  \cite{siegel2023optimal} shows that for any $f$ with $\|f\|_{\mathcal{B}(\mathbb{P}_X)} < K$ and any $0 \le r \le m$,  there exists $g \in \cN\cN(W,1,W(d+1),mK(d+1)^{m/2})$ such that
		\[
		\sup_{\|\bs\|_1 \le r}
		\big\| D^{\bs}(f - g) \big\|_{L^\infty(\mathbb{B}^d)}
		\;\lesssim\; K W^{-\frac{1}{2} - \frac{2(m-r)+1}{2d}}.
		\]
	Hence, the norm-constrained FNNs provide uniform Sobolev-type approximation to Barron functions. This also implies that our results can extend to Barron space when approximation error is quantified in this manner.
	\end{remark}
}


\section{Main results}\label{sec_main}
This section presents our main non-asymptotic error bounds for both the excess local worst-case risk and the excess natural risk under a shifted distribution. To facilitate the analysis,  the following assumption is needed.
\change{
\begin{assumption}\label{ass_Lipf}
	Suppose $f_0\in\cH^{\alpha}$ and there exist constants   $c_1,c_2,c_3,c_4>0$  such that the following conditions hold for all $f\in \cF_n $.
		
		(i)  $\mathcal{R}_{\Pbb}(f) -\inf_{f\in\cH^{\alpha}} \cR_{\Pbb}(f) \le  c_1+c_2\|f - f_0\|_{2}^2.$
		
		(ii)  $\cR_{\Pbb}(f) - \cR_{\Pbb}(f_0) \ge c_3
		\|f - f_0\|_{2}^2$ when  $\|f - f_0\|_{2}\geq c_4$.
	\end{assumption}

Assumption~\ref{ass_Lipf}  establishes a risk–norm equivalence condition linking the excess natural risk to the $L^2(\Pbb)$-norm $\|f-f_0\|_2^2=\Ebb_{\Pbb}[|f(X)-f_0(X)|^2]$, a standard regularity condition for achieving sharp convergence rates in nonparametric regression \cite{Bartlett2005,Farrell2021Eco}. The upper bound in (i) controls the approximation error, while  $c_1$ accounts for the bias that arises when $f_0$ is not the exact natural risk minimizer, as in robust estimation settings (see Section~\ref{sec_Huber}). Condition (ii) imposes localized strong convexity,  which enables effective control of the stochastic error in empirical risk minimization. Assumption~\ref{ass_Lipf} holds for a wide class of   loss functions, including the quadratic, Huber, and check losses, as explicitly verified in Section~\ref{sec_app}.}

\mainchange{
	\begin{remark}
		The constants $c_1$ and $c_4$ in Assumption~\ref{ass_Lipf} are introduced to allow for a more general framework. In many standard nonparametric settings,
$f_0$ is the exact minimizer of the natural risk, i.e., $f_0 = \arg\min_{f \in \mathcal{H}^{\alpha}} \mathcal{R}_{\mathbb{P}}(f)$.
In this situation, no bias is introduced, and one may set $c_1 = 0$ and $c_4 = 0$. This holds, for example, for least squares and quantile regression under appropriate conditions, as verified in Section~\ref{sec_app}. Conversely, in other settings, $f_0$ may not be the exact population-level minimizer. This occurs, for example, in robust regression with the Huber loss. In such cases, a bias term appears (captured by $c_1 > 0$), making the constants $c_1$ and $c_4$ non-negligible. Both scenarios are formally verified in Section~\ref{sec_app}.
\end{remark}}

For the estimation function space, let $\cF_n$ denote the truncated class $\cN\cN(W,L,\cS,K)$ at level $M_n>0$, where $M_n$ may grow with $n$. Formally,
\[
\cF_n=\big\{T_{M_n}g:\; g\in\cN\cN(W,L,\cS,K)\big\},
\]
where $T_{M_n}$ is the truncation operator restricting the output to $[-M_n, M_n]$, defined by $T_{M_n} y = y$ if $|y| \le M_n$ and $T_{M_n} y = M_n (y/|y|)$ otherwise. Allowing $M_n$ to diverge at an appropriate rate ensures that the approximation properties of $\cN\cN(W,L,\cS,K)$ are essentially preserved. With this choice of $\cF_n$, the estimator $\widehat{f}_{k,\delta}^{(n)}$ is defined as
	\begin{equation}\label{eq_estimator}
	\widehat{f}_{k,\delta}^{(n)}\in\argmin_{f\in \cF_n} \cR_{\Pbb_n,k}(f;\delta).
	\end{equation}


\subsection{Non-asymptotic error bounds for  excess local worst-case  risk}
To evaluate the performance of the estimator $	\widehat{f}_{k,\delta}^{(n)}$,   we consider the excess local worst-case risk defined by
\begin{equation*}
\cE_{\Pbb,k}(f;\delta)= \cR_{\Pbb,k}(f;\delta)-\inf_{f\in\cH^{\alpha}\cup \cF_n}\cR_{\Pbb,k}(f;\delta).
\end{equation*}
This is nonnegative and measures the gap between the estimator's population-level risk and the optimal achievable risk. Smaller values indicate that the estimator is closer to the theoretical optimum.  We analyze the non-asymptotic behavior of $\cE_{\Pbb,k}(\widehat{f}_{k,\delta}^{(n)};\delta)$ for the cases $k=1$ and $k>1$ separately. The analysis builds upon the robustness regularization properties characterized in Lemmas~\ref{lem_regu} and~\ref{lem_regu2}.
\change{
	\begin{theorem}[Case $k=1$]\label{thm_main1}
			Suppose Assumption~\ref{ass_Lipf}  holds with $\|\ell\|_{\Lip}<\infty$.  Assume $m\geq 1$, $W\ge  N,\mathcal{S}\ge N(d+1),$ $ K\!\gtrsim\max \{N^{\frac{d+2m+1-2\alpha}{2d}},\sqrt{\log N} \},$ and $M_n\asymp\log n$. Let $\beta =  \min\{(d+2m+1)/2, \alpha\}$. Then, for any $\delta>0$,  we have
\begin{equation*}
	\begin{split}
		\Ebb\big[{\cal E}_{\Pbb,1}(\widehat{f}_{1,\delta}^{(n)};\delta) \big]  &	\lesssim   c_1+ c_4^2+ N^{-\frac{2\beta}{d}}\log N+ n^{-1}\mathcal{S}\log(Kn)\log n+ \delta  K
		\\&	\qquad \; +
		\exp\{-\mathcal{S}\log(Kn)\}\log n.
	\end{split}
\end{equation*}
If further $\mathcal{S} \asymp W \asymp N \asymp n^{\frac{d}{d+2\beta}}$ and $K \asymp \max\big\{n^{\frac{d+2m+1-2\alpha}{2(d+2\beta)}},\sqrt{\log n}\big\}$, then
\begin{equation*}
	\Ebb\big[{\cal E}_{\Pbb,1}(\widehat{f}_{1,\delta}^{(n)};\delta) \big]\lesssim c_1+ c_4^2 + n^{-\frac{ 2\beta}{d+2\beta}}\{\log n\}^2+ \delta K.
\end{equation*}
\end{theorem}

\begin{cor}\label{cor_main1}
Suppose the conditions of Theorem~\ref{thm_main1} hold, and assume $c_1 = c_4 = 0$ and $\delta = o(n^{- \frac{d + 2m + 1 + 2\beta}{2(d + 2\beta)}})$. Then,
\begin{equation*}
	\Ebb\big[{\cal E}_{\Pbb,1}(\widehat{f}_{1,\delta}^{(n)};\delta) \big]\lesssim  n^{-\frac{ 2\beta}{d+2\beta}}\{\log n\}^2.
\end{equation*}
\end{cor}

Theorem~\ref{thm_main1} establishes a non-asymptotic error bound on the excess local worst-case risk for the case $k=1$. The bound explicitly decomposes the error into several distinct components: a population-level bias ($c_1+c_4^2$), an approximation error ($N^{-\frac{2\beta}{d}}\log N$), a stochastic error ($n^{-1}\mathcal{S}\log(Kn)\log n$), a term linear in the uncertainty level that reflects the cost of robustness ($\delta K$), and a residual term that decays exponentially.

 Corollary~\ref{cor_main1} further shows the convergence rate under optimally chosen tuning parameters. It indicates that, in a bias-free setting (i.e., $c_1 = c_4 = 0$) and when $\delta$ decreases sufficiently fast, the estimator achieves the convergence rate of $n^{-2\beta/(d+2\beta)}$ up to logarithmic factors.  \mainchange{This rate is governed by $\beta = \min\{(d+2m+1)/2, \alpha\}$. When $\beta = \alpha$, a condition often satisfied in high-dimensional settings, the rate $n^{-2\alpha/(d+2\alpha)}$ matches the minimax optimal rate for nonparametric regression \citep{stone1982optimal}. In the complementary regime, the rate becomes $n^{-(d+2m+1)/(2d+2m+1)}$, which corresponds to the minimax optimal rate for the Barron space of norm at most $1$ when $m=1$ \cite{Parhi2023}. This dual-rate characteristic arises from the proof technique used in Theorem~\ref{lem_appFNN}, where the Barron space serves as an intermediate function class. The overall convergence is thus dictated by the slower of the two approximation scales. The analysis framework is general and can incorporate sharper approximation bounds for norm-constrained FNNs to further refine the established rate. }

  \begin{theorem}[Case $k>1$]\label{thm_main2}
	Suppose Assumption~1 holds with  $\|\ell^{'}\|_{\infty}, \|\ell^{'}\|_{\Lip}<\infty$.   Assume  $\alpha>1$, $m\geq 2$, $W\ge  N,\mathcal{S}\ge N(d+1),$ $ K\!\gtrsim\max \{N^{\frac{d+2m+1-2\alpha}{2d}},\sqrt{\log N} \},$ and $M_n\asymp\log n$. Let  $\beta = \min\{(d+2m+1)/2, \alpha\}$. Then, for any $\delta>0$,  we have
	\begin{equation*}
		\begin{split}
		\Ebb[{\cal E}_{\Pbb,k}(\widehat{f}_{k,\delta}^{(n)};\delta)  ]
		& \lesssim  c_1+c_4^2+   N^{-\frac{2\beta}{d}}\log N
		+  n^{-1}\mathcal{S}\log(Kn) \log n\nonumber\\
		&\qquad	
		+\delta \big\{1+N^{-\frac{\beta-1}{d}}\sqrt{\log N}+ \{n^{-1}\mathcal{S}\}^{1/(2k^\star)} \{L  m^L    \}^{1/k^\star}K\big\}
		\\  &\qquad   + \delta^{\min\{k,2\}}(K^2+Lm^{L}K)
		+ \exp\{-\mathcal{S}\log(Kn)\}\log n.
		\end{split}
	\end{equation*}
	If further $\mathcal{S} \asymp W \asymp N \asymp n^{\frac{d}{d+2\beta}}$ and $K \asymp \max\big\{n^{\frac{d+2m+1-2\alpha}{2(d+2\beta)}},\sqrt{\log n}\big\}$, then
	\begin{align*}
		\Ebb\big[{\cal E}_{\Pbb,k}(\widehat{f}_{k,\delta}^{(n)};\delta) \big]&\lesssim
		c_1+c_4^2+ n^{-\frac{2\beta}{d+2\beta}} \{\log n\}^2 + \delta  \{1+ n^{-\frac{    \beta}{k^\star(d+2\beta)}} K \}
		+ \delta^{\min\{k,2\}} K^2.
	\end{align*}
\end{theorem}

 \begin{cor}\label{cor_main2}
	Suppose the conditions of Theorem~\ref{thm_main2} hold, and assume $c_1 = c_4 = 0$ and  $\delta  = o(n^{-\frac{c }{2(d+2\beta)}})$, where $c=\max\{4\beta, 2k^{-1}(d+2m+1), 2k^{-1}\beta+d+2m+1\}$. Then,
	\begin{equation*}
	\Ebb\big[{\cal E}_{\Pbb,k}(\widehat{f}_{k,\delta}^{(n)};\delta) \big] \lesssim n^{-\frac{2\beta}{d+2\beta}}\{\log n\}^2.
	\end{equation*}
 \end{cor}

Theorem~\ref{thm_main2} and Corollary~\ref{cor_main2} consider the case $k>1$ and establish a non-asymptotic error bound showing that the estimator can attain the rate $n^{-2\beta/(d+2\beta)}$, up to logarithmic factors, provided that $\delta$ decreases sufficiently fast. It is worth noting that the robustness-related terms exhibit a fundamentally different structure compared with the $k=1$ case. Instead of the simple $\delta K$ term, the bound contains two components, $\delta\{1+n^{-\frac{\beta}{k^\star(d+2\beta)}} K\}$ and $\delta^{\min\{k,2\}} K^2$, corresponding to first-order and higher-order regularization effects. This difference reflects the structural distinctions of the Wasserstein ambiguity sets, with $W_1$ inducing Lipschitz-type regularization as shown in Lemma~\ref{lem_regu}, and $W_k$ for $k>1$ introducing gradient-norm based higher-order effects as shown in Lemma~\ref{lem_regu2}. Taking the limit as $k \to 1^+$, the first component converges to $\delta K$, consistent with Theorem~\ref{thm_main1}, whereas the second approaches $\delta K^2$, which has no counterpart in the $k=1$ analysis as higher-order effects are not considered.

Corollary~\ref{cor_main2} also implies a trade-off between robustness strength and uncertainty tolerance. As $k$ increases, the constant $c$ in the required decay rate of $\delta$ decreases, indicating that larger $k$ imposes a weaker constraint on $\delta$. Consequently, the estimator can accommodate a greater level of uncertainty in the $W_k$ sense while preserving the optimal convergence rate. This behavior arises from the geometry of Wasserstein balls, where for $q>k$ the set $\{\Qbb: W_k(\Pbb,\Qbb)\le \delta\}$ contains $\{\Qbb: W_q(\Pbb,\Qbb)\le \delta\}$. Therefore, smaller values of $k$ provide stronger robustness guarantees, whereas larger $k$ allow for higher uncertainty levels, showing the need for careful design of robust models.}

\subsection{Robustness in the presence of model uncertainty}
An important property of the distributionally robust nonparametric estimator is its ability to perform reliably across a range of possible distributions, ensuring robust performance even in the presence of model uncertainty. Specifically, we define the excess natural risk as
\begin{equation}\label{eq_natralexcess}
{\cal E}_{\Pbb^{\prime}}(f)=\cR_{\Pbb^{\prime}}(f)- \inf_{f\in\cH^{\alpha}}  \cR_{\Pbb^{\prime}}(f),
\end{equation}
where the distribution $\Pbb^{\prime}$ is explicitly included in the notation to emphasize the specific distribution under which the natural risk is defined.

\change{ \begin{theorem}\label{thm_na}
	 Assume $\|\ell\|_{\Lip}<\infty$. For any  $\Pbb^{\prime}$ satisfying $W_k(\Pbb^{\prime},\Pbb)\leq \delta$, it holds that
\begin{align*}
	{\cal E}_{\Pbb^{\prime}}(f) \lesssim  \cE_{\Pbb,k}(f;\delta)+ \delta \|\ell \|_{\Lip}.
\end{align*}
Moreover, under the conditions of Theorem~\ref{thm_main1} for $k=1$, it holds that
\begin{align*}
 	\Ebb\big[	{\cal E}_{\Pbb^{\prime}}(\widehat{f}_{1,\delta}^{(n)})  \big]\lesssim c_1+ c_4^2 + n^{-\frac{ 2\beta}{d+2\beta}}\{\log n\}^2+ \delta K.
\end{align*}	
Alternatively, under the conditions of Theorem~\ref{thm_main2} for $k>1$, it holds that
\begin{align*}
\Ebb\big[	{\cal E}_{\Pbb^{\prime}}(\widehat{f}_{k,\delta}^{(n)}) \big] \lesssim c_1+c_4^2+ n^{-\frac{2\beta}{d+2\beta}} \{\log n\}^2 + \delta  \{1 + n^{-\frac{    \beta}{k^\star(d+2\beta)}} K \} + \delta^{\min\{k,2\}} K^2.
\end{align*}	
 \end{theorem}}

Theorem \ref{thm_na} shows that a smaller excess local worst-case risk leads to a correspondingly smaller excess natural risk under any distribution within the ambiguity set. In contrast, existing works \cite{KuhnMohajerin2018,Fournier2015,azizian2023exact,Shafieezadehmass2019, le2024universal, an2021generalization, gao2023finite} primarily focus on establishing bounds of the form:
\[
\cR_{\Pbb}(f)\leq \cR_{\Pbb_n,k}(f;\delta).
\]
These works typically aim to identify a suitable $\delta$ such that the empirical WDRO objective upper bounds the expected loss under the data-generating distribution. Theorem \ref{thm_na}, however, offers a distinct and more comprehensive perspective. It explicitly shows the impact of $\delta$ in controlling the excess natural risk under distributional perturbations, while also integrating the effect of model misspecification. This leads to a more detailed and precise evaluation of the estimator’s performance under both distributional and model uncertainty.

\change{\begin{cor}\label{cor_na}
			Suppose the conditions of Theorem~\ref{thm_na} hold and $c_1 = c_4 = 0$. Let $\delta = o(n^{- \frac{d + 2m + 1 + 2\beta}{2(d + 2\beta)}})$ for $k=1$ and $\delta  = o(n^{-\frac{c }{2(d+2\beta)}})$ for $k>1$, where $c=\max\{4\beta, 2k^{-1}(d+2m+1), 2k^{-1}\beta+d+2m+1\}$. For any $\Pbb^{\prime}$ satisfying $W_k(\Pbb^{\prime},\Pbb)\leq \delta$, it holds that
		\begin{align*}
			\Ebb\big[	{\cal E}_{\Pbb^{\prime}}(\widehat{f}_{k,\delta}^{(n)}) \big] \lesssim  n^{-\frac{2\beta}{d+2\beta}} \{\log n\}^2.
		\end{align*}	
\end{cor}}

 \noindent {\bf Continuation of Examples \ref{ex1} and \ref{ex3}.}
In the two examples, we illustrate simple cases of contaminated distributions and distributional inconsistency. In these scenarios, the problem arises from the inconsistency between the observation and target data distributions. Specifically, when the target data distribution $\Pbb^{\prime}$ is located within a $\delta$-Wasserstein ball, Theorem \ref{thm_na} shows that the excess natural risk of any estimator $f$ under this target distribution is bounded by ${\cal E}_{\Pbb^{\prime}}(f) \lesssim \cE_{\Pbb,k}(f;\delta)+ \delta  \|\ell \|_{\Lip}.$ This result guarantees that the estimator performs well under the target distribution, provided that $f$ maintains a small local worst-case risk.

\section{Applications}\label{sec_app}
To illustrate the results in Section~\ref{sec_main}, we consider four examples: nonparametric least squares regression, nonparametric robust regression, nonparametric quantile regression, and classification. In each example, Assumption~\ref{ass_Lipf} is explicitly verified, and the analysis is adapted to the corresponding setting.

 \subsection{Nonparametric least squares regression}\label{sec_ls}
Suppose the error $\varepsilon$ in regression model \eqref{model_re} satisfies $\Ebb[varepsilon|X]=0$
and $\Ebb(\varepsilon^2) < \infty.$
	Consider the quadratic loss $\ell(u)=u^2$, whose natural risk minimizer is the conditional mean, i.e., $f_0 =\argmin_{f \text{ measurable}}  \Ebb_{ \Pbb} [ \{Y-f(X)\}^2 ].$ It can be verified that Assumption~\ref{ass_Lipf} holds for the quadratic loss with $c_1=c_4=0$ and $c_2=c_3=1$. The corresponding distributionally robust nonparametric estimator is defined by
	\[
	\widehat{f}_{k,\delta}^{(n)}\in  \argmin_{f\in \cF_n} \sup_{\Qbb: W_k(\Qbb,\Pbb_n)\leq \delta}\Ebb_{ \Qbb}\left[\{Y-f(X)\}^2\right].
	\]
	Although the exact minimizer of the local worst-case risk is generally intractable due to the supremum over an infinite set, some intuition can be gained from the following analysis:
	 \begin{align*}
	 	\inf_{f \operatorname{measurable}} \sup_{\Qbb: W_k(\Qbb,\Pbb)\leq \delta }\Ebb_{ \Qbb}\left[ \{Y-f(X)\}^2 \right] &\geq \sup_{\Qbb: W_k(\Qbb,\Pbb)\leq \delta }\inf_{f \operatorname{measurable}}\Ebb_{ \Qbb}\left[ \{Y-f(X)\}^2 \right]
	 	\\&\geq\sup_{\Qbb: W_k(\Qbb,\Pbb)\leq \delta } \Ebb_{\Qbb}\left[ \{Y-\Ebb_{\Qbb}[Y|X]\}^2 \right],
	 \end{align*}
	 where $\Ebb_{\Qbb}[Y|X]$ denotes the conditional expectation  under   $\Qbb.$  This suggests that the distributional perturbation tends to find a distribution near $\Pbb$ that maximizes the variance of the residual $Y-\Ebb_{\Qbb}[Y|X]$. Establishing equality in the above inequalities requires additional conditions, presenting an interesting direction for future research.	
	
 \begin{assumption}\label{ass1}
 		There exists a constant $\sigma_{Y}>0$ such that $\Ebb[\exp\{\sigma_{Y}|Y|\}]<\infty.$
 	\end{assumption}  	
 	Assumption~\ref{ass1} is a mild condition on $Y$, allowing unbounded cases with sub-exponential tails and relaxing the requirement of strict boundedness. This assumption  is important for deriving non-asymptotic error bounds for the excess local worst-case risk under the quadratic loss. The main challenge is controlling the unbounded derivatives of the loss. We address this by employing a truncation technique that restricts the analysis to a bounded region where the derivatives are controlled, while tail behavior is accounted for through Assumption~\ref{ass1} and the Wasserstein constraint. Moreover, standard least squares regression requires $\Ebb[\varepsilon^2] < \infty$, and to ensure this condition holds under perturbations within the ambiguity set, we focus on $k \ge 2$.

\change{	\begin{theorem}\label{thm_ls}
Suppose that $\Ebb(\varepsilon|X)=0$, $\Ebb(\varepsilon^2) < \infty,$
and Assumption~\ref{ass1} holds.  Assume $\alpha\!>\!1$, $m\!\geq \!2$, $W \! \ge \! N,\mathcal{S}\!\ge\! N(d+1),$ $ K\!\gtrsim \!\max \{N^{\frac{d+2m+1-2\alpha}{2d}},\sqrt{\log N} \},$ and $M_n\asymp\log n$. Let $\beta = \min\{(d+2m+1)/2, \alpha\}$. Then, for any $\delta>0$ and $k\geq 2$, we have
		\begin{equation*}
			\begin{split}
				\Ebb\big[ {\cal E}_{\Pbb,k}(\widehat{f}_{k,\delta}^{(n)};\delta) \big] & \lesssim N^{-\frac{2\beta}{d}}\log N + n^{-1}\mathcal{S}\log(Kn)\{ \log n \}^3 + n^{-1} \log n \\
				&\qquad + \delta \big\{ 1+ N^{-\frac{\beta-1}{d}}\sqrt{\log N} + \{n^{-1}\mathcal{S}\}^{1/(2k^\star)} \{L m^L  \}^{1/k^\star}K \big\}\log n \\
				&\qquad + \delta^2 \big\{1+K^2+Lm^{L}K\big\}\log n.
			\end{split}
		\end{equation*}
		If further $\mathcal{S} \asymp W\asymp N \asymp n^{\frac{d}{d+2\beta}}$ and $K \asymp \max\big\{n^{\frac{d+2m+1-2\alpha}{2(d+2\beta)}},\sqrt{\log n}\big\}$, then
		\begin{align*}
			\Ebb\big[{\cal E}_{\Pbb,k}(\widehat{f}_{k,\delta}^{(n)};\delta) \big]  \lesssim n^{-\frac{2\beta}{d+2\beta}}\{ \log n \}^4 + \delta \big\{1  + n^{-\frac{ \beta}{k^\star(d+2\beta)}} K \big\}\log n
			+ \delta^2 K^2\log n.
		\end{align*}
		In particular, if $ \delta = o\big(n^{- \frac{c}{2(d+2\beta)} }\big)$, with $c=\max\{4\beta, 2k^{-1}\beta+d+2m+1\}$,  then
		\[
		\Ebb\big[{\cal E}_{\Pbb,k}(\widehat{f}_{k,\delta}^{(n)};\delta) \big] \lesssim n^{-\frac{2\beta}{d+2\beta}}\{ \log n \}^4.
		\]
	\end{theorem}
	Theorem~\ref{thm_ls} shows that the theoretical results established in Theorem~\ref{thm_main2} extend to least squares regression, despite the quadratic loss not satisfying the bounded derivative condition.
	For sufficiently small $\delta$, the proposed estimator achieves a convergence rate of $n^{-2\beta/(d+2\beta)}$, up to logarithmic factors. Further, Theorem~\ref{thm_ls} enables us to derive bounds on the $L^2$-error under distributional shifts.  Specifically, for any distribution $\Pbb^{\prime}$ and $(X^{\prime},Y^{\prime}) \sim \Pbb^{\prime}$, let the corresponding target function be $f_0^{\prime}(\bx) = \Ebb_{\Pbb^{\prime}}[Y^{\prime} | X^{\prime}=\bx],$ and define the $L^2(\Pbb^{\prime}_X)$-norm by $\|f\|_{L^2(\Pbb^{\prime}_X)} = \Ebb_{\Pbb^{\prime}}[|f(X^{\prime})|^2]^{1/2}.$

 	\begin{theorem}\label{thm_ls2}
				Suppose the conditions of Theorem 5.1 hold and $ \delta = o\big(n^{- \frac{c}{2(d+2\beta)} }\big)$, with $c=\max\{4\beta, 2k^{-1}\beta+d+2m+1\}$. For any    $\Pbb^{\prime}$ satisfying  $W_k(\Pbb^{\prime}, \Pbb)\leq \delta$, and assuming $f_0^{\prime}\in\cH^{\alpha}$, we have
			\begin{align*}
				\Ebb\big[	\|\widehat{f}_{k,\delta}^{(n)}-f_0^{\prime}\|_{L^2(\Pbb^{\prime}_X)}^2\big] \lesssim n^{-\frac{2\beta}{d+2\beta}}\{ \log n \}^4.
			\end{align*}
		\end{theorem}
Theorem~\ref{thm_ls2} indicates that for any $\Pbb^{\prime}$ within the $W_k$ ambiguity set, the estimator achieves a convergence rate of $n^{-2\beta/(d+2\beta)}$, up to logarithmic factors,  in the $L^2(\Pbb^{\prime}_X)$-norm with respect to its corresponding target function $f_0^{\prime}$. \mainchange{As discussed in the general analysis in Section~\ref{sec_main}, this rate coincides with the minimax optimal rate for nonparametric regression in $\cH^{\alpha}$ when $\alpha$ is less than $(d+2m+1)/2$, a condition often satisfied in high-dimensional settings.}

 Our analysis extends existing work to a distributionally robust framework. Previous studies have shown that least squares regression with deep neural networks can achieve the  minimax optimal convergence rate \cite{Schmidt2020,Kohler2021,jiao2021deep}, and norm-constrained FNNs in least-squares settings have been analyzed in \cite{jiao2022approximation,Yang2024}. In contrast, the present work extends these studies by explicitly incorporating model uncertainty and provides theoretical guarantees under distributional variation.}

\subsection{Nonparametric robust regression}\label{sec_Huber}
 Suppose the error in the regression model \eqref{model_re} satisfies $\Ebb[\varepsilon|X]=0.$
To reduce the sensitivity of the quadratic loss to outliers in response, a common approach is to consider the Huber loss $\ell_\tau(\cdot)$, defined by
 	\begin{align*}
 	\ell_\tau(u) = \begin{cases}
 	\frac{1}{2} u^2, & \qquad |u| \le \tau, \\
 	\tau |u| - \frac{1}{2}\tau^2, & \qquad |u| > \tau,
 	\end{cases}
 	\end{align*}
 	where  $\tau >0$ is a tuning parameter. This  loss is continuously differentiable, with derivative $\ell_\tau^{\prime}(u) = \min \{ \max( -\tau, u ), \tau \}$. The derivative is 1-Lipschitz and uniformly bounded by $\tau$. To further enhance robustness against  model uncertainty, we consider
 	\begin{equation*}
 	\widehat{f}_{\delta,\tau}^{(n)}\in\argmin_{f\in \cF_n} \sup_{\Qbb: W_k(\Qbb,\Pbb_n)\leq \delta}\Ebb_{\Qbb} [ \ell_{\tau}(Y-f(X))].
 	\end{equation*}
 	
 	The statistical properties of the natural risk $\cR_{\Pbb}(f) = \Ebb_{\Pbb}[\ell_\tau(Y - f(X))]$  are characterized by the conditional distribution of $\varepsilon$ given $X$, as described below.
 	 	\begin{assumption}\label{con_H1}
 	 	(i) There exists   $r\geq 1$ such that  $\mathbb{E} [ | \varepsilon|^r|X=\bx ] \le v_r< \infty $   for all $\bx\in\cX;$
 	 	(ii) The conditional  distribution of $\varepsilon$ given $X$   is symmetric around zero.
 	 \end{assumption}
 	 When only the $r$-th moment condition in Assumption~\ref{con_H1}(1) holds, \cite{fan2022noise} shows that for any $f \in \cF_n$, if $\tau \ge 2\max\{2M_n, (2v_r)^{1/r}\}$, then
 	 \begin{equation}\label{eq_Huber_ineq}
 	 -\tau^{1-r} v_r \|f - f_0\|_2 + \tfrac{1}{4}\|f - f_0\|_2^2
 	 \leq
 	 \cR_{\Pbb}(f) - \cR_{\Pbb}(f_0)
 	 \leq
 	 \tfrac{1}{2} \tau^{2-2r} v_r^2 + \|f - f_0\|_2^2,
 	 \end{equation}
 	 where $\|\cdot\|_2=\|\cdot\|_{L^2(\Pbb_{X})}$. 	 In this case, the  minimizer $f_\tau = \argmin_f \cR_{\Pbb}(f)$ may differ from $f_0$, with the bias bounded by $\|f_\tau - f_0\|_2 \le 4\tau^{1-r} v_r$. Combining this with~\eqref{eq_Huber_ineq} further yields
 	 \[
 	 \cR_{\Pbb}(f) - \cR_{\Pbb}(f_{\tau}) \leq 5\tau^{2-2r} v_r^2 + \|f - f_0\|_2^2.
 	 \]
 	 This result implies that Assumption~\ref{con_H1}(i) ensures Assumption~\ref{ass_Lipf}  with $c_1 = 5\tau^{2-2r} v_r^2$, $c_2 = 1$,   $c_3 = 1/8$, and $c_4=8\tau^{1-r} v_r$. Moreover,  if  the conditional distribution of $\varepsilon$ is symmetric, that is, Assumption~\ref{con_H1}(ii) also holds,  the inequalities in \eqref{eq_Huber_ineq} simplify to
 	 \[
 	 \tfrac{1}{4}\|f - f_0\|_2^2
     \leq
 	 \cR_{\Pbb}(f) - \cR_{\Pbb}(f_0)
      \leq
 	 \tfrac{1}{2}\|f - f_0\|_2^2.
 	 \]
 	 In this symmetric setting, the bias term vanishes, implying that  $f_0$ is the exact population Huber minimizer $f_\tau$, and no bias is introduced by the Huberization process.  Therefore, when Assumption~\ref{con_H1} holds,  Assumption~\ref{ass_Lipf} is satisfied with $c_1 = c_4 = 0$, $c_2 = 1/2$, and $c_3 = 1/4$.
 	
 \change{
    \begin{prop} \label{pro_HU}
  	Suppose the conditions of Theorem \ref{thm_na} and  Assumption \ref{con_H1} hold, and let $\tau \geq  2\max\{ 2 M_n,  (2v_r)^{1/r} \}$. If $k=1$, then
 	\begin{align*}
    \Ebb\big[ \|\widehat{f}_{\delta,\tau}^{(n)}-f_0\|_{L^2(\Pbb_{X})}^2\big]\lesssim  \tau^2 n^{-\frac{ 2\beta}{d+2\beta}}\{\log n\}^2+ \delta  \tau K,
 	\end{align*}
 	where    $K \asymp \max\big\{n^{\frac{d+2m+1-2\alpha}{2(d+2\beta)}},\sqrt{\log n}\big\}$.
 	Further, if $k>1$, then
 	\begin{align*}
 	  \Ebb\big[ \|\widehat{f}_{\delta,\tau}^{(n)}-f_0\|_{L^2(\Pbb_{X})}^2\big] \lesssim   \tau^2  n^{-\frac{2\beta}{d+2\beta}} \{\log n\}^2 + \delta  \tau \{1 + n^{-\frac{    \beta}{k^\star(d+2\beta)}} K \} + \delta^{\min\{k,2\}} \tau K^2.
 	\end{align*}	
    \end{prop}
 Proposition~\ref{pro_HU} establishes the $L^2$ convergence rate of the proposed estimator, quantifying the influence of  $\tau$. The bounds indicate that both the statistical error and the robustness-related terms grow with $\tau$. While a larger $\tau$ is required to accommodate non-outlier data with large magnitudes, it simultaneously reduces the estimator’s robustness by treating larger errors as inliers. Ths causes the theoretical error bound to increase with $\tau$.}

 A key challenge within this distributionally robust framework is that the noise symmetry guaranteed by Assumption~\ref{con_H1}(2) is generally not preserved under distributional perturbations. Consider a target distribution $\Pbb^{\prime}$  satisfying $W_k(\Pbb^{\prime},\Pbb)\leq \delta$, and denote the corresponding regression model by $Y^{\prime}=f_0^{\prime}(X^{\prime})+\varepsilon^{\prime}$. Even if the nominal noise $\varepsilon$ is symmetric, the perturbed noise $\varepsilon^{\prime}$ is generally not. For example, a simple perturbation of the form $Y^{\prime} = Y+\lambda\varepsilon_1$, where $\varepsilon_1$ is asymmetric  and $\lambda>0$ is small, will result in an asymmetric noise $\varepsilon' = \varepsilon + \lambda\varepsilon_1$. Consequently,
the Huberization bias, which is absent in the nominal model, reappears under the perturbed distribution.  Denote the minimizer of the natural risk under $\Pbb^{\prime}$ by
\[
f_{0,\tau}^{\prime}=\argmin_{f\in\cH^{\alpha}}\Ebb_{\Pbb^{\prime}}[\ell_{\tau}(Y^{\prime}-f(X^{\prime}))].
\]
\change{When $\Pbb^{\prime}$ also satisfies the $r$-th moment condition in Assumption~\ref{con_H1}(1), the lower bound in \eqref{eq_Huber_ineq} remains valid. Combined with Young’s inequality, this yields the $L^2(\Pbb^{\prime}_X)$-norm relation.
\[
\|f-f_{0,\tau}^{\prime}\|_{L^2(\Pbb_{X}^{\prime})}^2\leq 16 \left\{ \mathcal{R}_{\Pbb^{\prime}}(f) - \mathcal{R}_{\Pbb^{\prime}}(f_0^{\prime}) + 4\tau^{2-2r}v_r^2 \right\},
\]
which serves as the key step in translating the excess natural risk bounds from Theorem~\ref{thm_na} into an $L^2$ convergence rate relative to $f_{0,\tau}^{\prime}$.
  \begin{prop} \label{pro_HU1}
 	Suppose the conditions of Theorem \ref{thm_na} and Assumption \ref{con_H1} hold, and let $\tau \ge  2\max\{ 2 M_n,  (2v_r)^{1/r} \}$. For any $\Pbb^{\prime}$ satisfying $W_k(\Pbb^{\prime},\Pbb)\leq \delta$, assume that $\Pbb^{\prime}$ also satisfies $f_{0}^{\prime}\in\cH^{\alpha}$ and Assumption \ref{con_H1}(1).  If $k=1$, then
	\begin{align*}
 		\Ebb\big [\|\widehat{f}_{\delta,\tau}^{(n)}-  f_{0,\tau}^{\prime}\|_{L^2(\Pbb_{X}^{\prime})}^2 \big]\lesssim  \tau^{2-2r}v_r^2+\tau^2 n^{-\frac{ 2\beta}{d+2\beta}}\{\log n\}^2+ \delta  \tau K,
 	\end{align*}
 	where   $K \asymp \max\big\{n^{\frac{d+2m+1-2\alpha}{2(d+2\beta)}},\sqrt{\log n}\big\}$.
 	Further, if $k>1$, then
 \begin{align*}
 	\Ebb\big[ \|\widehat{f}_{\delta,\tau}^{(n)}-f_{0,\tau}^{\prime}\|_{L^2(\Pbb_{X}^{\prime})}^2\big] &\lesssim  \tau^{2-2r}v_r^2+ \tau^2  n^{-\frac{2\beta}{d+2\beta}} \{\log n\}^2 \\
 	&\qquad \;+ \delta  \tau \{1 + n^{-\frac{    \beta}{k^\star(d+2\beta)}} K \} + \delta^{\min\{k,2\}} \tau K^2.
 \end{align*}	
   \end{prop}
Proposition~\ref{pro_HU1} establishes the $L^2$ convergence rate relative to $f_{0,\tau}^{\prime}$ and explicitly captures the Huberization bias $\tau^{2-2r}v_r^2$, which arises under asymmetric noise. This bias remains constant for heavy-tailed noise  when $r=1$  and decays with $\tau$ for $r>1$. Since both the statistical and robustness-related error terms increase with $\tau$, its selection requires careful balancing to optimize overall performance.

	\begin{cor}
		Suppose the conditions of Proposition~\ref{pro_HU1} hold. Let $\delta = o(n^{- \frac{d + 2m + 1 + 2\beta}{2(d + 2\beta)}})$ for $k=1$ and $\delta  = o(n^{-\frac{c }{2(d+2\beta)}})$ for $k>1$, where $c=\max\{4\beta, 2k^{-1}(d+2m+1), 2k^{-1}\beta+d+2m+1\}$.   For any   $\Pbb^{\prime}$ satisfying $W_k(\Pbb^{\prime},\Pbb)\leq \delta$, assume that $\Pbb^{\prime}$ also satisfies $f_{0}^{\prime}\in\cH^{\alpha}$ and Assumption \ref{con_H1}(1). Then,
		\begin{align*}
		\Ebb\big[ \|\widehat{f}_{\delta,\tau}^{(n)}-f_{0,\tau}^{\prime}\|_{L^2(\Pbb_{X}^{\prime})}^2\big] &\lesssim  \tau^{2-2r}v_r^2+ \tau^2  n^{-\frac{2\beta}{d+2\beta}} \{\log n\}^2.
		\end{align*}	
	\end{cor}
}

\subsection{Nonparametric quantile regression}
 For a given quantile level $\varrho \in (0,1)$,   consider the quantile regression model
	\begin{equation}\label{eq_qun}
		Y = f_{\varrho}(X) + \varepsilon,
	\end{equation}
	where $f_{\varrho}$ is the true $\varrho$-th conditional quantile of $Y$ given $X$, and $\varepsilon$ satisfies $P(\varepsilon \le 0 \mid X) = \varrho.$  Quantile regression provides a detailed characterization of how $X$ affects the conditional distribution of $Y$, which is particularly useful in the presence of skewness or heteroscedasticity. Estimation of $f_{\varrho}$ is based on minimizing the natural risk $\cR_{\Pbb}(f) = \Ebb[\ell_\varrho(Y - f(X))]$, where $\ell_\varrho$ is the check loss function:
	\begin{align*}
		\ell_\varrho(u) = u\big\{\varrho-I(u\leq 0)\big\}.
	\end{align*}
	The loss $\ell_\varrho$ is convex and Lipschitz continuous with $\|\ell_\varrho\|_{\Lip}= \max\{1-\varrho, \varrho\}$, and $f_{\varrho}$ is the unique minimizer of  $\cR_{\Pbb}(f)$ under standard regularity conditions. The corresponding distributionally robust nonparametric estimator is defined as
	\begin{equation*}
		\widehat{f}_{\delta,\varrho}^{(n)} \in \argmin_{f\in \cF_n} \sup_{\Qbb: W_1(\Qbb,\Pbb_n)\leq \delta}\Ebb_{\Qbb} [ \ell_{\varrho}(Y-f(X))].
	\end{equation*}
	A key component of the analysis is the following assumption on $P_{Y|X=\bx}(\cdot)$, which denotes the conditional cumulative distribution function of $Y$ given $X=\bx$.
	\begin{assumption} \label{ass_calib}
 		There exist constants $L, \nu_1, \nu_2 > 0$ such that for all $\bx \in \cX$,
	\begin{align*}
 			\text{(1)} \quad \big| P_{Y|X=\bx}(f_{\varrho}(\bx)+u) - P_{Y|X=\bx}(f_{\varrho}(\bx)) \big| &\le \nu_1 |u|, \quad \text{for all } u \in \mathbb{R}, \\
 			\text{(2)} \quad \big| P_{Y|X=\bx}(f_{\varrho}(\bx)+u) - P_{Y|X=\bx}(f_{\varrho}(\bx)) \big| &\ge \nu_2 |u|, \quad \text{for all } |u| \le L.
 		\end{align*}
 	\end{assumption}
	
	Assumption~\ref{ass_calib} characterizes the properties of the conditional distribution around the true quantile. Condition~(1) imposes a global Lipschitz constraint on the conditional cumulative distribution function, guaranteeing that condition~(i) of Assumption~\ref{ass_Lipf} is satisfied with $c_1 = 0$ and $c_2 = \nu_1/2$. Condition~(2), often referred to as the self-calibration condition \cite{Shen2024JM}, requires the conditional density of $\varepsilon$ near zero to be bounded away from zero, ensuring localized strong convexity of the natural risk. In particular, it implies that condition~(ii) of Assumption~\ref{ass_Lipf} holds with $c_4 = 0$ and $c_3 = \min\{\nu_2/2, \nu_2 L/(8 M_n)\}$, which simplifies to $c_3 = \nu_2/2$ when $\|f-f_{\varrho}\|_\infty \le L$.

\change{\begin{prop} \label{pro_Q1}
		Suppose the conditions of Theorem \ref{thm_na} and Assumption \ref{ass_calib}  hold. For any   $\Pbb^{\prime}$  satisfying $W_1(\Pbb^{\prime},\Pbb)\leq \delta$, assume that $\Pbb^{\prime}$ also satisfies Assumption \ref{ass_calib}(2)  and  the corresponding quantile function $f_{\varrho}^{\prime}$
		satisfies $f_{\varrho}^{\prime}\in\cH^{\alpha}$. Then,
		\begin{align*}
		 	\Ebb\big[\|\widehat{f}_{\delta,\varrho}^{(n)}-f_{\varrho}^{\prime}\|_{L^2(\Pbb_{X}^{\prime})}^2  \big]\lesssim  n^{-\frac{ 2\beta}{d+2\beta}}\{\log n\}^3+ \delta K \log n,
		\end{align*}
			where   $K \asymp \max\big\{n^{\frac{d+2m+1-2\alpha}{2(d+2\beta)}},\sqrt{\log n}\big\}$. If further $\delta = o(n^{- \frac{d + 2m + 1 + 2\beta}{2(d + 2\beta)}})$, then
				\begin{align*}
				\Ebb\big[\|\widehat{f}_{\delta,\varrho}^{(n)}-f_{\varrho}^{\prime}\|_{L^2(\Pbb_{X}^{\prime})}^2  \big]\lesssim  n^{-\frac{ 2\beta}{d+2\beta}}\{\log n\}^3.
			\end{align*}			
	\end{prop}}


\subsection{Classification}
In a binary classification task, we seek to predict a label $Y \in \{-1, 1\}$ from covariates $X$. This task is typically evaluated using the classification loss  $\ell_{\operatorname{class}}(u,y)=1\big\{\operatorname{sign}(u)y\leq 0\big\},$ where  $u \in \mathbb{R}$ is a score and $\sign(u) =1$ if $u\geq 0$  and $-1$ otherwise. Given a score function $f: \mathbb{R}^d \mapsto \mathbb{R}$, the associated classifier is $\operatorname{sign} (f)$, and its natural classification risk is defined by
\[
\cR_{\operatorname{class},\Pbb}(f) = \Ebb_{\Pbb} \big[1\big\{\operatorname{sign} f(X)\ne Y\big\}\big].
\]
 The minimum natural classification risk, known as the Bayes risk, is
\begin{align*}
	\cR_{\operatorname{class},\Pbb}^{\star}= \inf_{f \text{ measurable}}\cR_{\operatorname{class},\Pbb}(f)
	=\Ebb\big[\min\{\eta(X),1-\eta (X)\}\big],
\end{align*}
where  $\eta(\bx)=P(Y=1| X=\bx)$ is the conditional class probability.  This infimum is reached by Bayes regression function  $f^{\star}(\bx)= 2\eta(\bx)-1$ \cite{svm2008}.

Since the classification loss is non-smooth and non-convex, its direct minimization is computationally intractable. This motivates the use of margin-based surrogate losses, which take the form $\ell(u,y)=\phi(uy)$ for some convex margin loss $\phi: \mathbb{R} \mapsto [0, \infty)$. A primary requirement for selecting $\phi$ is to ensure classification consistency \citep{svm2008}, meaning that minimizing the surrogate risk yields a classifier converging to the Bayes optimal classifier $\operatorname{sign} (f^{\star})$. Although margin-based methods have been extensively studied, standard estimators remain sensitive to label noise, which may arise from measurement error or other sources of uncertainty. This type of error is distinct from the covariate perturbations typically considered in adversarial risk. To address uncertainty in both $X$ and $Y$, we consider the  distributionally robust nonparametric estimator
\begin{equation} \label{eq_class}
	\widehat{f}_{k,\delta}^{(n)}\in\argmin_{f\in \cF_n} \sup_{\Qbb: W_k(\Qbb,\Pbb_n)\leq \delta}\Ebb_{\Qbb} [\phi(f(X)Y)].
\end{equation}

\change{
Although this multiplicative loss form $\phi(f(X)Y)$ differs from the additive structure in our main analysis, the framework can be extended. For simplicity, we focus on the quadratic surrogate $\phi(u)=(1-u)^2$, which connects the classification and least-squares frameworks. The corresponding risk satisfies $\cR_{\Pbb}(f) = \Ebb[(1-f(X)Y)^2] = \Ebb[(Y-f(X))^2],$ since $Y^2=1$. Thus, its population minimizer $f_0(\bx) = \Ebb[Y|X=\bx]$ is exactly the Bayes regression function $f^{\star}(\bx)$. This equivalence allows us to apply the $L^2 $ error bounds established in Theorem~\ref{thm_ls2}. For any $\Pbb^{\prime}$ satisfying $W_k(\Pbb^{\prime}, \Pbb)\leq \delta$, let $f_0^{\prime}(\bx)=\Ebb[Y^{\prime}|X^{\prime}=\bx]$ denote its corresponding Bayes regression function. Assuming $f_0^{\prime}\in\cH^{\alpha}$, the results of Theorem~\ref{thm_ls2} yield the following $L^2$ convergence rate.
\begin{align*}
	\Ebb\big[	\|\widehat{f}_{k,\delta}^{(n)}-f_0^{\prime}\|_{L^2(\Pbb^{\prime}_X)}^2\big] \lesssim n^{-\frac{2\beta}{d+2\beta}}\{ \log n \}^4.
\end{align*}
While this $L^2$ bound is a direct consequence of the previous analysis, the primary goal in classification is to bound the natural classification risk. The excess classification risk is related to the $L^2$ error via the following  inequality \cite{svm2008}.
\begin{align*}
	\cR_{\operatorname{class},\Pbb^{\prime}}(f) - \cR_{\operatorname{class},\Pbb^{\prime}}^{\star}
	\leq  \|f- f_0^{\prime}\|_{L^2(\Pbb_{X}^{\prime})}.
\end{align*}
Combining these results yields  the following result for the excess classification risk.
\begin{prop} \label{pro_Q3}
	Suppose the conditions of Theorem~\ref{thm_ls2} hold. For any $\Pbb^{\prime}$ satisfying $W_k(\Pbb^{\prime}, \Pbb)\leq \delta$, and assuming its Bayes regression function  satisfies $f_0^{\prime}\in\cH^{\alpha}$, we have
	\begin{align*}
		\Ebb\big[	\cR_{\operatorname{class},\Pbb^{\prime}}(\widehat{f}_{k,\delta}^{(n)}) -  \cR_{\operatorname{class},\Pbb^{\prime}}^{\star} \big]\lesssim n^{-\frac{\beta}{d+2\beta}}\{ \log n \}^2.
	\end{align*}
\end{prop}
	
 This proposition is particularly relevant for learning under label noise. Let $\Pbb_{\mathrm{true}}$ and $\Pbb_{\mathrm{obs}}$ denote the true and observed distributions, respectively, where $\Pbb_{\mathrm{obs}}$ coincides with $\Pbb_{\mathrm{true}}$ except that labels are independently flipped with probability $\xi\in(0,1)$. In this case, the Wasserstein distance between the two distributions is bounded by $W_k(\Pbb_{\mathrm{true}}, \Pbb_{\mathrm{obs}}) \leq 2\xi^{1/k}$. If the estimator $\widehat{f}_{k,\delta}^{(n)}$ is trained on samples from $\Pbb_{\mathrm{obs}}$  and we choose an uncertainty level $\delta \geq 2\xi^{1/k}$, then   $\Pbb_{\mathrm{true}}$ is ensured to be in the ambiguity set. Proposition \ref{pro_Q3} is thus directly applicable by setting $\Pbb^{\prime} = \Pbb_{\mathrm{true}}$, yielding a bound on the excess classification risk under the true data-generating distribution.

\begin{cor}
	Let $\Pbb_{\mathrm{true}}$ be the true distribution and $\Pbb_{\mathrm{obs}}$ be the observed version where labels are misclassified with probability $\xi\in(0,1)$. Let the estimator $\widehat{f}_{k,\delta}^{(n)}$ in \eqref{eq_class} be constructed from samples from $\Pbb_{\mathrm{obs}}$ with   $\delta \geq 2\xi^{1/k}$. Suppose the conditions of Proposition \ref{pro_Q3} hold. Then,
	\begin{align*}
	 	\Ebb\big[\cR_{\operatorname{class},\Pbb_{\mathrm{true}}}(\widehat{f}_{k,\delta}^{(n)}) -  \cR_{\operatorname{class},\Pbb_{\mathrm{true}}}^{\star}\big]\lesssim n^{-\frac{\beta}{d+2\beta}}\{ \log n \}^2.
	 \end{align*}
	\end{cor}
This result indicates that, despite being constructed from potentially misclassified data, the distributionally robust nonparametric estimator retains consistency under the true data-generating distribution, provided the uncertainty level $\delta$ is chosen to be larger than the noise level. This result is further illustrated in the numerical study in Section \ref{sec_num}.
}

\section{Numerical results}\label{sec_num}
In this section, we conduct numerical experiments to evaluate the finite-sample performance of the WDRO method. Its performance is compared to that of empirical risk minimization (ERM), which optimizes the natural risk, across both regression and classification tasks. To illustrate its practical utility, we also include an application to the MNIST dataset. Additional algorithmic details are provided in the Supplementary Material.

 \subsection{Simulation}
 For regression, we considered three loss functions: quadratic loss, Huber loss (with $\tau = 1$), and check loss (with $\rho = 0.5$). For classification, the cross-entropy loss was used. Model uncertainty was introduced through a shift probability, which controlled the likelihood of replacing a standard training sample with one drawn from a shifted distribution. When the shift probability was set to zero, the training data remained unchanged.

 Specifically, the covariate vector $X=(x_1,\dots, x_{10})^{\top}$ consisted of  five continuous variables sampled  independently from $U(0, 1)$ and five binary variables sampled independently from $\text{Bernoulli}(0.5)$. The response satisfied $Y = f_0(X) + \varepsilon$, where $\varepsilon \sim N(0, 0.5)$ and
 \[
 f_0(X) = 3x_1^2 - 2x_2 + 4x_3 x_7+ 2 \sin(2\pi x_4) + \cos(3\pi x_9)+  \exp(x_5  x_8)/10.
 \]
 For classification,  the response $Y$ was then converted to binary labels: samples with $Y > 0$ were labeled as class 1, and those with $Y \leq 0$ as class $-1$.

 To simulate model uncertainty, training samples were replaced according to a specified shift probability. In   regression, both $X$  and $Y$ were perturbed. The first five continuous variables of $X$ were modified by adding the vector $(0.3, -0.2, 0.25, -0.15, 0.2)^\top$, followed by clipping to  $[0, 1]$. Binary variables $x_6$ and $x_8$ were flipped with probability $0.5$. The response was then generated from a shifted function
 \[
 f_{\text{shifted}}(X) = x_1^2 +2x_3 x_7+ 2 \sin(2\pi x_4) + \cos(3\pi x_9)+  \exp(x_5  x_8)/10,
 \]
 with added noise $e \sim N(0, 0.6)$ such that $Y = f_{\text{shifted}}(X) + e$.  In   classification, the labels of shifted samples were reversed to simulate label noise.

 We experimented with shift probabilities in $\{0.0, 0.05, 0.1, 0.15, 0.2, 0.25\}$, using training sets of $1000$ samples. The WDRO uncertainty level was selected via cross-validation over   $\{0.0, 0.05, 0.1, 0.15, 0.2, 0.25\}$. An FNN with three hidden layers of sizes $32, 16,$ and $8$ was used. The output layer consisted of a single unit for regression or a binary classifier for classification. Norm constraints were applied to all linear layers to enforce a Lipschitz constant not exceeding 10.

 The methods were evaluated on both standard (unperturbed) and perturbed test sets, with the latter generated using twice the training shift probability. For classification task, an imbalanced test set was also considered (50\% of class 1 labels flipped). All test sets comprised of 500 samples, and each experiment was  repeated independently 20 times.

 \begin{table}[htbp]
 	\centering
 	\caption{Regression error comparison of WDRO and ERM methods on standard and perturbed test sets across different losses and shift probabilities (presented as mean $\pm$ standard deviation).}
 	\label{Tab_regress}
 	\begin{adjustbox}{width=0.82\textwidth}
 		\begin{tabular}{c|ccccc}
 			\toprule
 			\text{Loss} & \text{Test Set} & \text{Shift Probability} & \text{ERM} & \text{WDRO} & \text{Improvement} \\
 			\midrule
 			\multirow{12}{*}{\text{Quadratic}}
 			& \multirow{6}{*}{Standard}
 			& 0.00 & 1.088 $\pm$ 2.85\% & 0.973 $\pm$ 2.56\% & 9.721\% $\pm$ 2.93\% \\
 			& & 0.05 & 1.117 $\pm$ 3.29\% & 1.009 $\pm$ 2.42\% & 8.404\% $\pm$ 3.13\% \\
 			& & 0.10 & 1.149 $\pm$ 3.59\% & 1.068 $\pm$ 2.47\% & 5.480\% $\pm$ 3.43\% \\
 			& & 0.15 & 1.197 $\pm$ 3.50\% & 1.116 $\pm$ 2.57\% & 5.517\% $\pm$ 3.15\% \\
 			& & 0.20 & 1.237 $\pm$ 3.36\% & 1.148 $\pm$ 2.90\% & 5.986\% $\pm$ 3.31\% \\
 			& & 0.25 & 1.290 $\pm$ 3.12\% & 1.217 $\pm$ 2.63\% & 4.771\% $\pm$ 2.72\% \\
 			\cmidrule{2-6}
 			& \multirow{6}{*}{Perturbed}
 			& 0.00 & 1.099 $\pm$ 2.28\% & 0.982 $\pm$ 2.87\% & 10.05\% $\pm$ 3.05\% \\
 			& & 0.05 & 1.334 $\pm$ 2.23\% & 1.176 $\pm$ 2.89\% & 11.44\% $\pm$ 2.59\% \\
 			& & 0.10 & 1.446 $\pm$ 3.56\% & 1.284 $\pm$ 2.56\% & 10.35\% $\pm$ 2.67\% \\
 			& & 0.15 & 1.523 $\pm$ 3.46\% & 1.356 $\pm$ 3.00\% & 10.36\% $\pm$ 2.45\% \\
 			& & 0.20 & 1.506 $\pm$ 3.97\% & 1.345 $\pm$ 2.71\% & 9.76\% $\pm$ 2.55\% \\
 			& & 0.25 & 1.555 $\pm$ 4.87\% & 1.377 $\pm$ 3.32\% & 10.67\% $\pm$ 2.09\% \\
 			\midrule
 			\multirow{12}{*}{\text{Huber}}
 			& \multirow{6}{*}{Standard}
 			& 0.00 & 1.057 $\pm$ 3.60\% & 0.945 $\pm$ 2.44\% & 8.90\% $\pm$ 3.54\% \\
 			&  & 0.05 & 1.047 $\pm$ 3.37\% & 1.005 $\pm$ 2.46\% & 2.02\% $\pm$ 4.20\% \\
 			&  & 0.10 & 1.103 $\pm$ 3.94\% & 1.009 $\pm$ 2.38\% & 6.29\% $\pm$ 4.05\% \\
 			&  & 0.15 & 1.164 $\pm$ 3.72\% & 1.072 $\pm$ 2.92\% & 6.60\% $\pm$ 3.23\% \\
 			&  & 0.20 & 1.222 $\pm$ 3.82\% & 1.133 $\pm$ 3.05\% & 6.19\% $\pm$ 3.06\% \\
 			&  & 0.25 & 1.277 $\pm$ 3.76\% & 1.188 $\pm$ 2.95\% & 6.05\% $\pm$ 2.69\% \\
 			\cmidrule{2-6}
 			& \multirow{6}{*}{Perturbed}
 			& 0.00 & 1.060 $\pm$ 2.74\% & 0.946 $\pm$ 2.82\% & 9.81\% $\pm$ 3.31\% \\
 			&  & 0.05 & 1.276 $\pm$ 2.82\% & 1.175 $\pm$ 3.19\% & 6.92\% $\pm$ 3.44\% \\
 			&  & 0.10 & 1.426 $\pm$ 3.80\% & 1.255 $\pm$ 2.52\% & 10.91\% $\pm$ 2.81\% \\
 			&  & 0.15 & 1.494 $\pm$ 4.05\% & 1.313 $\pm$ 2.42\% & 11.16\% $\pm$ 2.39\% \\
 			&  & 0.20 & 1.492 $\pm$ 4.08\% & 1.313 $\pm$ 2.45\% & 11.17\% $\pm$ 2.11\% \\
 			&  & 0.25 & 1.490 $\pm$ 4.49\% & 1.333 $\pm$ 2.27\% & 9.36\% $\pm$ 2.61\% \\
 			\midrule
 			\multirow{12}{*}{\text{Check}}
 			& \multirow{6}{*}{Standard}
 			& 0.00 & 1.110 $\pm$ 3.66\% & 1.008 $\pm$ 2.62\% & 7.51\% $\pm$ 3.67\% \\
 			&  & 0.05 & 1.108 $\pm$ 3.81\% & 1.060 $\pm$ 2.47\% & 2.32\% $\pm$ 3.93\% \\
 			&  & 0.10 & 1.158 $\pm$ 3.84\% & 1.069 $\pm$ 2.78\% & 5.83\% $\pm$ 3.88\% \\
 			&  & 0.15 & 1.202 $\pm$ 3.40\% & 1.137 $\pm$ 2.79\% & 4.15\% $\pm$ 3.49\% \\
 			&  & 0.20 & 1.257 $\pm$ 3.93\% & 1.219 $\pm$ 3.30\% & 1.37\% $\pm$ 3.92\% \\
 			&  & 0.25 & 1.326 $\pm$ 4.47\% & 1.248 $\pm$ 3.00\% & 4.43\% $\pm$ 3.25\% \\
 			\cmidrule{2-6}
 			& \multirow{6}{*}{Perturbed}
 			& 0.00 & 1.121 $\pm$ 2.80\% & 1.006 $\pm$ 3.03\% & 9.13\% $\pm$ 3.77\% \\
 			&  & 0.05 & 1.341 $\pm$ 2.98\% & 1.221 $\pm$ 2.60\% & 8.00\% $\pm$ 2.98\% \\
 			&  & 0.10 & 1.489 $\pm$ 4.12\% & 1.319 $\pm$ 3.01\% & 10.29\% $\pm$ 2.90\% \\
 			&  & 0.15 & 1.547 $\pm$ 3.84\% & 1.375 $\pm$ 2.71\% & 10.40\% $\pm$ 2.24\% \\
 			&  & 0.20 & 1.549 $\pm$ 5.09\% & 1.394 $\pm$ 2.64\% & 8.42\% $\pm$ 3.02\% \\
 			&  & 0.25 & 1.501 $\pm$ 4.92\% & 1.385 $\pm$ 2.24\% & 6.48\% $\pm$ 2.33\% \\
 			\bottomrule
 		\end{tabular}
 	\end{adjustbox}
 \end{table}

 \begin{table}[htbp]
 	\centering
 	\caption{Classification accuracy comparison of WDRO and ERM methods on standard, perturbed, and imbalanced test sets across different shift probabilities (presented as mean $\pm$  standard deviation).}
 	\label{Tab_class}
 	\begin{adjustbox}{width=0.7\textwidth}	\begin{tabular}{ccccc}
 		\toprule
 		\text{Test Set} & \text{Shift Probability} & \text{ERM   (\%)} & \text{WDRO  (\%)} & \text{Improvement (\%)} \\
 		\midrule
 		\multirow{7}{*}{Standard } & 0.00& 77.2 $\pm$ 0.7& 86.8 $\pm$ 0.2& 9.6 $\pm$ 0.6 \\
 		& 0.05 & 78.4 $\pm$ 0.6 & 86.5 $\pm$ 0.3 & 8.1 $\pm$ 0.6 \\
 		& 0.10& 78.7 $\pm$ 0.5 & 85.9 $\pm$ 0.2 & 7.1 $\pm$ 0.5 \\
 		& 0.15 & 79.8 $\pm$ 0.5 & 85.2 $\pm$ 0.3 & 5.4 $\pm$ 0.6 \\
 		& 0.20 & 80.7 $\pm$ 0.7 & 84.3 $\pm$ 0.4 & 3.6 $\pm$ 0.7 \\
 		& 0.25 & 81.3 $\pm$ 0.6 & 83.0 $\pm$ 0.4 & 1.6 $\pm$ 0.7 \\
 		\midrule
 		\multirow{7}{*}{Perturbed }& 0.00 & 77.3 $\pm$ 0.5 & 86.6 $\pm$ 0.4 & 9.3 $\pm$ 0.5 \\
 		& 0.05 & 72.6 $\pm$ 0.7 & 78.3 $\pm$ 0.6 & 5.7 $\pm$ 0.5 \\
 		& 0.10 & 66.7 $\pm$ 0.7 & 71.0 $\pm$ 0.5 & 4.3 $\pm$ 0.5 \\
 		& 0.15 & 61.3 $\pm$ 0.5 & 63.6 $\pm$ 0.4 & 2.4 $\pm$ 0.3 \\
 		& 0.20& 55.9 $\pm$ 0.4 & 57.1 $\pm$ 0.4 & 1.3 $\pm$ 0.3 \\
 		& 0.25 & 49.2 $\pm$ 0.4 & 50.7 $\pm$ 0.5 & 1.5 $\pm$ 0.3 \\
 		\midrule
 		\multirow{7}{*}{Imbalanced} & 0.00 &  44.6 $\pm$ 0.7 &  61.3 $\pm$ 0.5 &  16.7 $\pm$ 0.7 \\
 		& 0.05 &  44.5 $\pm$ 0.6 &  59.7 $\pm$ 0.5 &  15.2 $\pm$ 0.7   \\
 		& 0.10 &  43.8 $\pm$ 0.7 &  58.0 $\pm$ 0.5 &  14.2 $\pm$ 0.6   \\
 		& 0.15 &  43.9 $\pm$ 0.7 & 56.3 $\pm$ 0.5 &  12.4 $\pm$ 0.6   \\
 		& 0.20 &  43.9 $\pm$ 1.0 &  54.5 $\pm$ 0.5 &  10.6 $\pm$ 0.8   \\
 		& 0.25 &  43.1 $\pm$ 1.0 &  52.9 $\pm$ 0.7 &  9.7 $\pm$ 0.9  \\
 		\bottomrule
 	\end{tabular}
 	 	\end{adjustbox}
 \end{table}

 Table~\ref{Tab_regress} shows that WDRO consistently achieves lower prediction errors than ERM across all regression settings, with more significant improvements under perturbed test sets, reaching approximately 10\%. This pattern holds across all three loss functions, indicating WDRO exhibits greater robustness under distributional shifts. Table~\ref{Tab_class} shows that WDRO consistently outperforms ERM across all shift probabilities and test settings in classification tasks. Although the performance gap decreases as the shift probability increases, WDRO still maintains a clear advantage, particularly in imbalanced and perturbed scenarios.

 \subsection{Application to MNIST dataset}

We evaluated the WDRO method on the MNIST dataset \citep{Lecun1998}, available through PyTorch's \texttt{torchvision} library. For computational efficiency, we subsampled $3000$ training and $600$ test samples from the original dataset. The network architecture followed a modified LeNet design, consisting of two convolutional layers (with $6$ and $16$ channels, $5 \times 5$ kernels, ReLU activations, and $2 \times 2$ max pooling) and three fully connected layers (with $120$, $84$, and $10$ units, respectively). The model outputs logits for $10$-class classification, and the cross-entropy loss is applied during training to optimize the model.

\begin{figure}[htbp]
 	\centering
 	
 	\begin{minipage}{\textwidth}
 		\centering
 		\includegraphics[width=0.9\textwidth, trim=0 50 0 55, clip]{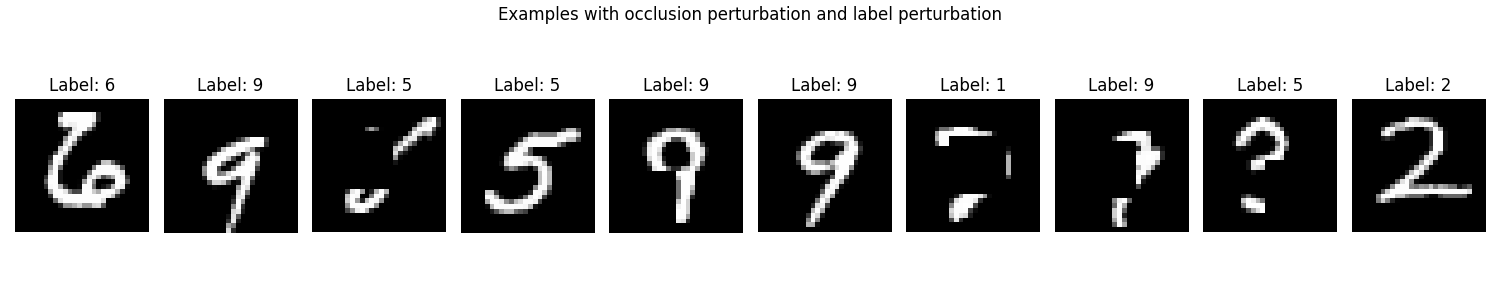} 
 		\subcaption{Occlusion  with similar label shift.} \label{fig:sub1}
 	\end{minipage}
 	\vspace{0.1cm}
 	
 	\begin{minipage}{\textwidth}
 		\centering
 		\includegraphics[width=0.9\textwidth, trim=0 50 0 55, clip]{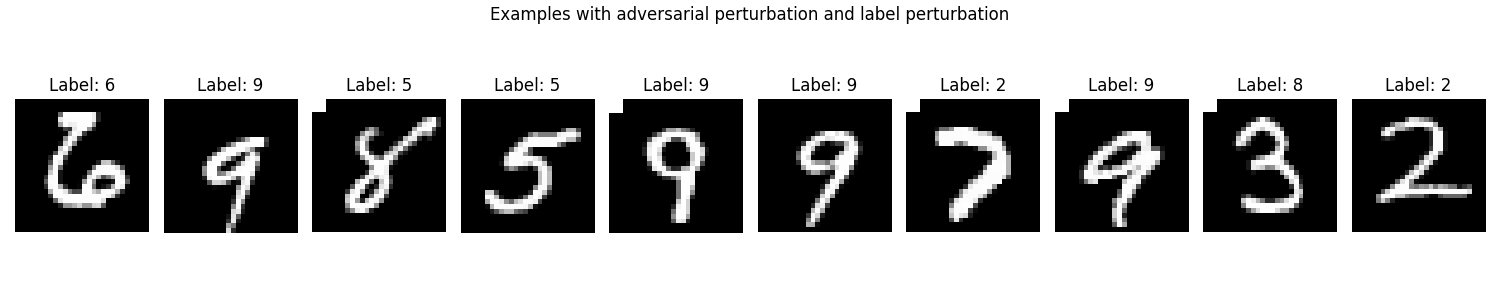} 
 		\subcaption{Corner corruption  with similar label shift.} \label{fig:sub2}
 	\end{minipage}
 	\vspace{0.1cm}
 	
 	\begin{minipage}{\textwidth}
 		\centering
 		\includegraphics[width=0.9\textwidth, trim=0 50 0 55, clip]{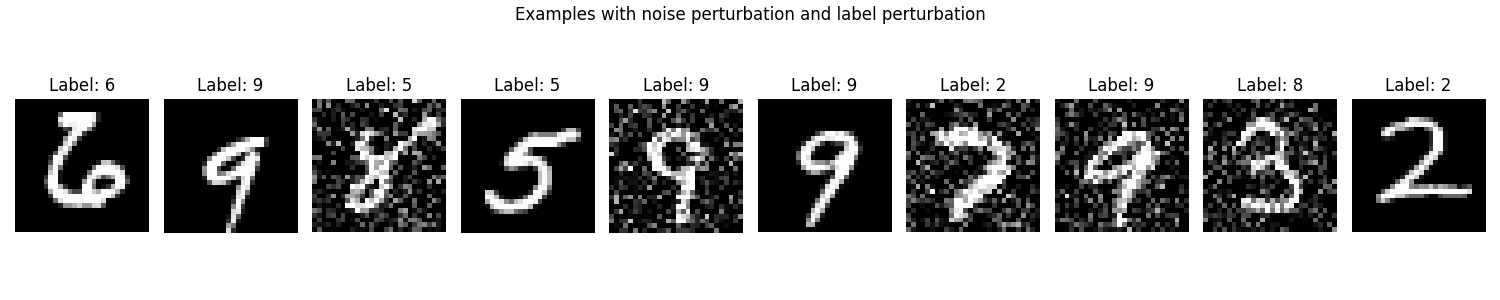} 
 		\subcaption{Pixel-wise noise   with similar label shift.} \label{fig:sub3}
 	\end{minipage}
 	\caption{Examples of the three types of perturbations applied during training and evaluation. }
 	\label{fig:overall_caption}
 \end{figure}

 To simulate real-world distortions, we applied three types of perturbations: (i) Occlusion: randomly masks a $12 \times 12$ region of the image by setting pixel values to zero, simulating localized information loss;
 (ii) Corner corruption: adds a fixed bright patch (pixel value 1) to the top-left corner, mimicking structured noise;
 (iii) Pixel-wise noise: adds Gaussian noise to each pixel with a noise level of $0.3$, clipping the values to the range $[0, 1]$.
 Additionally, all perturbations incorporated a label-shift strategy, replacing certain digits with visually similar alternatives based on common confusion patterns $(3 \leftrightarrow 5 \leftrightarrow 8,\ 1 \leftrightarrow 2 \leftrightarrow 7)$. Perturbations were applied to randomly selected samples with shift probabilities from $\{0.1, 0.2, 0.3\}$. The examples of the three types of perturbations are shown in Figure \ref{fig:overall_caption}.

 Both WDRO and ERM methods were trained on perturbed data and evaluated on both standard (unperturbed) and perturbed test sets, with the latter generated using twice the training shift probability. The WDRO uncertainty level was selected via cross-validation over $\{0.0, 0.1, 0.2, 0.3\}$. Each experiment was repeated independently 20 times.

 \begin{table}[htbp]
 	\centering
 	\caption{Classification accuracy comparison of  WDRO and ERM methods on standard and perturbed test sets across different perturbation types and shift probabilities (presented as mean $\pm$  standard deviation).}
 	\label{tab:perturbation_results}
 	\begin{adjustbox}{width=0.85\textwidth}	\begin{tabular}{cccccc}
 		\toprule
 		\multirow{2}{*}{Perturbation Type} & \multirow{2}{*}{Shift Probability} & \multicolumn{2}{c}{Standard (\%)} & \multicolumn{2}{c}{Perturbed (\%)} \\
 		\cmidrule(lr){3-4} \cmidrule(lr){5-6}
 		& & ERM  & WDRO & ERM  & WDRO \\
 		\midrule
 		\multirow{3}{*}{Occlusion}
 		& 0.1 & 95.70 $\pm$ 0.23 &96.24 $\pm$ 0.20 &84.90 $\pm$ 0.29&   85.37 $\pm$ 0.26\\
 		& 0.2 & 95.39 $\pm$ 0.26 &95.81 $\pm$ 0.21 &76.93 $\pm$ 0.25&   77.32 $\pm$ 0.33\\
 		& 0.3 & 94.63 $\pm$ 0.23 &95.30 $\pm$ 0.21 &  68.93 $\pm$ 0.48&   69.52 $\pm$ 0.45 \\
 		\midrule
 		\multirow{3}{*}{Corner corruption}
 		& 0.1 & 95.65 $\pm$  0.24 &96.52 $\pm$ 0.16&89.30 $\pm$ 0.29&  90.23 $\pm$ 0.27\\
 		& 0.2 & 95.74 $\pm$  0.23 &96.60 $\pm$ 0.16&83.78 $\pm$ 0.28& 85.02 $\pm$ 0.35\\
 		& 0.3 & 95.54 $\pm$  0.27&96.33 $\pm$ 0.17&78.46 $\pm$ 0.33& 79.21 $\pm$ 0.27\\
 		\midrule
 		\multirow{3}{*}{Pixel-wise noise}
 		& 0.1 & 95.75 $\pm$  0.19 &96.66 $\pm$  0.16&89.21 $\pm$  0.23& 89.86 $\pm$  0.26\\
 		& 0.2 & 95.81 $\pm$  0.19 &96.28 $\pm$  0.16&83.47 $\pm$  0.24& 84.18 $\pm$  0.21\\
 		& 0.3 & 95.36 $\pm$  0.26 &96.10 $\pm$  0.17&78.09 $\pm$  0.24& 77.78 $\pm$  0.25 \\
 		\bottomrule
 	\end{tabular}
 	 	 	\end{adjustbox}
 \end{table}

 Table~\ref{tab:perturbation_results} shows that WDRO consistently outperforms ERM under both standard and perturbed test sets across nearly all perturbation types and shift probabilities.
 In particular, WDRO exhibits enhanced robustness in the corner corruption setting, with an average performance gain exceeding 0.9\%. In contrast, the improvement under occlusion perturbation is more modest, averaging around 0.5\%.

 \section{Discussion}\label{sec_dis}
\change{In this paper, we develop a general analytical framework for studying the generalization performance of estimators derived from the local worst-case risk under model misspecification. We show that the local worst-case risk is closely connected to the regularized natural risk, which implicitly enforces gradient regularization, and we provide explicit bounds for this relationship when different $W_k$  are considered. The results for the $k=1$ case further illustrate the distinction between the local worst-case risk and the adversarial risk, with the former mainly influenced by points where the Lipschitz constant is attained and the latter shaped by local behavior.  We propose a distributionally robust nonparametric estimator by minimizing the local worst-case risk within a norm-constrained FNN space. The norm constraint plays an important role by providing both a Lipschitz upper bound and a uniform bound on the Hessian norm. By jointly analyzing the function and its gradient, we establish bounds on the space’s complexity and approximation properties. This fundamental property is particularly important for the analysis in the $k>1$ case. The main theoretical result establishes non-asymptotic error bounds for the excess local worst-case risk. With an appropriate choice of $\delta$, the estimator achieves the minimax optimal rate when $\alpha<(d+2m+1)/2$. The bounds also precisely characterize the dependence of the robustness-related term, showing a linear structure for $k=1$ and a higher-order dependence for $k>1$. The assumptions for this analysis are verified in typical nonparametric regression and classification problems. Finally, numerical comparisons in simulations and applications to the MNIST dataset demonstrate the superior robustness of the proposed estimator compared to its natural risk-based counterpart.

\mainchange{Several important problems remain for further study. Developing higher-order approximation theories for norm-constrained FNNs will deepen the understanding of their expressive power. It is also of interest to investigate ambiguity balls tailored to specific application settings to enhance practical relevance and performance.}
}


\section{Supplementary Material}
The Supplementary Material is organized as follows. Section~\ref{sec_A} introduces auxiliary lemmas, with their proofs provided in Section~\ref{sec_prof_aux}. Section~\ref{sec_B} presents the proofs for the lemmas from the main text, and Section~\ref{sec_C} contains the proofs for the theorems.  For clarity, the original statements of these results are restated before their respective proofs. Finally, Section~\ref{sec_algorithm} describes the algorithm used in the numerical study presented in Section~6 of the main text.
\\~\\
{\bf Notations.}
We first outline some notations to be used in the sequel. Let the set of positive integers be denoted by $\mathbb{N}= \{1, 2, . . .\}$ and let  $\mathbb{N}_0=\mathbb{N}\cup \{0\}.$ For a integer $d$, let $[d]=\{1,\dots, d\}.$ For any interger $q\in \mathbb{N}$, let $q^{\star}$ be the integer such that $1/q+1/q^{\star}=1$.
 If $a$ and $b$ are two quantities, we use $a\lesssim b $ or $b\gtrsim a$ to denote the statement that $a\leq Cb$ for some constant $C>0$. We denote $a\asymp b$ when $a\lesssim b\lesssim a$. Let $\lceil a \rceil$ denote the smallest integer larger than or equal to quantity $a$. For a vector $\bx$ and $p\in[1,\infty]$, we use $\|\bx \|_{p}$ to denote the $p$-norm of $\bx$. For a function $f$, we denote its supremum norm by $\|f \|_{\infty}$ and its gradient by $\nabla f$.  For integer $p>0$,   define $\|f\|_p=\{\Ebb[|f(X)|^p]\}^{1/p}$ and $\|f\|_{p,n}=\{\frac{1}{n}\sum_{i=1}^n|f(X_i)|^p\}^{1/p}$.
When it is necessary to emphasize the distribution of $X$, we write  $\|f\|_{L^p(\Pbb)}  =  \{\Ebb_{\Pbb}[|f(X)|^p] \}^{1/p}.$
 For a function $f$ defined on a set $\Omega$, its $L^{\infty}$ norm  is defined as
$\|f\|_{L^{\infty}(\Omega)} = \inf\{C \ge 0 : |f(x)| \le C \ \text{for almost all } x \in \Omega\}.$
For a measurable function $H$, let $\esssup(H)$ denote  its essential supremum.

\begin{appendix}

 \section{Auxiliary lemmas}\label{sec_A}

 The following result follows directly from Example 5.8 of \cite{Wainwright_2019}.
 \begin{lemma}\label{lem_covercall}
 	Let $\|\cdot\|$ be a norm on $\Rbb^d$, and denote by $\mathbb{B}^d = \{\bx \in \Rbb^d : \|\bx\| \leq 1\}$ the corresponding unit ball. Then,
 	\[
 	(1/u)^d \leq \cN(u, \mathbb{B}^d ,\|\cdot\|)\leq ( 2/u +1)^d.
 	\]
 \end{lemma}

 The following lemma provides an upper bound for the covering number of a class of bounded matrices, with the proof deferred to Section \ref{sec_prof_aux}.
 \begin{lemma}\label{lem_covermatrix}
 	Define the set
 	\begin{equation*}
 		\mathcal{M}(m,n,r)=\Big\{ \bW\in \Rbb^{m\times n}:\|\bW\|_{\infty}\leq r\Big\},
 	\end{equation*}
 	where $\|\cdot\|_{\infty}$  denotes  the infinity matrix norm  satisfying $\|\bW\|_{\infty}=\sup_{\|\bx\|_{\infty}\leq 1}\|\bW\bx\|_{\infty}$.
 	The $u$-covering number  associated with the  infinity matrix norm for the class  $\mathcal{M}(m,n,r)$ satisfies
 	\[
 	\mathcal{N}\left(u, \mathcal{M}(m,n,r),\|\cdot\|_{\infty}\right)\leq (2r/u+1)^{mn}.
 	\]
 \end{lemma}

 The following lemma shows   any $g \in \cN\cN(W,L,\mathcal{S})$ can be reparameterized so that each weight matrix in the hidden layers has norm at most one, with    proof deferred to Section \ref{sec_prof_aux}.

 \begin{lemma}\label{lem_normalize}
 	Fix $m\in\mathbb{N}$ and $q\in \mathbb{N}\cup\{\infty\}$. For every $g \in \cN\cN(W,L,\mathcal{S})$, there exists a reparameterization
 	\[
 	g(\bx) = \tilde{g}_L \circ \tilde{g}_{L-1} \circ \cdots \circ \tilde{g}_0(\bx),
 	\]
 	where $\tilde{g}_\ell(\bx) = \sigma_m(\tilde{\bA}_\ell \bx + \tilde{\bb}_\ell)$ for $0\leq\ell\leq L-1$, and $\tilde{g}_L(\bx) = \tilde{\bA}_L \bx$. The parameters satisfy $\|(\tilde{\bA}_\ell, \tilde{\bb}_\ell)\|_q \le 1$ for $0\leq\ell\leq L-1$, and
 	$\tilde{\bA}_L = \bA_L ( \prod_{s=0}^{L-1} r_s^{m^{L-s}}),$ with $ r_s = \max\{\|(\bA_s, \bb_s)\|_q, 1\}.$
 \end{lemma}

 \begin{lemma}  \label{lem_gradient}
 	Let $g \in \cN\cN(W,L,\mathcal{S})$ be an FNN with RePU activation $\sigma_m$ for $m\in\mathbb{N}$, and assume the hidden-layer weights and biases are normalized so that $\|(\bA_\ell, \bb_\ell)\|_\infty \le 1$ for $0 \le \ell \le L-1$. Let $\cX =\{\bx\in\Rbb^d:\|\bx\|_{\infty}\leq 1\}$. Then, the gradient and Hessian of $g$ satisfy
 	\begin{align*}
 		\sup_{\bx\in\cX}	\|\nabla g(\bx)\|_1 \le m^L \|\bA_L\|_\infty\quad\text{ and }\quad
 		\sup_{\bx\in\cX}	\big\|\nabla^2 g(\bx) \big\|_{\infty} \leq   Lm^{2L}\|\bA_L\|_{\infty}.
 	\end{align*}
 \end{lemma}

 Define two classes of functions over the unit ball $\mathbb{B}^d  = \{ \bx \in \mathbb{R}^d : \|\bx\|_2 \le 1 \}$ as follows.
 \begin{equation*} \label{def_shallowFNN}
 	\cF_{\sigma_m}(W,K) := \Bigl\{ f(\bx) = \sum_{i=1}^W a_i \, \sigma_m((\bx^{\top},1)\bv_i) : \bv_i \in \mathbb{S}^d,\; \sum_{i=1}^W |a_i| \le K \Bigr\},
 \end{equation*}
 \begin{equation*}
 	\cF_{\sigma_m}(K) := \left\{ f(\bx) = \int_{\mathbb{S}^d} \sigma_m((\bx^{\top},1)\bv) d\mu(\bv): \|\mu\| \le K \right\},
 \end{equation*}
 where $\|\mu\|$ denotes the total variation and $\mathbb{S}^d=\{ \bx \in \mathbb{R}^{d+1} : \|\bx\|_2 = 1 \}$.  The following lemma, based on Theorems 2 and 3 of \cite{siegel2023optimal}, establishes that $\cF_{\sigma_m}(K)$ admits a simultaneous approximation in both function values and derivatives by $\cF_{\sigma_m}(W,K)$.
 \begin{lemma} \label{lem_shallow_approximation}
 	Let    $0 \leq r \leq m$. Then there exists a constant
 	$C = C(d,m) > 0$ such that  for any $W\in\mathbb{N}$, there holds
 	\[
 	\sup_{f \in \mathcal{F}_{\sigma_m}(1)} \;
 	\inf_{g \in \mathcal{F}_{\sigma_m}(W,1)}
 	\Big\{ \;
 	\sup_{\|\bs\|_1 \leq r}
 	\big\| D^{\bs}(f - g) \big\|_{L^\infty(\mathbb{B}^d)}
 	\;\Big\}
 	\leq  C(d,m) W^{-\frac{1}{2} - \frac{2(m-r)+1}{2d}} .
 	\]
 \end{lemma}

 The next lemma shows properties of the operator $S_m$, introduced in \cite{bach2017breaking,yang2024optimal}, which plays an important role in the proof of the approximation result stated in Theorem 3.4 of the main text.  The detailed proof of the lemma can be found in Section \ref{sec_prof_aux}.
 \begin{lemma}\label{lem_Sk_property}
 	Let $h\in C^1(\mathbb{S}^d)$ and define the operator $S_m: C^1(\mathbb{S}^d)\mapsto C^1(\mathbb{B}^d)$ as
 	\[
 	S_m(h)(\bx) = (1+\|\bx\|_2^2)^{m/2}   h\big(\bu(\bx)\big), \quad  \text{with }\;
 	\bu(\bx) = \frac{1}{\sqrt{1+\|\bx\|_2^2}} \begin{pmatrix} \bx \\ 1 \end{pmatrix} \;\text{ for }\; \bx \in \mathbb{B}^d.
 	\]
 	Then
 	$\|S_m (h)\|_{L^{\infty}(\mathbb{B}^d)}\leq 2^{m/2}\|h\|_{L^{\infty}(\mathbb{S}^d)}$.  For $1\leq i\ne j\leq d+1$, define the angular derivative
 	$D_{i,j} =u_i\partial_{u_j}-u_j\partial_{u_i}.$ There exists a constant $C>0$, depending only on $m$ and $d$, such that
 	\begin{align*}
 		\sup_{\bx\in\mathbb{B}^d}	\|	\nabla_{\bx} S_m(h)(\bx)\|_{\infty}\leq
 		C\;\Big\{ \|h \|_{L^{\infty}(\mathbb{S}^d)}+\sum_{1\le i<j\le d+1} \|D_{i,j}h\|_{L^{\infty}(\mathbb{S}^d)}\Big\}.
 	\end{align*}
 \end{lemma}

 Next, we introduce several fundamental tools from empirical process theory to support the main analysis. Specifically,  for a given function class $\cF$ and a radius  $r>0$,   the localized empirical Rademacher complexities under $\|\cdot\|_2$ and $\|\cdot\|_{2,n}$ are respectively defined as
 \[
 \overline{\cR}_n(r;\cF): =\Ebb_{\bm{\sigma}}\Big[\sup_{\substack{g \in \mathcal{F} \\ \|g\|_{2} \leq r}}\Big|\frac{1}{n}\sum_{i=1}^n\sigma_ig(\bx_i)\Big|\Big],\;\quad \; \widehat{\cR}_n(r;\cF): =\Ebb_{\bm{\sigma}}\Big[\sup_{\substack{g \in \mathcal{F} \\ \|g\|_{2,n} \leq r}}\Big|\frac{1}{n}\sum_{i=1}^n\sigma_ig(\bx_i)\Big|\Big],
 \]
 where $\boldsymbol{\sigma} = (\sigma_1,\dots, \sigma_n)^{\top}$ denotes a vector of independent Rademacher variables, independent of the samples $\{\bx_i\}_{i=1}^n$.
 If $\mathcal{F}$ is star-shaped, i.e., $a g \in \mathcal{F}$ for every $g \in \mathcal{F}$ and $a \in [0,1]$, then the ratio $\overline{\mathcal{R}}_n(r;\mathcal{F}) / r$ is non-increasing in $r$. Specifically, this property is stated in the following lemma, whose proof is deferred to Section~\ref{sec_prof_aux}.

 \begin{lemma}\label{lem_star}
 	Suppose $\cF$ is a star-shaped function class. Then, for any $0 < r_1 \leq r_2$,
 	\begin{align*}
 		\frac{1}{r_2}\overline{\cR}_n(r_2;\cF)
 		\leq \frac{1}{r_1}\overline{\cR}_n(r_1;\cF).
 	\end{align*}
 \end{lemma}


For completeness, we restate several key inequalities from empirical process theory.
 \begin{lemma}[Dudley's integral covering number bound  \cite{srebro2010note}]\label{lem_cover}
 	 Suppose the function class $\cF$  satisfies $\sup_{f\in\cF}\|f\|_{2,n}\leq B$ for some positive constant $B$. Then,
 	\begin{equation*}
 		\begin{split}
 			\Ebb_{\boldsymbol{\sigma}}\Big\{\sup_{f\in\cF }  \frac{1}{n}\sum_{i=1}^n\sigma_if( X_i)  \Big\}
 			\leq \inf_{\eta\geq 0}\Big\{
 			4\eta+12\int_{\eta}^{B}\sqrt{\frac{\log\mathcal{N}(u,\cF,L^2(\Pbb_n))}{n}}du
 			\Big\},
 		\end{split}
 	\end{equation*}
 	where    $\boldsymbol{\sigma} = (\sigma_1,\dots, \sigma_n)^{\top}$ denotes a vector of independent Rademacher variables, independent of the samples.
 \end{lemma}

 \begin{lemma}[Ledoux–Talagrand contraction inequality \cite{Ledoux1991}]\label{lem_Contraction}
 	Let $F: \mathbb{R} \to \mathbb{R}$ be a convex and increasing function.
 	For each $i \in [n]$, let $\varphi_i: \mathbb{R} \to \mathbb{R}$ be a function with Lipschitz constant $1$ and $\varphi_i(0)=0$.  Then, for any bounded subset $T \subseteq \mathbb{R}^n$,
 	\[
 	\mathbb{E} F\left(\frac{1}{2}
 	\sup_{(t_1,\dots, t_n)\in T}\left|\sum_{i=1}^n \sigma_i \varphi_i\left(t_i\right)\right|\right) \leq \mathbb{E} F\left(\sup_{(t_1,\dots, t_n)\in T}\left|\sum_{i=1}^n\sigma_i t_i\right|\right).
 	\]
 \end{lemma}


 \begin{lemma}[Talagrand concentration for empirical processes \cite{Wainwright_2019}]\label{lemma_talagrand}
 	Let $\mathcal{F}$ be a countable class of functions satisfying $\sup_{f\in\mathcal{F}}\|f\|_\infty \le B$ for some constant $B$. Define
 	\[
 	Z=\sup_{f\in\cF}\left\{ \frac{1}{n}\sum_{i=1}^nf(X_i)\right\}~~\text{ or }~~ Z=\sup_{f\in\cF}\left| \frac{1}{n}\sum_{i=1}^nf(X_i)\right|.
 	\]
 	Then, for any $\delta>0$, it holds that
 	\[
 	P\Big\{Z \geq \mathbb{E}[Z]+\delta\Big\}\leq 2 \exp \left(\frac{-n \delta^2}{8 e \mathbb{E}\left[\Sigma^2\right]+4 B \delta}\right).
 	\]
 	Here, $\Sigma^2=\sup _{f \in \mathcal{F}} \frac{1}{n} \sum_{i=1}^n f^2\left(X_i\right)$, and its expectation satisfies $\Ebb[\Sigma^2]\leq \sigma^2+2B\Ebb[Z]$, where $\sigma^2=\sup _{f \in \mathcal{F}} \Ebb[f^2\left(X\right)]$.
 \end{lemma}

 \begin{lemma}[McDiarmid’s inequality]\label{lem_MC}
 	Let $f: \mathcal{X}^n \rightarrow \mathbb{R}$ satisfy the bounded differences property with constants $c_i$, i.e., for each $i \in [n]$, all $(x_1,\dots, x_n) \in \mathcal{X}^n$, and any $x_i^{\prime} \in \mathcal{X}$,
 	\[
 	\left| f\left(x_1, \dots, x_{i-1}, x_i, x_{i+1}, \dots, x_n\right)
 	- f\left(x_1, \dots, x_{i-1}, x_i^{\prime}, x_{i+1}, \dots, x_n\right) \right| \leq c_i.
 	\]
 	Let $X_1,\dots, X_n$ be independent random variables. Then, for any $t>0$,
 	\[
 	P\Big\{\Big|f\left(X_1,\dots, X_n\right)-\mathbb{E}\left[f\left(X_1,\dots, X_n\right)\right]\Big| \geq t\Big\} \leq 2\exp\Big\{-\frac{2 t^2}{\sum_{i=1}^n c_i^2}\Big\}.
 	\]
 \end{lemma}

 The next lemma is instrumental in establishing the convergence rates of our main results. Its proof, guided by the localization techniques in Chapter~14 of \cite{Wainwright_2019}, is deferred to Section~\ref{sec_prof_aux}.
 \begin{lemma}
 	\label{lemma_conrate}
 	Let  $\cF_n$ be a  class of functions satisfying $\sup_{f\in\cF_n}\|f\|_{\infty}\leq M_n$, 	where $M_n$ is a positive constant that may depend on $n$,  and define its star-shaped shifted   class as
 	$ \cF_n^*:= \{a(f-f_0): a\in[0,1], f\in\cF_n \},$ where the reference function $f_0$ also satisfies $\|f_0\|_{\infty}\leq M_n$. Let $r_n$  be defined such that $ \Ebb[ \overline{R}_n(r_n;\cF_n^*) ]= r_n^2/(64M_n).$ For any $f \in \cF_n$, let $H_n(f;f_0):=\Ebb[\ell(Z;f)-\ell(Z;f_0)]-\frac{1}{n}\sum_{i=1}^n\{  \ell(Z_i;f)-\ell(Z_i;f_0)\}$, where $\ell(\bz;f)=\ell(f(\bx)-y)$ with $\bz=(\bx,y)$. Then there exist positive constants
 	$b_1, b_2$ such that, with probability at least
 	$1-2\exp\{- \frac{nb_1 r_n^2}{2M_n^2}+b_2\log(\log(M_nr_n^{-1}))\}$, we have
 	\[
 	\big|H_n(f;f_0)\big| \leq  2 \|\ell\|_{\Lip} M_n^{-1}r_n(\|f-f_0\|_2+r_n)~~~~\text{ for any $f\in\cF_n$}.
 	\]
 \end{lemma}

 We note that Lemma \ref{lemma_conrate} extends beyond the specific loss $\ell(f(\bx)-y)$. The result holds for any loss of the form $\ell(\bz;f)=\phi(f(\bx),y)$ that is $L$-Lipschitz with respect to its function argument, i.e.,
 \[
 |\ell(\bz;f)-\ell(\bz;\tilde{f})|\leq L|f(\bx)-\tilde{f}(\bx)| \quad \text{for some } L > 0.
 \]
  In such a generalization, the term $\|\ell\|_{\Lip}$ in the lemma's statement would   be replaced by $L$. However, for consistency with the primary setting of the main text, we present Lemma \ref{lemma_conrate} in its current, more specific form.


 As analyzing the localized empirical Rademacher complexity $\widehat{\cR}_n(r;\cF_n^*)$ is often more tractable than   $\mathbb{E}[\overline{\cR}_n(r;\cF_n^*)]$, we establish a connection between the two in the following lemma. The proof, guided by \cite{Wainwright_2019}, is deferred to Section~\ref{sec_prof_aux}. Together with Theorem~3.3 in the main text, this lemma further implies Lemma~\ref{lem_rnn}.

 \begin{lemma}\label{lem_rr}
 	Suppose $\cF_n^*$ is a star-shaped function class satisfying $\sup_{g \in \cF_n^*}\|g\|_{\infty} \le 2M_n$,
 	where $M_n$ is a positive constant that may depend on $n$.
 	Let $\hat{r}_n$ and $r_n$ be positive numbers such that $ \widehat{\cR}_n(\hat{r}_n;\cF_n^*) \leq c\hat{r}_n^2$ for a positive constant $c$, and $
 	\Ebb[ \overline{R}_n(r_n;\cF_n^*) ]=  r_n^2 /(64M_n)$. Assume $r_n/M_n=o(1)$. Then there exist positive constants $s_1, s_2$ such that
 	\[
 	\hat{r}_n\geq s_1r_n/(cM_n)
 	\]
 	with probability at least $1- \exp\{-\frac{n r_n^4}{s_2M_n^4}\} $.
 \end{lemma}

 \begin{lemma}\label{lem_rnn}
 	Let $\cF_n=\{T_{M_n}g:g\in\cN\cN(W,L,\mathcal{S},K)\}$ and $\cF_n^*=\{a(f-f_0):a\in[0,1], f\in\cF_n\}$, where $M_n$ is a positive constant that may depend on $n$. Assume that the reference function $f_0$ satisfies $\|f_0\|_{\infty} \le M_n$,
 	and that  $M_n\lesssim K $ and $ \mathcal{S}\log(Kn)/n=o(1)$. For $r_n$  satisfying  $
 	\Ebb[ \overline{R}_n(r_n;\cF_n^*) ]=  r_n^2 /(64M_n)$,   there exists a positive constant $c$ such that
 	\[
 	r_n \leq  c M_n n^{-1/2}\sqrt{\mathcal{S}\log(Kn)}.
 	\]
 \end{lemma}

 \begin{lemma}\label{lem_ab}
 	For any $a,b\ge 0$ and $r\in(0,1]$, we have
 	$|a^{r}-b^{r}| \;\le\; |a-b|^{r}.$
 \end{lemma}

\section{Proof of lemmas}\label{sec_B}
\phantomsection
\begin{statement} {Lemma 2.3}
	Suppose the composed loss $\ell(\cdot;f)$ has a bounded Lipschitz constant $\|\ell_{f}\|_{\Lip}<\infty$, then
	\begin{align*}
		\cR_{\Pbb,1}(f;\delta)-\cR_{\Pbb}(f) 	\leq \delta \|\ell_{f}\|_{\Lip}.
	\end{align*}
	Furthermore, there holds
	\begin{align*}
		\cR_{\Pbb,1}(f;\delta)-\cR_{\Pbb}(f)=\delta  \|\ell_{f}\|_{\Lip},
	\end{align*}
	if one of the following conditions is satisfied:
	\begin{itemize}
		\item[(i)] There exists $\bz_0\in\cZ$ such that
		\[
		{\lim\sup}_{\|\bz-\bz_0\|\rightarrow\infty}\frac{\ell(\bz;f)-\ell(\bz_0;f)}{\|\bz-\bz_0\|}= \|\ell_{f}\|_{\Lip}.
		\]
		\item[(ii)] There exists $\bz_0\in\cZ$, a positive constant $\tau$, and a sequence $\{r_m\}_{m\geq 1}\rightarrow 0$ such that
		\[
		{\sup}_{\|\bz-\bz_0\|\geq \tau}\frac{\ell(\bz;f)-\ell(\bz_0;f)}{\|\bz-\bz_0\|}= \|\ell_{f}\|_{\Lip}
		\]
		and $P(\|Z-\bz_0\|\leq r_m)>0$ for any $m\in\mathbb{N}$.
	\end{itemize}
\end{statement}
	\addcontentsline{toc}{subsection}{Lemma 2.3}
\begin{proof}[{\bf Proof of Lemma 2.3}]
     From the definition of the Wasserstein distance, the worst-case risk $\cR_{\Pbb,1}(f;\delta)$ can be reformulated as
	\begin{align*}
	\cR_{\Pbb,1}(f;\delta)  =\sup_{\tilde{Z}\in\cP_1(\cZ):\Ebb[\|\tilde{Z}-Z\|]\leq \delta}\Ebb[\ell(\tilde{Z};f)]
	=\sup_{\Delta\in\cP_1(\cZ):\Ebb[\|\Delta\|]\leq \delta}\Ebb[\ell(Z+\Delta;f)].
	\end{align*}
	It directly follows an upper bound of $\cR_{\Pbb,1}(f;\delta)-\cR_{\Pbb,1}(f;0)$ based on the definition of the Lipschitz constant.
	\begin{align*}
	\cR_{\Pbb,1}(f;\delta)-\cR_{\Pbb,1}(f;0)&=\sup_{\Ebb[\|\Delta\|]\leq \delta} \Ebb\big[\ell(Z+\Delta;f)-\ell(Z;f)\big]
\\&	\leq \sup_{\Ebb[\|\Delta\|]\leq \delta} \Ebb\big[ \|\ell_f\|_{\Lip}\|\Delta\|\big]
	\leq \delta \|\ell_{f}\|_{\Lip}.
	\end{align*}
	
	Moreover, under the  conditions (i) or (ii), there exists $\bz_0$ and a sequence $\{\bz_m\}_{m\geq 1}\subseteq \cZ$ such that
	\[
	\frac{\ell(\bz_m;f)-\ell(\bz_0;f)}{\|\bz_m-\bz_0\|}\geq \|\ell_{f}\|_{\Lip}-m^{-1}.
	\]
	Define $\mathbb{F}_{m}(\bz_0)=\{\bz\in\cZ:\|\bz-\bz_0\|\leq r_m\}$ for a positive nonincreasing sequence $\{r_m\}_{m\geq 1}$. It is required  that $P(Z\in \mathbb{F}_m(\bz_0))>0$ and  $\{r_m\}_{m\geq 1}$ will be specified subsequently.
	For any $\bz\in\mathbb{F}_{m}(\bz_0), $ we have
	\begin{equation}\label{eq_zmlow}
	\begin{split}
	\ell(\bz_m;f)-\ell(\bz;f)&\geq  \ell(\bz_m;f)-\ell(\bz_0;f)-\big| \ell(\bz_0;f)-\ell(\bz;f)\big|
	\\&\geq  \big\{\|\ell_{f}\|_{\Lip}-m^{-1}\big\}\big\|\bz_m-\bz_0\big\|- \|\ell_{f}\|_{\Lip}\big\|\bz-\bz_0\big\|
	\\&\geq  \big\{\|\ell_{f}\|_{\Lip}-m^{-1}\big\}\big\|\bz_m-\bz\big\|- 2\|\ell_{f}\|_{\Lip}\big\|\bz-\bz_0\big\|
	\\&\geq  \big\{\|\ell_{f}\|_{\Lip}-m^{-1}\big\}\big\|\bz_m-\bz\big\|- 2\|\ell_{f}\|_{\Lip}r_m.
	\end{split}
	\end{equation}
	Consider a sequence of   random vectors defined by
	\[
	Z_m=\{(1-\pi_m)Z+\pi_m \bz_m\}1\{Z\in\mathbb{F}_m(\bz_0)\}+Z1\{Z\notin\mathbb{F}_m(\bz_0)\},
	\]
	where $\pi_m\in\{0,1\}$ is independent of $Z$ satisfying $P(\pi_m=1)=\varepsilon_m$ for some $\varepsilon_m\in(0,1)$.
	Then, we have
	\begin{align*}
	\Ebb\big[\|Z_m-Z\|\big]&= \Ebb\big[\|Z_m-Z\|\mid \pi_m=1 \big]P(\pi_m=1)+\Ebb\big[\|Z_m-Z\|\mid \pi_m=0 \big]P(\pi_m=0)
	\\&=\varepsilon_m \Ebb\big[\| \bz_m 1\{Z\in\mathbb{F}_m(\bz_0)\}+Z1\{Z\notin\mathbb{F}_m(\bz_0)\}-Z\|\big]
	\\&=\varepsilon_m \Ebb\big[\| \bz_m  -Z\|1\{Z\in\mathbb{F}_m(\bz_0)\}\big]
	\\&\geq \varepsilon_m (\|\bz_m-\bz_0\| -r_m)P\big(Z\in\mathbb{F}_m(\bz_0)\big).
	\end{align*}
	Select $\varepsilon_m$ such that  $ \Ebb\big[\|Z_m-Z\|\big]= \delta$.  Then,  it follows from \eqref{eq_zmlow} that
	\begin{align*}
	& \Ebb\big[\ell(Z_m;f)\big]-  \Ebb\big[\ell(Z;f)\big]
	\\&=  \Ebb\big[\ell(Z_m;f)-\ell(Z;f)\mid \pi_m=1\big]P(\pi_m=1)+ \Ebb\big[\ell(Z_m;f)-\ell(Z;f)\mid \pi_m=0\big]P(\pi_m=0)
	\\&=\varepsilon_m\Ebb\big[\ell(\bz_m 1\{Z\in\mathbb{F}_m(\bz_0)\}+Z1\{Z\notin\mathbb{F}_m(\bz_0)\};f)-\ell(Z;f)  \big]
	\\&=\varepsilon_m\Ebb\big[\{\ell(\bz_m;f)-\ell(Z;f) \}1\{Z\in\mathbb{F}_m(\bz_0)\} \big]
	\\&\geq  \big\{\|\ell_{f}\|_{\Lip}-m^{-1}\big\}\varepsilon_m\Ebb\big[ \big\|\bz_m-Z\big\|1\{Z\in\mathbb{F}_m(\bz_0)\}\big]-2\|\ell_{f}\|_{\Lip}\varepsilon_mr_mP\big(Z\in\mathbb{F}_m(\bz_0)\big)
	\\&= \delta \big\{\|\ell_{f}\|_{\Lip}-m^{-1}\big\}-2\|\ell_{f}\|_{\Lip}\varepsilon_mr_mP\big(Z\in\mathbb{F}_m(\bz_0)\big).
	\end{align*}
	In addition, the following condition
	\begin{align*}
	\varepsilon_m r_mP\big(Z\in\mathbb{F}_m(\bz_0)\big)\leq \frac{r_mP\big(Z\in\mathbb{F}_m(\bz_0)\big)\delta}{(\|\bz_m-\bz_0\| -r_m)P\big(Z\in\mathbb{F}_m(\bz_0)\big)}\leq \frac{r_m\delta }{(\|\bz_m-\bz_0\| -r_m)}\rightarrow 0
	\end{align*}
	can be satisfied if $\|\bz_m-\bz_0\| \rightarrow \infty$ and $r_m$ is selected as a   fixed value, or when $\|\bz_m-\bz_0\|\geq \tau$ for some $\tau>0$ and  $r_m\rightarrow0.$  In such case,
	\begin{align*}
	\cR_{\Pbb,1}(f;\delta)-\cR_{\Pbb,1}(f;0)\geq \Ebb\big[\ell(Z_m;f)\big]-  \Ebb\big[\ell(Z;f)\big] \quad \text{ for all  }m\in\mathbb{N}.
	\end{align*}
	Therefore, we show
		\begin{align*}
	\cR_{\Pbb,1}(f;\delta)-\cR_{\Pbb,1}(f;0)= \delta \|\ell_{f}\|_{\Lip}.
	\end{align*}

	The first case that $\|\bz_m-\bz_0\| \rightarrow \infty$  is satisfied under the condition (i), that is,
	\[
	{\lim\sup}_{\|\bz-\bz_0\|\rightarrow\infty}\frac{\ell(\bz;f)-\ell(\bz_0;f)}{\|\bz-\bz_0\|}= \|\ell_{f}\|_{\Lip}.
	\]
	The second case that $\|\bz_m-\bz_0\|\geq \tau$ for some $\tau>0$ and $r_m\rightarrow0$ is satisfied under the condition (ii), that is,
	\[
	{\sup}_{\|\bz-\bz_0\|\geq \tau}\frac{\ell(\bz;f)-\ell(\bz_0;f)}{\|\bz-\bz_0\|}= \|\ell_{f}\|_{\Lip}
	\]
	and  $P(Z\in \mathbb{F}_m(\bz_0)) >0$. Hence, the proof is completed.
\end{proof}

\phantomsection
   \addcontentsline{toc}{subsection}{Lemma 2.4}
 	\begin{statement}{Lemma 2.4}
 	Suppose there exists a positive number $q>0$ and a measuable function $H(\bz;f)$ such that for any $\tilde{\bz},\bz\in\cZ,$ there holds
 	\[
 	\|\nabla \ell(\tilde{\bz};f)-\nabla \ell(\bz;f)\|_1\leq H(\bz;f)\|\tilde{\bz}-\bz\|_{\infty}^q.
 	\]
 	Let $\mathbb{A}=\{\bz:\|\nabla\ell(\bz;f)\|_1=\esssup(\|\nabla\ell(Z;f)\|_1)\}$.   Suppose    $\Ebb[H(Z;f) 1\{Z\in\mathbb{A}\}]<\infty$ and $P(Z\in \mathbb{A})>0$,  then
 	\begin{align*}
 		\delta\esssup(\|\nabla\ell(Z;f)\|_1)-  \delta^{1+q}  \frac{\Ebb\big[H(Z;f) 1\{Z\in\mathbb{A}\}\big] }{P(Z\in \mathbb{A})^{1+q}}\leq \cR_{\Pbb,1}(f;\delta)-\cR_{\Pbb}(f) \leq \delta \|\ell_f\|_{\Lip}.
 	\end{align*}
 \end{statement}

\begin{proof}[{\bf Proof of Lemma 2.4}]
	First, the upper bound can be   derived from  Lemma 2.3 in the main text. Next, to derive the lower bound,   define $\mathbb{A}=\{\bz:\|\nabla\ell(\bz;f)\|_1=\esssup(\|\nabla\ell(Z;f)\|_1)\}$ as the set of points where the norm of $\nabla\ell(\bz; f)$ reaches its essential supremum. Here,   $\nabla\ell(\bz;f)$ denotes the gradient of $\ell(\bz;f)$ with respect to $\bz.$ Define
	\[
	\widetilde{\Delta}_1=\frac{\delta\operatorname{sgn}(\nabla\ell(Z;f))1\{Z\in\mathbb{A}\}}{P(Z\in \mathbb{A})},
	\]
	where $\operatorname{sgn}(\cdot)$ is applied element-wise and satisfies $\operatorname{sgn}(x)=x/|x|$ for $x\ne0$ and $\operatorname{sgn}(x)=0$ when $x=0$.
Then, we have  $\Ebb[\|\widetilde{\Delta}_1\|_{\infty}]\leq \delta$ and $\Ebb[\widetilde{\Delta}_1^{\top}\nabla\ell(Z;f)]=\delta\esssup(\|\nabla\ell(Z;f)\|_1).$ This in fact indicates that $\widetilde{\Delta}_1$ is selected such that $\sup_{\Ebb[\|\Delta\|_{\infty}]\leq \delta} \Ebb[\Delta^{\top}\nabla\ell(Z;f)]$ is reached, which is upper bounded by $\delta\esssup(\|\nabla\ell(Z;f)\|_1)$ due to H{\"o}lder's inequality. It then follows
	\begin{align*}
	&\cR_{\Pbb,1}(f;\delta)-\cR_{\Pbb,1}(f;0) =\sup_{\Ebb[\|\Delta\|_{\infty}]\leq \delta} \Ebb\big[\ell(Z+\Delta;f)-\ell(Z;f)\big]
	\\&=\sup_{\Ebb[\|\Delta\|_{\infty}]\leq \delta} \Ebb\big[\int_0^1\Delta^{\top}\nabla\ell(Z+t\Delta;f)dt\big]
	\\&=\sup_{\Ebb[\|\Delta\|_{\infty}]\leq \delta} \Ebb\big[\Delta^{\top}\nabla\ell(Z;f)+ \int_0^1\Delta^{\top}\{\nabla\ell(Z+t\Delta;f)-\nabla\ell(Z;f)\}dt\big]
	\\&\geq  \Ebb\big[\widetilde{\Delta}_1^{\top}\nabla\ell(Z;f)+ \int_0^1\widetilde{\Delta}_1^{\top}\{\nabla\ell(Z+t\widetilde{\Delta}_1;f)-\nabla\ell(Z;f)\}dt\big]
	\\&\geq  \delta\esssup(\|\nabla\ell(Z;f)\|_1)-\int_0^1\Ebb\big[\big|\widetilde{\Delta}_1^{\top}\{\nabla\ell(Z+t\widetilde{\Delta}_1;f)-\nabla\ell(Z;f)\}\big|\big]dt.
	\end{align*}
	To bound the second term in the last inequality,  note that the growth condition  for the gradient of the  loss function implies that
	\begin{align*}
	\|\nabla\ell(Z+t\widetilde{\Delta}_1;f)-\nabla\ell(Z;f)\|_1\leq H(Z;f)\|t\widetilde{\Delta}_1\|_{\infty}^q,
	\end{align*}
	from which we derive
	\begin{align*}
&	\int_0^1\Ebb\big[\big|\widetilde{\Delta}_1^{\top}\{\nabla\ell(Z+t\widetilde{\Delta}_1;f)-\nabla\ell(Z;f)\}\big|\big]dt
\\	&\leq  \int_0^1\Ebb\big[\|\widetilde{\Delta}_1\|_{\infty}\|\nabla\ell(Z+t\widetilde{\Delta}_1;f)-\nabla\ell(Z;f)\|_1\big]dt
	\\&\leq  \int_0^1\Ebb\big[\|\widetilde{\Delta}_1\|_{\infty}H(Z;f)\|t\widetilde{\Delta}_1\|_{\infty}^q\big]dt
	\\&\leq \Ebb\big[H(Z;f)\|\widetilde{\Delta}_1\|_{\infty}^{1+q}\big]
 \leq  \delta^{1+q}  \frac{\Ebb\big[H(Z;f) 1\{Z\in\mathbb{A}\}\big] }{P(Z\in \mathbb{A})^{1+q}}.
	\end{align*}
	Therefore, from $\cR_{\Pbb,1}(f;0)=\cR_{\Pbb}(f)$, we derive
	\begin{align*}
	\delta\esssup(\|\nabla\ell(Z;f)\|_1)-  \delta^{1+q}  \frac{\Ebb\big[H(Z;f) 1\{Z\in\mathbb{A}\}\big] }{P(Z\in \mathbb{A})^{1+q}}\leq  \cR_{\Pbb,1}(f;\delta)-\cR_{\Pbb}(f) \leq \delta \|\ell_f\|_{\Lip}.
	\end{align*}
	This completes the proof.
\end{proof}

\phantomsection
 \addcontentsline{toc}{subsection}{Lemma 2.5}
 	\begin{statement}{Lemma 2.5}
 	Suppose there exists a measuable function $H(\bz;f)$ such that  for any $\tilde{\bz},\bz\in\cZ,$ there holds
 	\[
 	\big\|\nabla \ell(\tilde{\bz};f)-\nabla \ell(\bz;f)\big\|_1\leq H(\bz;f)\big\|\tilde{\bz}-\bz\big\|_{\infty}^{\min\{1,k-1\}}.
 	\]
 	When $k\in(2,\infty)$ and  $H(\bz;f)\in L^{\frac{k}{k-2}}(\Pbb)$, we have
 	\begin{align*}
 		\big|\cR_{\Pbb,k}(f;\delta)-\cR_{\Pbb}(f) - \delta\Ebb\big[\|\nabla\ell(Z;f)\|_1^{k^{\star}}\big]^{ 1/k^{\star} }\big|\leq \delta^2 \Ebb\big[H(Z;f)^{\frac{k}{k-2}}\big]^{1-\frac{2}{k}}.
 	\end{align*}
 	When $k\in(1,2]$ and  $H(\bz;f)\in L^{\infty}(\Pbb)$,  we have
 	\begin{align*}
 		\big|	\cR_{\Pbb,k}(f;\delta)-\cR_{\Pbb}(f) - \delta\Ebb\big[\|\nabla\ell(Z;f)\|_1^{k^{\star}}\big]^{ 1/k^{\star}  }\big|\leq \delta^{k}\esssup\big(H(Z;f)\big).
 	\end{align*}	
\end{statement}

\begin{proof}[{\bf Proof of Lemma  2.5}]
	From the definition of the Wasserstein distance, the worst-case risk $\cR_{\Pbb,k}(f;\delta)$ can be reformulated as
	\begin{align*}
	\cR_{\Pbb,k}(f;\delta)  =\sup_{\tilde{Z}\in\cP_k(\cZ):\Ebb[\|\tilde{Z}-Z\|_{\infty}^k]\leq \delta^k}\Ebb [\ell(\tilde{Z};f) ]
	=\sup_{\Delta\in\cP_k(\cZ):\Ebb[\|\Delta\|_{\infty}^k]\leq \delta^k}\Ebb [\ell(Z+\Delta;f) ].
	\end{align*}
	Define $k^{\star}$ as the integer such that $1/k+1/k^{\star}=1$.  Then,
	\begin{align*}
&	\cR_{\Pbb,k}(f;\delta)-\cR_{\Pbb,k}(f;0)=\sup_{\Ebb[\|\Delta\|_{\infty}^k]\leq \delta^k} \Ebb\big[\ell(Z+\Delta;f)-\ell(Z;f)\big]
\\& =\sup_{\Ebb[\|\Delta\|_{\infty}^k]\leq \delta^k} \Ebb\big[\int_0^1\Delta^{\top}\nabla\ell(Z+t\Delta;f)dt\big]
	\\&=\sup_{\Ebb[\|\Delta\|_{\infty}^k]\leq \delta^k} \Ebb\big[\Delta^{\top}\nabla\ell(Z;f)+ \int_0^1\Delta^{\top}\{\nabla\ell(Z+t\Delta;f)-\nabla\ell(Z;f)\}dt\big]
	\\&\leq\sup_{\Ebb[\|\Delta\|_{\infty}^k]\leq \delta^k} \Ebb\big[\|\Delta\|_{\infty}\|\nabla\ell(Z;f)\|_1+\int_0^1\Delta^{\top}\{\nabla\ell(Z+t\Delta;f)-\nabla\ell(Z;f)\}dt\big]
	\\&\leq \delta\Ebb\big[\|\nabla\ell(Z;f)\|_1^{k^{\star}}\big]^{\frac{1}{k^{\star}}}+\sup_{\Ebb[\|\Delta\|_{\infty}^k]\leq \delta^k}\Ebb\big[\int_0^1\Delta^{\top}\{\nabla\ell(Z+t\Delta;f)-\nabla\ell(Z;f)\}dt\big],
	\end{align*}
	where the last two inequalities follow from  Hölder's inequality. Define
	\[
	\widetilde{\Delta}=\frac{\delta\operatorname{sgn}(\nabla\ell(Z;f))}{\|\nabla\ell(Z;f)\|_1^{1-k^{\star}}}\Ebb[\|\nabla\ell(Z;f)\|_1^{k^{\star}}]^{-\frac{1}{k}}.
	\]
	It satisfies $\Ebb[\|\widetilde{\Delta}\|_{\infty}^k]\leq \delta^k$ and   $ \Ebb [\widetilde{\Delta}^{\top}\nabla\ell(Z;f)]=\delta\Ebb\big[\|\nabla\ell(Z;f)\|_1^{k^{\star}}\big]^{1/k^{\star}}$.  Then, we have
	\begin{align*}
	\cR_{\Pbb,k}(f;\delta)-\cR_{\Pbb,k}(f;0)&  =\sup_{\Ebb[\|\Delta\|_{\infty}^k]\leq \delta^k} \Ebb\big[\Delta^{\top}\nabla\ell(Z;f)+ \int_0^1\Delta^{\top}\{\nabla\ell(Z+t\Delta;f)-\nabla\ell(Z;f)\}dt\big]
	\\&\geq  \Ebb\big[\widetilde{\Delta}^{\top}\nabla\ell(Z;f)+ \int_0^1\widetilde{\Delta}^{\top}\{\nabla\ell(Z+t\widetilde{\Delta};f)-\nabla\ell(Z;f)\}dt\big]
	\\&\geq  \delta\Ebb\big[\|\nabla\ell(Z;f)\|_1^{k^{\star}}\big]^{\frac{1}{k^{\star}}}-\int_0^1\Ebb\big[\big|\widetilde{\Delta}^{\top}\{\nabla\ell(Z+t\widetilde{\Delta};f)-\nabla\ell(Z;f)\}\big|\big]dt.
	\end{align*}
	
	When $k\in(2,\infty)$,   the second term in the last inequality can be upper bounded. From Hölder's inequality, we have
	\begin{align*}
	\int_0^1\Ebb\big[\big|\widetilde{\Delta}^{\top}\{\nabla\ell(Z+t\widetilde{\Delta};f)-\nabla\ell(Z;f)\}\big|\big]dt
	&\leq  \int_0^1\Ebb\big[\|\widetilde{\Delta}\|_{\infty}\|\nabla\ell(Z+t\widetilde{\Delta};f)-\nabla\ell(Z;f)\|_1\big]dt
	\\&\leq  \int_0^1\Ebb\big[\|\widetilde{\Delta}\|_{\infty}H(Z;f)\|t\widetilde{\Delta}\|_{\infty}\big]dt
	\\&\leq \Ebb\big[H(Z;f)\|\widetilde{\Delta}\|_{\infty}^2\big]
	\\&\leq  \Ebb\big[H(Z;f)^{\frac{k}{k-2}}\big]^{1-\frac{2}{k}}\Ebb\big[\|\widetilde{\Delta}\|_{\infty}^k\big]^{\frac{2}{k}}
	\\&\leq \delta^2 \Ebb\big[H(Z;f)^{\frac{k}{k-2}}\big]^{1-\frac{2}{k}}.
	\end{align*}
 Following a similar analysis, we also derive
	\begin{align*}
	\sup_{\Ebb[\|\Delta\|_{\infty}^k]\leq \delta^k}\Ebb\big[\int_0^1\Delta^{\top}\{\nabla\ell(Z+t\Delta;f)-\nabla\ell(Z;f)\}dt\big]
	\leq \delta^2 \Ebb\big[H(Z;f)^{\frac{k}{k-2}}\big]^{1-\frac{2}{k}}.
	\end{align*}
	Consequently, 	from $\cR_{\Pbb,k}(f;0)=\cR_{\Pbb}(f)$, we derive
	\begin{align*}
	\big|\cR_{\Pbb,k}(f;\delta)-\cR_{\Pbb}(f) - \delta\Ebb\big[\|\nabla\ell(Z;f)\|_1^{k^{\star}}\big]^{\frac{1}{k^{\star}}}\big|\leq \delta^2 \Ebb\big[H(Z;f)^{\frac{k}{k-2}}\big]^{1-\frac{2}{k}}.
	\end{align*}
	
	When $k\in(1,2]$,  the condition $\|\nabla\ell(\tilde{\bz};f)-\nabla\ell(\bz;f)\|_1\leq H(\bz;f)\|\tilde{\bz}-\bz\|_{\infty}^{k-1}$  and Hölder's inequality  imply
	\begin{align*} &\int_0^1\Ebb\big[\big|\widetilde{\Delta}^{\top}\{\nabla\ell(Z+t\widetilde{\Delta};f)-\nabla\ell(Z;f)\}
\big|\big]dt
	\\&\leq  \int_0^1\Ebb\big[\|\widetilde{\Delta}\|_{\infty}\|\nabla\ell(Z+t\widetilde{\Delta};f)-\nabla\ell(Z;f)\|_1\big]dt
	\\&\leq  \int_0^1\Ebb\big[\|\widetilde{\Delta}\|_{\infty}H(Z;f)\|t\widetilde{\Delta}\|_{\infty}^{k-1}\big]dt
	\\&\leq \Ebb\big[H(Z;f)\|\widetilde{\Delta}\|_{\infty}^k\big]
	\\&\leq  \delta^{k}\esssup(H(Z;f)).
	\end{align*}
 Following a similar analysis, we also derive
	\begin{align*}
	\sup_{\Ebb[\|\Delta\|_{\infty}^k]\leq \delta^k}\Ebb\big[\int_0^1\Delta^{\top}\{\nabla\ell(Z+t\Delta;f)-\nabla\ell(Z;f)\}dt\big]
	\leq\delta^{k}\esssup(H(Z;f)),
	\end{align*}
	from which it follows
	\begin{align*}
	\big|\cR_{\Pbb,k}(f;\delta)-\cR_{\Pbb}(f) - \delta\Ebb\big[\|\nabla\ell(Z;f)\|_1^{k^{\star}}\big]^{\frac{1}{k^{\star}}}\big|\leq \delta^{k}\esssup(H(Z;f)).
	\end{align*}
	This completes the proof.
\end{proof}

\section{Proof of theorems}\label{sec_C}
 \phantomsection
 \addcontentsline{toc}{subsection}{Theorem 3.3}
 \begin{statement}{Theorem 3.3}
 	Let $\cN\cN(W,L,\mathcal{S},K)$  be defined over $  \mathcal{X}$. For any $u > 0$, the following bounds hold.
 	(i) For the supremum norm $\|\cdot\|_{\infty}$, we have
 	\begin{equation*}
 		\begin{split}
 			\log \cN(u, \cN\cN(W,L,\mathcal{S},K) ,\|\cdot\|_{\infty})\lesssim \mathcal{S}\log\left(1+ Ku^{-1} \right).
 		\end{split}
 	\end{equation*}
 	(ii) If further $m \ge 2$, then
 	\begin{equation*}
 		\log \cN(u, \cN\cN(W,L,\mathcal{S},K), \rho_\infty) \lesssim \mathcal{S}\log\left(1+ LK m^Lu^{-1} \right),
 	\end{equation*}
 	where   $\rho_\infty$ is defined as $\rho_\infty(g,\tilde{g}) = \sup_{\bx\in \mathcal{X}}   \{ |g(\bx)-\tilde{g}(\bx)| + \|\nabla g(\bx)-\nabla \tilde{g}(\bx)\|_1 \}.$
 \end{statement}

\begin{proof}[{\bf Proof of Theorem 3.3}]
	Let $B=m^{-L}K$. By Lemma~\ref{lem_normalize}, it suffices to study the subclass
	\[
	\mathcal{T}_m(\cN\cN(W,L,\mathcal{S},K)):=\Big\{ g\in\cN\cN(W,L,\mathcal{S},K):  \|\bA_L \|_{\infty} \leq B, \| (\bA_{\ell}, \boldsymbol{b}_{\ell} ) \|_{\infty} \leq 1, \; 0 \leq \ell \leq L-1 \Big\},
	\]
	which satisfies $\mathcal{T}_m(\cN\cN(W,L,\mathcal{S},K))=\cN\cN(W,L,\mathcal{S},K)$.  Define
	\[
	\Phi_1:=\Big\{\phi_1(\bx)=\sigma_m(\bA_0\bx+\bb_0): \|(\bA_{0},\bb_{0})\|_{\infty}\leq 1\Big\}.
	\]
	For $1\leq \ell\leq L-1$, inductively define
	\[
	\Phi_{\ell+1}:=\Big\{\phi_{\ell+1}(\bx)=\sigma_m(\bA_\ell \phi_{\ell}(\bx)+\bb_\ell):\;\phi_{\ell}\in \Phi_\ell,\;\|(\bA_{\ell},\bb_{\ell})\|_{\infty}\leq 1\Big\}.
	\]
	Then $\mathcal{T}_m(\cN\cN(W,L,\mathcal{S},K))$ can be represented as
	\[
	\mathcal{T}_m(\cN\cN(W,L,\mathcal{S},K))=\Big\{g(\bx)=\bA_L \phi_L(\bx): \phi_L\in\Phi_L,\;\|\bA_L\|_\infty\leq B\Big\}.
	\]
	We construct layer-wise coverings as follows.
	
	\smallskip

	(1) We first construct the cover $\widetilde{\Phi}_1$ for $\Phi_1$. From Lemma \ref{lem_covermatrix}, there exists a finite set
	\[
	\widetilde{\mathcal{M}}_1 :=\big\{(\tilde{\bA}_j,\tilde{\bb}_j): j=1,\dots, (2/t+1)^{d_1(d+1)}\big\}\subseteq \mathcal{M}(d_1,d+1,1)
	\]
	such that for any $\phi_1(\bx)=\sigma_m(\bA_0\bx+\bb_0)\in\Phi_1$,  one can find $(\tilde{\bA}_0,\tilde{\bb}_0)\in\widetilde{\mathcal{M}}_1$ with $\|(\tilde{\bA}_0,\tilde{\bb}_0)\|_{\infty}\leq 1$ satisfying
	\[
	\|(\bA_0-\tilde{\bA}_0,\bb_0-\tilde{\bb}_0)\|_{\infty}\leq t.
	\]
	Define  $\tilde{\phi}_1(\bx)=\sigma_m(\tilde{\bA}_0 \bx+\tilde{\bb}_0 )$.  Then,
	\begin{align*}
		\|\phi_1(\bx)-\tilde{\phi}_1(\bx) \|_{\infty}
		&\leq m	\| (\bA_0-\tilde{\bA}_0,\bb_0-\tilde{\bb}_0)( \bx^{\top},1)^{\top} \|_{\infty}
		\\&\leq m\| (\bA_0-\tilde{\bA}_0,\bb_0-\tilde{\bb}_0)\|_{\infty} \|( \bx^{\top},1)^{\top}\|_{\infty}
		\\&\leq mt.
	\end{align*}
	Define the set
	$\widetilde{\Phi}_1:=\big\{\tilde{\phi}_1(\bx) =\sigma_m(\tilde{\bA} \bx+\tilde{\bb} ):(\tilde{\bA} ,\tilde{\bb})\in	\widetilde{\mathcal{M}}_1\big\}$.
	The above analysis demonstrates that $\|\tilde{\phi}_1(\bx)\|_{\infty}\leq 1$ over $\cX$ and $\widetilde{\Phi}_1$ is a $(mt)$-cover of $\Phi_1$. In addition, the cardinality of $\widetilde{\Phi}_1$ satisfies
	\[
	|\widetilde{\Phi}_1| \leq (2/t+1)^{d_1(d+1)}.
	\]

	(2) For $1 \leq \ell \leq L-1$, suppose we have constructed a set $\widetilde{\Phi}_{\ell}$ that is a $(t \cdot m \frac{m^\ell-1}{m-1})$-cover of $\Phi_{\ell}$, with each element $\tilde{\phi}_\ell \in \widetilde{\Phi}_\ell$ satisfying $\|\tilde{\phi}_{\ell}(\bx) \|_{\infty}\leq 1$. Further, suppose its cardinality satisfies
	\[
	|\widetilde{\Phi}_{\ell}| \leq (2/t+1)^{\sum_{s=0}^{\ell-1}d_{s+1}(d_{s}+1)}.
	\]
	We construct the cover for the next layer. Define
	\[
	\widehat{\Phi}_{\ell+1}:=\big\{\widehat{\phi}_{\ell+1}(\bx)= \sigma_m(\bA_{\ell} \tilde{\phi}_{\ell}(\bx)+\bb_{\ell}): \tilde{\phi}_{\ell}\in\widetilde{\Phi}_{\ell},\;\| (\bA_{\ell}, \boldsymbol{b}_{\ell} ) \|_{\infty} \leq 1 \big\}.
	\]
	From Lemma \ref{lem_covermatrix}, there exists a finite set
	\[
	\widetilde{\mathcal{M}}_{\ell+1} := \{(\tilde{\bA},\tilde{\bb}) : j=1,\dots,(2/t+1)^{d_{\ell+1}(d_{\ell}+1)}\} \subseteq \mathcal{M}(d_{\ell+1},d_{\ell}+1,1)
	\]
	such that for any   $(\bA_\ell,\bb_\ell)$ associated with $\widehat{\phi}_{\ell+1} \in \widehat{\Phi}_{\ell+1}$, one can find $(\tilde{\bA}_\ell,\tilde{\bb}_\ell) \in \widetilde{\mathcal{M}}_{\ell+1}$ with $\|(\tilde{\bA}_\ell,\tilde{\bb}_\ell) \|_{\infty}\leq 1$ satisfying
	\[
	\|(\bA_\ell-\tilde{\bA}_\ell,\;\bb_\ell-\tilde{\bb}_\ell)\|_\infty \leq t.
	\]
	Let $\tilde{\phi}_{\ell+1}(\bx)=\sigma_m(\tilde{\bA}_\ell \tilde{\phi}_{\ell}(\bx)+\tilde{\bb}_\ell)$. Then $\|\tilde{\phi}_{\ell+1}(\bx) \|_{\infty}\leq 1$ and
	\begin{align*}
		\|\widehat{\phi}_{\ell+1}(\bx)-\tilde{\phi}_{\ell+1}(\bx) \|_{\infty}
		&\leq m \| (\bA_{\ell}-\tilde{\bA}_\ell ,\bb_{\ell}-\tilde{\bb}_\ell )( \tilde{\phi}_{\ell}(\bx)^{\top},1)^{\top} \|_{\infty}
		\\&\leq m \| (\bA_{\ell}-\tilde{\bA}_\ell ,\bb_{\ell}-\tilde{\bb}_\ell )\|_{\infty} \|( \tilde{\phi}_{\ell}(\bx)^{\top},1)^{\top}\|_{\infty}
		\\&\leq mt.
	\end{align*}
	Define the cover for layer $\ell+1$ as
	\[
	\widetilde{\Phi}_{\ell+1}:=\big\{\tilde{\phi}_{\ell+1}(\bx) =\sigma_m(\tilde{\bA} \tilde{\phi}_{\ell}(\bx)+\tilde{\bb} ): \tilde{\phi}_{\ell}\in\widetilde{\Phi}_{\ell}, (\tilde{\bA} ,\tilde{\bb})\in	\widetilde{\mathcal{M}}_{\ell+1}\big\}.
	\]
	The cardinality of this set satisfies
	\[
	|\widetilde{\Phi}_{\ell+1}| \leq |\widetilde{\mathcal{M}}_{\ell+1}| |\widetilde{\Phi}_{\ell}|\leq (2/t+1)^{d_{\ell+1}(d_{\ell}+1)}\times(2/t+1)^{\sum_{s=0}^{\ell-1}d_{s+1}(d_{s}+1)}=(2/t+1)^{\sum_{s=0}^{\ell}d_{s+1}(d_{s}+1)}.
	\]
	Now, we show that $\widetilde{\Phi}_{\ell+1}$ is a   cover for $\Phi_{\ell+1}$. For any $ \phi_{\ell+1}\in\Phi_{\ell+1}$, denoted as $\phi_{\ell+1}(\bx)=\sigma_m(\bA_\ell\phi_{\ell}(\bx) +\bb_\ell)$, there exists $\tilde{\phi}_{\ell}\in \widetilde{\Phi}_{\ell}$ such that $\|\phi_{\ell}(\bx)-\tilde{\phi}_{\ell}(\bx)\|_{\infty}\leq t \cdot m \frac{m^\ell-1}{m-1}$. Let $\widehat{\phi}_{\ell+1}(\bx)= \sigma_m(\bA_{\ell} \tilde{\phi}_{\ell}(\bx)+\bb_{\ell})$. We have shown there is a $\tilde{\phi}_{\ell+1}\in\widetilde{\Phi}_{\ell+1}$ such that $\|\widehat{\phi}_{\ell+1} (\bx)-\tilde{\phi}_{\ell+1} (\bx)\|_{\infty} \leq mt.$ Combining these errors using the triangle inequality,
	\begin{equation*}
		\begin{split}
			\|\phi_{\ell+1} (\bx)- \tilde{\phi}_{\ell+1}(\bx) \|_\infty &  \leq  \|\phi_{\ell+1}(\bx) -\widehat{\phi}_{\ell+1} (\bx)\|_{\infty}+ \|\widehat{\phi}_{\ell+1}(\bx) - \tilde{\phi}_{\ell+1} (\bx)\|_{\infty}
			\\&\leq  m\| \bA_\ell (\phi_{\ell}(\bx)-\tilde{\phi}_{\ell}(\bx))\|_{\infty}+mt
			\\&\leq m\| \bA_\ell \|_{\infty}\|\phi_{\ell}(\bx)-\tilde{\phi}_{\ell}(\bx)\|_{\infty}+mt
			\\&\leq m \left(t \cdot m \frac{m^\ell-1}{m-1}\right) + mt  = t\cdot m \frac{m^{\ell+1}-1}{m-1}.
		\end{split}
	\end{equation*}
	Thus, the inductive hypothesis holds.
	\smallskip
	
	(3) Now we construct the cover for the   FNN class. For any $g\in\mathcal{T}_m(\cN\cN(W,L,\mathcal{S},K))$,  it can be represented by
	\begin{equation*}
		g(\boldsymbol{x}) =\bA_L \phi_L(\boldsymbol{x}),\quad \text{for } \phi_L\in \Phi_{L}\text{ and } \|\bA_L \|_{\infty} \leq B.
	\end{equation*}
	Since $\widetilde{\Phi}_{L}$ is a $(t \cdot m \frac{m^L-1}{m-1})$-cover of $\Phi_{L}$, there exists $\tilde{\phi}_{L}\in\widetilde{\Phi}_{L}$ with $\|\tilde{\phi}_{L}(\bx)\|_{\infty}\leq 1$ and $\|\tilde{\phi}_{L}(\bx)-\phi_{L}(\bx)\|_{\infty}\leq t \cdot m \frac{m^L-1}{m-1}$.
	From Lemma~\ref{lem_covermatrix}, there exists a finite set
	\[
	\widetilde{\mathcal{M}}_{L+1} := \{\tilde{\bA}_j: j=1,\dots,(2B/\nu+1)^{d_{L}}\} \subseteq \mathcal{M}(1,d_{L},B)
	\]
	such that for any $\bA_L$ satisfying $ \|\bA_L \|_{\infty} \leq B$, one can find $\tilde{\bA}_L \in \widetilde{\mathcal{M}}_{L+1}$ with $\|\tilde{\bA}_L\|_{\infty} \leq B$ satisfying
	\[
	\|\bA_L - \tilde{\bA}_L\|_\infty \leq \nu.
	\]
	Let $\tilde{g}(\boldsymbol{x}) =\tilde{\bA}_L\tilde{\phi}_{L}(\bx)$. Then,  we have
	\begin{equation}\label{eq_cons_cover}
		\begin{split}
			\|g-\tilde{g}\|_{\infty} &= \|\bA_L \phi_L - \tilde{\bA}_L\tilde{\phi}_L \|_{\infty}
			\\& \leq  \|\bA_L \phi_L -\bA_L\tilde{\phi}_L \|_{\infty}+ \|\bA_L \tilde{\phi}_L - \tilde{\bA}_L\tilde{\phi}_L \|_{\infty}
			\\&\leq B \sup_{\bx\in\cX}\|\phi_L(\bx) -\tilde{\phi}_L(\bx)\|_{\infty}+\|\bA_L -\tilde{\bA}_L\|_{\infty}\sup_{\bx\in\cX}\|\tilde{\phi}_L(\bx)\|_{\infty}
			\\&\leq  B \left(t \cdot m \frac{m^L-1}{m-1}\right) + \nu.
		\end{split}
	\end{equation}
	Define the final cover set $\widetilde{\Phi}:=\big\{\tilde{g}(\bx) = \tilde{\bA} \tilde{\phi}_{L}(\bx) :\tilde{\phi}_{L}\in\widetilde{\Phi}_{L}, \tilde{\bA} \in	\widetilde{\mathcal{M}}_{L+1}\big\}$. Its cardinality satisfies
	\begin{equation}\label{eq_covernu}
		|\widetilde{\Phi}|\leq |\widetilde{\Phi}_{L}| |\widetilde{\mathcal{M}}_{L+1}|\leq (2/t+1)^{\sum_{s=0}^{L-1}d_{s+1}(d_{s}+1)}(2B/\nu+1)^{d_{L}}.
	\end{equation}
	Let $\nu=\frac{u}{2}$ and $Btm \frac{m^L-1}{m-1} =\frac{u}{2}$. This implies $t = \frac{u m^{L-1}(m-1)}{2K(m^L-1)}$. Therefore, we show
	\begin{equation*}
		\begin{split}
		&	\log \cN(u, \cN\cN(W,L,\mathcal{S},K) ,\|\cdot\|_{\infty})
			\\&\leq \sum_{s=0}^{L-1}d_{s+1}(d_{s}+1)\log\left(1+\frac{4K(m^L-1)}{u m^{L-1}(m-1)}\right)+d_{L}\log\left(1+\frac{4Km^{-L}}{u}\right)
			\\&\lesssim \sum_{s=0}^{L-1}d_{s+1}(d_{s}+1)\log\left(1+ K/u \right).
		\end{split}
	\end{equation*}
	
	\medskip
	
	Next, we derive a covering bound that controls both the function values and their gradients.  Let $g\in	\mathcal{T}_m(\cN\cN(W,L,\mathcal{S},K))$, and let $\tilde{g}$  be the cover element constructed in \eqref{eq_cons_cover}. We define their respective pre-activation vectors and gradients. For $g$, let $\phi_0(\bx) = \bx$. The pre-activation vector at layer $\ell$ ($0 \le \ell \le L-1$) is defined as	$\bz_{\ell}(\bx) = \bA_\ell \phi_\ell(\bx) + \bb_\ell$, and the transposed gradient is given by
	\[
	\nabla g(\bx)^\top = \bA_L \bD_{L-1} \bA_{L-1} \cdots \bD_0 \bA_0,\quad \bD_\ell = \text{diag}(\sigma_m'(z_{\ell,1}),\dots, \sigma_m'(z_{\ell,d_{\ell+1}})).
	\]
	Similarly, for $\tilde{g}$, let $\tilde{\phi}_0(\bx) = \bx$. The pre-activation vector at layer $\ell$ ($0 \le \ell \le L-1$) is defined as $\tilde{\bz}_{\ell}(\bx) = \tilde{\bA}_\ell \tilde{\phi}_\ell(\bx) + \tilde{\bb}_\ell$.
	The transposed gradient is
	\[
	\nabla\tilde{g}(\bx)^\top = \tilde{\bA}_L \tilde{\bD}_{L-1} \tilde{\bA}_{L-1} \cdots \tilde{\bD}_0 \tilde{\bA}_0, \quad \tilde{\bD}_\ell = \text{diag}(\sigma_m'(\tilde{z}_{\ell,1}),\dots, \sigma_m'(\tilde{z}_{\ell,d_{\ell+1}})).
	\]
	Define $\Delta\bA_\ell = \bA_\ell - \tilde{\bA}_\ell$ and $\Delta\bD_\ell = \bD_\ell - \tilde{\bD}_\ell$. The difference between the two gradients can be expressed as the telescoping sum.
	\begin{align*}
		\nabla g(\bx)^\top - \nabla\tilde{g}(\bx)^\top = & \big(\Delta\bA_L\big) \big( \tilde{\bD}_{L-1} \cdots \tilde{\bA}_0 \big) \\
		& + \sum_{\ell=0}^{L-1} \big( \bA_L \bD_{L-1} \cdots \bA_{\ell+1} \big) \big(\Delta\bD_{\ell}\big) \big( \tilde{\bA}_{\ell} \cdots \tilde{\bA}_0 \big) \\
		& + \sum_{\ell=0}^{L-1} \big( \bA_L \bD_{L-1} \cdots \bD_{\ell} \big) \big(\Delta\bA_{\ell}\big) \big( \tilde{\bD}_{\ell-1} \cdots \tilde{\bA}_0 \big).
	\end{align*}
	By the triangle inequality, the norm of each term can be bounded separately.
	
	First,	we analyze the norm of terms containing $\Delta\bA_\ell$. For $0\leq\ell\leq L-1$, define
	\[
	T_{\bA_\ell} := \big( \bA_L \bD_{L-1}\cdots \bD_{\ell} \big) (\Delta\bA_{\ell}) \big( \tilde{\bD}_{\ell-1} \cdots \tilde{\bA}_0 \big).
	\]
	By the sub-multiplicative property of the matrix norm, together with the bounds $\|\bA_L\|_\infty \le B$, $\|\bA_s\|_\infty, \|\tilde{\bA}_s\|_\infty \le 1$, and $\|\bD_s\|_\infty, \|\tilde{\bD}_s\|_\infty \le m$ for $s\in[L-1]$, we obtain
	\begin{align*}
		\|T_{\bA_\ell}\|_\infty &\le \|\bA_L\|_\infty \prod_{s=\ell+1}^{L-1} (\|\bD_s\|_\infty \|\bA_s\|_\infty)\|\bD_{\ell}\|_\infty    \|\Delta\bA_\ell\|_\infty   \prod_{s=0}^{\ell-1} (\|\tilde{\bD}_s\|_\infty \|\tilde{\bA}_s\|_\infty) \\
		&\le  B   m^{L-\ell-1}   m t   m^\ell  =t B m^L.
	\end{align*}
	Further, define $T_{\bA_L} = (\Delta\bA_L)  ( \tilde{\bD}_{L-1} \cdots \tilde{\bA}_0  )$. Its norm satisfies
	\begin{align*}
		\|T_{\bA_L}\|_\infty &\le \|\Delta\bA_L\|_\infty    \prod_{s=0}^{L-1} (\|\tilde{\bD}_s\|_\infty \|\tilde{\bA}_s\|_\infty)  \leq \nu   m^L.
	\end{align*}

	Next, we investigate the terms containing $\Delta\bD_{\ell}$. For $0\leq\ell\leq L-1$, define
	\[
	T_{\bD_\ell} := \big( \bA_L \bD_{L-1}\cdots \bA_{\ell+1} \big) \big(\Delta\bD_{\ell}\big) \big( \tilde{\bA}_{\ell} \cdots \tilde{\bA}_0 \big).
	\]
	Its norm can be bounded by
	\begin{align*}
	\|T_{\bD_\ell}\|_\infty &\leq  \|\bA_{L}\|_{\infty}\prod_{s=\ell+1}^{L-1} (\| {\bD}_s\|_\infty \| {\bA}_s\|_\infty)  \|\Delta{\bD}_\ell\|_\infty\|\tilde{\bA}_{\ell}\|_\infty  \prod_{s=0}^{\ell-1} (\|\tilde{\bD}_s\|_\infty \|\tilde{\bA}_s\|_\infty)
\\&	\leq B m^{L-1}\|\Delta{\bD}_\ell\|_\infty.
	\end{align*}
	We now  bound $\|\Delta{\bD}_\ell\|_\infty$. Since $\sigma_m'(z) = m(\max\{z,0\})^{m-1}$, we have
	\[
	\|\bD_\ell - \tilde{\bD}_\ell\|_\infty = \max_i |\sigma_m'(z_{\ell,i}) - \sigma_m'(\tilde{z}_{\ell,i})| \le m(m-1) \|\bz_{\ell} - \tilde{\bz}_{\ell}\|_\infty.
	\]
	Applying the triangle inequality, we obtain
	\begin{align*}
		\|\bz_{\ell} - \tilde{\bz}_{\ell}\|_\infty &= \big\| (\bA_\ell \phi_\ell(\bx) + \bb_\ell ) -  (\tilde{\bA}_\ell \tilde{\phi}_\ell (\bx)+ \tilde{\bb}_\ell )\big\|_\infty \\
		&\le\big \|\bA_\ell(\phi_\ell (\bx)- \tilde{\phi}_\ell(\bx))\big\|_\infty + \big\|(\bA_\ell - \tilde{\bA}_\ell)\tilde{\phi}_\ell (\bx)+ (\bb_\ell - \tilde{\bb}_\ell)\big\|_\infty \\
		&\le \|\bA_\ell\|_\infty \|\phi_\ell (\bx)- \tilde{\phi}_\ell(\bx)\|_\infty + \|(\bA_\ell - \tilde{\bA}_\ell, \bb_\ell - \tilde{\bb}_\ell)\|_\infty \|( \tilde{\phi}_{\ell}(\bx)^{\top},1)^{\top}\|_{\infty} \\
		&\le 1 \cdot \left(t \cdot m \frac{m^\ell-1}{m-1}\right) + t \cdot 1 = t \cdot \frac{m^{\ell+1}-1}{m-1}.
	\end{align*}
This implies that
	\begin{align*}
		\|T_{{\bD}_\ell}\|_\infty &\le B m^{L-1} \|\Delta{\bD}_{\ell}\|_\infty
		\leq B m^{L-1} \left( m(m-1) \|\bz_{\ell} - \tilde{\bz}_{\ell}\|_\infty \right) \\
		&\le B m^{L-1} \left( m(m-1) \cdot t \frac{m^{\ell+1}-1}{m-1} \right)
		=t  B   m^L  (m^{\ell+1}-1).
	\end{align*}
	Thus, we derive
	\begin{equation} \label{eq_grad_error}
		\begin{split}
			\sup_{\bx\in\cX}	\big	\|\nabla g(\bx)- \nabla \tilde{g}(\bx)\big\|_1 &\le  \nu m^L+ \sum_{\ell=0}^{L-1} tB   m^L  + \sum_{\ell=0}^{L-1} tB   m^L  (m^{\ell+1}-1) \\
			&\leq \nu m^L + tL  B m^L +t Bm^{2L+1} .
		\end{split}
	\end{equation}

	Combining the results from \eqref{eq_cons_cover} and \eqref{eq_grad_error}, we obtain
	\[
	\rho_\infty(g, \tilde{g}) \le \nu(m^L+1)+t \left( B  m \frac{m^L-1}{m-1}  + LBm^{L} +Bm^{2L+1} \right).
	\]
	To construct an $u$-cover, we choose $t$ and $\nu$ such that the right-hand side is at most  $u$. This can be done by requiring each component to be at most $u/2$. Setting $\nu(m^L+1) = u/2$ gives   $\nu   = u/ (2m^L+2)$. For the terms involving  $t$, it is  dominated  by $tL B  m^{2L} $, which requires
	$t \asymp  \frac{u}{2 L K m^L }.$ From \eqref{eq_covernu}, the cardinality of the cover $\widetilde{\Phi}$ is bounded by
	\begin{equation*}
		|\widetilde{\Phi}| \leq (2/t+1)^{\sum_{s=0}^{L-1}d_{s+1}(d_{s}+1)}(2B/\nu+1)^{d_{L}}.
	\end{equation*}
	Plugging the chosen values of $t$ and $\nu$, we obtain
	\begin{align*}
	&	\log \cN(u, \cN\cN(W,L,\mathcal{S},K) , \rho_\infty)
	\\ &\lesssim \sum_{s=0}^{L-1}d_{s+1}(d_{s}+1)\log\left(1+\frac{4 L K m^L  }{u}\right) + d_L \log\left(1+\frac{4 K (1+m^{-L})}{u }\right)
		\\&\lesssim \sum_{s=0}^{L-1}d_{s+1}(d_{s}+1)\log\left(1+ \frac{LK m^L}{u} \right).
	\end{align*}
		This completes the proof.
\end{proof}

\phantomsection
	\addcontentsline{toc}{subsection}{Theorem 3.4}
\begin{statement}{Theorem 3.4}
For any $f\in\mathcal{H}^\alpha(\mathcal{X})$, if  $\cN\cN(W,L,\mathcal{S},K)$ satisfies $W\!\ge\! N$,  $\mathcal{S}\!\ge\! N(d+1)$, and $K\!\gtrsim\!\max\{N^{\frac{d+2m+1-2\alpha}{2d}},\sqrt{\log N}\}$, then
	there exists $g\in\cN\cN(W,L,\mathcal{S},K)$ such that
	\begin{align*}
		\sup_{\bx\in\mathcal{X}}|g(\bx)-f(\bx)|&\lesssim
		N^{-\frac{\min\{d+2m+1,2\alpha\}}{2d}}\sqrt{\log N}\\
		\sup_{\bx\in\mathcal{X}}\|\nabla(g-f)(\bx)\|_\infty&\lesssim
		N^{-\frac{\min\{d+2m-1,2(\alpha-1)\}}{2d}}\sqrt{\log N}.
	\end{align*}
\end{statement}

\begin{proof}[{\bf Proof of Theorem 3.4}]
	The proof first establishes the approximation results of
	\[
	\cF_{\sigma_m}(W,K) := \Bigl\{ f(\bx) = \sum_{i=1}^W a_i \, \sigma_m((\bx^{\top},1)\bv_i) : \bv_i \in \mathbb{S}^d, \sum_{i=1}^W |a_i| \le K \Bigr\}
	\]
	on the unit ball $\mathbb{B}^d := \{\bx \in \mathbb{R}^d : \|\bx\|_2 \le 1\}$ follows the techniques of \cite{bach2017breaking,yang2024optimal}, and then extends the result to the hypercube $\mathcal{X} := \{\bx \in \mathbb{R}^d : \|\bx\|_\infty \le 1\}$ via $\mathcal{X} \subseteq \{\bx \in \mathbb{R}^d : \|\bx\|_2 \le \sqrt{d}\}$. Since
	$\cF_{\sigma_m}(W,K)\subseteq\cN\cN(W, 1, W(d+1), mK(d+1)^{m/2})\subseteq \cN\cN(\tilde{W}, \tilde{L}, \tilde{\mathcal{S}}, \tilde{K})$, for any $\tilde{W}\geq W$, $\tilde{L}\geq 1$, $\tilde{\mathcal{S}}\geq W(d+1)$, and $\tilde{K}\geq mK(d+1)^{m/2}$, any approximation result established for   $\cF_{\sigma_m}(W,K)$ directly extends to these larger classes.
	
	The analysis of $\cF_{\sigma_m}(W,K)$ over $\mathbb{B}^d$ proceeds by lifting the target function $h \in \mathcal{H}^\alpha(\mathbb{B}^d)$ to a smooth function $\widetilde{h}$ on the sphere $\mathbb{S}^d \subseteq \mathbb{R}^{d+1}$. The extension $\widetilde{h}$  is then approximated by a spherical polynomial. Restricting this polynomial back to the unit ball $\mathbb{B}^d$ yields a function that approximates $h$ while preserving key structural properties. Finally, the results of \cite{siegel2023optimal} show that this polynomial approximation can be further approximated by  $\cF_{\sigma_m}(W,K)$.
	
	In particular, Proposition 3.3 of \cite{yang2024optimal} shows that for any $h\in \cH^\alpha(\mathbb{B}^d)$, there exists $\widetilde{h}\in C^{r,\beta}(\mathbb{S}^d)$ with $\alpha=r+\beta$, where $r\in\mathbb{N}_0$ and $\beta\in(0,1]$, such that
	\[
	S_{m} \widetilde{h} =h\; \text{ on } \mathbb{B}^d, \quad \text{ and } \quad\omega_{2s^*}(\widetilde{h};t)_\infty \le C t^{\alpha},
	\]
	where $S_m$ is the operator defined in Lemma~\ref{lem_Sk_property}, $s^*\in \mathbb{N}$ is the smallest integer satisfying $\alpha\leq 2s^*$,   $\omega_{2s^*}(\widetilde{h};t)_\infty$ denotes the $2s^*$-th order modulus of smoothness of $\widetilde{h}$, and $C$ is a constant independent of $h$. Moreover, $\widetilde{h}$ can be chosen odd (resp. even) when $m$ is even (resp. odd).
	
	Next, we approximate  $\widetilde{h}$ on the sphere $\mathbb{S}^d$ using spherical polynomials.  This procedure is based on the theory of spherical harmonics, with the key definitions summarized below  (see \cite{Daibook} for details). Let $\mathcal{H}_{\ell}^{d+1}$ denote the space of spherical harmonics of degree $\ell \in \mathbb{N}_0$, i.e., the restrictions to $\mathbb{S}^d$ of real homogeneous harmonic polynomials of degree $\ell$.  Its dimension is    $a_{\ell}^{d+1} := \frac{2\ell+d-1}{\ell}\binom{\ell+d-2}{d-1}$ for $\ell\neq 0$, and $a_0^{d+1}:=1$.  As shown in Theorem 2.2.2 of \cite{Daibook}, $\mathcal{H}_{\ell}^{d+1}$ are mutually orthogonal and their direct sum is dense in $L^2(\mathbb{S}^d)$,  yielding the decomposition $L^2(\mathbb{S}^d)=\bigoplus_{\ell=0}^\infty  \mathcal{H}_{\ell}^{d+1}$. Then, any $f\in L^2(\mathbb{S}^d)$
	can be represented as $f=\sum_{\ell=0}^{\infty} \operatorname{proj}_{\ell}f$, where $\operatorname{proj}_{\ell} : L^2(\mathbb{S}^d) \mapsto \mathcal{H}_{\ell}^{d+1}$ is the orthogonal projection from  $L^2(\mathbb{S}^d)$  onto $\mathcal{H}_{\ell}^{d+1}$, which admits the   representation
	\[
	\operatorname{proj}_{\ell}f(\bx) = a_{\ell}^{d+1} \int_{\mathbb{S}^d} f(\by) P_{\ell}(\bx^\top \by) \, d\tau_d(\by),
	\]
	where $\tau_d$ is the normalized surface area measure on $\mathbb{S}^d$ and $P_{\ell}$ is the Gegenbauer polynomial
	\[
	P_{\ell}(t) := \frac{(-1)^{\ell}}{2^{\ell}} \frac{\Gamma(d/2)}{\Gamma(\ell+d/2)} (1-t^2)^{1-\frac{d}{2}} \left(\frac{d}{dt}\right)^{\ell}\Big( (1-t^2)^{\ell+\frac{d-2}{2}}\Big), \quad t \in [-1,1],
	\]
	normalized so that $P_{\ell}(1)=1$. For $\ell \neq 0$, $P_{\ell}$ is odd (resp. even) when $\ell$ is odd (resp. even).

	Let  $\Pi_{\ell}(\mathbb{S}^d)$ denote  the space of  spherical polynomials of degree at most $\ell$. For any $f\in C(\mathbb{S}^d)$,  the error of the best approximation to $f$ by  $\Pi_{\ell}(\mathbb{S}^d)$  is defined as
	\[
	E_{\ell}(f)_{\infty}:=\inf_{g\in\Pi_{\ell}(\mathbb{S}^d)}\|f-g\|_{L^\infty(\mathbb{S}^d)} .
	\] 	
	Since $\Pi_{\ell}(\mathbb{S}^d)$ is finite-dimensional, a best approximation exists but is generally intractable. In practice, near-best approximations are sufficient. To this end,  we let $\eta$ be a $C^\infty$ function supported on $[0,2]$ with $\eta(t)=1$ for $0 \le t \le 1$, and define
	\[
	L_{\ell} f(\bx) :=f*L_{\ell}(\bx)=   \int_{\mathbb{S}^{d}}f(\by) L_{\ell}(\bx^\top \by) \, d\tau_d(\by), \quad \bx \in \mathbb{S}^d,
	\]
	where  $L_{\ell}(t)  := \sum_{r=0}^{\infty} \eta\left(\frac{r}{\ell}\right)a_r^{d+1}  P_r(t)$. Equivalently,
	\[
	L_{\ell} f(\bx)  =\sum_{r=0}^{\infty}\eta\left(\frac{r}{\ell}\right)\operatorname{proj}_r f(\bx), \quad \bx \in \mathbb{S}^d.
	\]
	Since $ \eta\left(\frac{r}{\ell}\right)=0$ for $r \ge 2\ell$,  $L_{\ell} f$ is  a spherical polynomial of degree at most $2\ell-1$.  The best approximation error can be characterized in terms of the modulus of smoothness, i.e.,
	\begin{equation}\label{eq_appromu}
		\|f-L_{\ell}f\|_{L^\infty(\mathbb{S}^d)} \lesssim 	E_{\ell}(f)_{\infty}\lesssim \omega_r(f;\ell^{-1})_{\infty},
	\end{equation}
	where the first inequality follows from Theorem 2.6.3 of \cite{Daibook} and the second from Theorems 4.7.1 and 4.4.2 of \cite{Daibook}.  For $1\leq i\ne j\leq d+1$, define the angular derivative  $D_{i,j} =u_i\partial_{u_j}-u_j\partial_{u_i}.$
	Theorem 4.5.5  of \cite{Daibook} further shows that $L_{\ell}f$ also provides a near-best simultaneous approximation to $D_{i,j} f$, in the sense that
	\begin{equation} \label{eq_approderi}
		\|D_{i,j}f-D_{i,j}L_{\ell}f\|_{L^\infty(\mathbb{S}^d)} \lesssim E_{\ell}(D_{i,j} f)_{\infty} \lesssim \omega_r(D_{i,j} f;\ell^{-1})_{\infty}.
	\end{equation}
	
	Now, we define the approximation polynomial $g_{\ell}=L_{\ell} \tilde{h}$. Since $\omega_{2s^*}(\widetilde{h};t)_{\infty} \lesssim t^{\alpha}$, and noting from a direct calculation that $S_m D_{i,j}\widetilde{h}$ consists of polynomial multiples of the derivatives of $h$, \eqref{eq_appromu} and \eqref{eq_approderi} yield
	\begin{equation*}
		\|\tilde{h}-g_{\ell}\|_{L^\infty(\mathbb{S}^d)} \lesssim   \ell^{-\alpha},\quad \text{and }\;\|D_{i,j}\tilde{h}-D_{i,j}g_{\ell}\|_{L^\infty(\mathbb{S}^d)} \lesssim \ell^{-(\alpha-1)}.
	\end{equation*}
   Following \cite{yang2024optimal}, we consider three smoothness regimes and summarize their main results. In  Case~I ($\alpha >(d+2m+1)/2$ or $\alpha =(d+2m+1)/2$ is an even integer), we have  $\tilde{h}\in  \mathcal{G}_{\sigma_m}(C(d,m,\alpha))$ for some constant $C(d,m,\alpha)$ depending on $d,m,$ and $\alpha$. Here,  $\cG_{\sigma_m}(K) $ is a class of functions defined over $\mathbb{S}^d$ as follows.
	\[
	\cG_{\sigma_m}(K): = \Big \{ g \in L^\infty(\mathbb{S}^d):  \;\;
	g(\bu) = \int_{\mathbb{S}^d} \sigma_m(\bu^{\top} \bv)  d\mu(\bv),\;  \|\mu\|\le K\Big \}.
	\]
	In  Case~II ($\alpha =(d+2m+1)/2$ is not an even integer), we have $g_{\ell} \in \mathcal{G}_{\sigma_m}(K)$ with $K \lesssim  \sqrt{\log \ell}$, leading to
	\[
	\|\tilde h - g_{\ell}\|_{L^\infty(\mathbb{S}^d)}
	\lesssim \exp\{-\alpha K^2\},\quad \text{and }\; \|D_{i,j}(\tilde h - g_{\ell})\|_{L^\infty(\mathbb{S}^d)}
	\lesssim \exp\{-(\alpha-1)K^2\}.
	\]
	In  Case~III ($\alpha < (d+2m+1)/2$), we have $g_{\ell} \in \mathcal{G}_{\sigma_m}(K)$ with $K \lesssim \ell^{\frac{d+2m+1-2\alpha}{2}}$, leading to
	\[
	\|\tilde h - g_{\ell}\|_{L^\infty(\mathbb{S}^d)}
	\lesssim K^{-\frac{2\alpha}{ d+2m+1-2\alpha}},\quad \text{and }\; \|D_{i,j}(\tilde h - g_{\ell})\|_{L^\infty(\mathbb{S}^d)}
	\lesssim K^{-\frac{  2(\alpha-1)}{d+2m+1-2\alpha}}.
	\]
	
We first study Case~III.  Define $\phi  = S_m g_{\ell}$.   Proposition~3.3 of \cite{yang2024optimal} shows  that $\phi \in\mathcal{F}_{\sigma_m}(K)$.  By Lemma~\ref{lem_Sk_property}, we obtain the following approximation bounds for both the function value and its gradient,
	\[
	\|h - \phi\|_{L^\infty(\mathbb{B}^d)}
	\le 2^{m/2}\,\|\tilde h - g_{\ell}\|_{L^\infty(\mathbb{S}^d)}
	\lesssim K^{-\frac{2\alpha}{\,d+2m+1-2\alpha}},
	\]
	and
	\begin{align*}
		\sup_{\bx\in\mathbb{B}^d}	\|	\nabla  (h-\phi)(\bx)\|_{\infty}\lesssim   \| \tilde{h}-g_{\ell}\|_{L^{\infty}(\mathbb{S}^{d})}
		+  \sum_{1\le i<j\le d+1} \|D_{i,j}(\tilde{h}-g_{\ell})\|_{L^{\infty}(\mathbb{S}^{d})}
		\lesssim    K^{-\frac{ 2(\alpha-1)}{d+2m+1-2\alpha}}.
	\end{align*}
	Thus, $\phi \in \cF_{\sigma_m}(K)$ provides a simultaneous approximation of $h$ in both function value  and gradient. Further, Lemma~\ref{lem_shallow_approximation} shows that there exists $\phi_W \in \mathcal{F}_{\sigma_m}(W,K)$ such that
  \begin{align*}
		\big\|  \phi_W-\phi \big\|_{L^\infty(\mathbb{B}^d) }
 	\lesssim K\cdot W^{-\frac{1}{2} - \frac{2m+1}{2d}},  \text{ and }
		\sup_{\|\bs\|_1 \leq 1}
		\big\| D^{\bs}( \phi_W-\phi) \big\|_{L^\infty(\mathbb{B}^d) }
 	\lesssim K\cdot W^{-\frac{1}{2} - \frac{2m-1}{2d}}.
	\end{align*}
	Combining this with the previous bounds yields
	\[
	\big\|\phi_W-h\big\|_{L^\infty(\mathbb{B}^d) }\leq \big\|\phi_W-\phi\big\| _{L^\infty(\mathbb{B}^d) }+\big\|\phi-h\big\|_{L^\infty(\mathbb{B}^d) }\lesssim  K\cdot W^{-\frac{1}{2} - \frac{2m+1}{2d}}+K^{-\frac{ 2\alpha}{d+2m+1-2\alpha}},
	\]
	and
	\begin{align*}
		\sup_{\bx\in \mathbb{B}^d}  \big\|\nabla(\phi_W-h)(\bx) \big\|_{\infty} & \leq \sup_{\bx\in \mathbb{B}^d}  \big\|\nabla(\phi_W-\phi) (\bx)\big\|_{\infty} +\sup_{\bx\in \mathbb{B}^d}  \big\|\nabla(\phi-h)(\bx) \big\|_{\infty}
		\\&\lesssim K\cdot W^{-\frac{1}{2} - \frac{2m-1}{2d}} +  K^{-\frac{ 2(\alpha-1)}{d+2m+1-2\alpha}}.
	\end{align*}
	Therefore, with $K \gtrsim W^{\frac{d+2m+1-2\alpha}{2d}}$, for Case~III there exists  $\phi_W \in \mathcal{F}_{\sigma_m}(W,K)$ such that
	\begin{align*}
		\big\|\phi_W-h\big\|_{L^\infty(\mathbb{B}^d) } \lesssim  W^{-\frac{\alpha}{d}},\quad\text{and} \quad\sup_{\bx\in \mathbb{B}^d}  \big\|\nabla(\phi_W-h) (\bx)\big\|_{\infty}  \lesssim   W^{-\frac{\alpha-1}{d}}.
	\end{align*}
   For Case~II, by a similar argument, there exists $\phi_W \in \mathcal{F}_{\sigma_m}(W,K)$ such that
		\begin{align*}
		\big\|\phi_W-h\big\|_{L^\infty(\mathbb{B}^d) }& \lesssim   K\cdot W^{-\frac{1}{2} - \frac{2m+1}{2d}}+\exp\{-\alpha K^2\}, 	\quad \text{ and }
	\nonumber	\\\sup_{\bx\in \mathbb{B}^d}  \big\|\nabla(\phi_W-h) (\bx)\big\|_{\infty} & \lesssim   K\cdot W^{-\frac{1}{2} - \frac{2m-1}{2d}}+ \exp\{-(\alpha-1) K^2\}.
	\end{align*}
   When $K^2 \gtrsim \frac{d+2m+1}{2d\alpha}\log W$, this yields
			\begin{align*}
		\big\|\phi_W-h\big\|_{L^\infty(\mathbb{B}^d) }   \lesssim  W^{- \frac{d+2m+1}{2d}}\sqrt{\log W}
	  \text{ and }  \sup_{\bx\in \mathbb{B}^d}  \big\|\nabla(\phi_W-h) (\bx)\big\|_{\infty}  \lesssim    W^{- \frac{d+2m-1}{2d}}\sqrt{\log W}.
	\end{align*}
	For Case~I, there exists $\phi_W \in \mathcal{F}_{\sigma_m}(W,C(d,m,\alpha))$ such that
		\begin{align*}
		\big\|\phi_W-h\big\|_{L^\infty(\mathbb{B}^d) }  \lesssim   W^{  - \frac{d+2m+1}{2d}}
		\quad \text{ and }\quad	 \sup_{\bx\in \mathbb{B}^d}  \big\|\nabla(\phi_W-h) (\bx)\big\|_{\infty}  \lesssim     W^{ - \frac{d+2m-1}{2d}} .
	\end{align*}
Consequently, for all three cases, when $K \gtrsim \max\{W^{\frac{d+2m+1-2\alpha}{2d}},\sqrt{\log W} \}$,  there exists   $\phi_W \in \cF_{\sigma_m}(W,K)$ such that
	\begin{align} \label{eq_error1}
	\big\|\phi_W-h\big\|_{L^\infty(\mathbb{B}^d) } & \lesssim   W^{- \frac{\min\{d+2m+1,2\alpha\}}{2d}} \sqrt{\log W},
	\quad \text{ and }\nonumber
	\\\sup_{\bx\in \mathbb{B}^d}  \big\|\nabla(\phi_W-h) (\bx)\big\|_{\infty} & \lesssim     W^{ - \frac{\min\{d+2m-1,2(\alpha-1)\}}{2d}}\sqrt{\log W}.
\end{align}

	Next, we extend the approximation results from   $\mathbb{B}^d$ to   $\mathcal{X} = \{\bx \in \mathbb{R}^d : \|\bx\|_\infty \le 1\}$.  Let $f \in \mathcal{H}^\alpha(\mathcal{X})$ be the target function. Define a rescaled function on $\mathbb{B}^d$ by
	\[
	\tilde{f}(\by):= f(\sqrt{d}\by), \quad \text{ with }  \by=\frac{\bx}{\sqrt{d} } \in \mathbb{B}^d.
	\]
	Then  $\tilde{f}$ inherits the smoothness of $f$, satisfying $d^{-\alpha/2}\tilde{f} \in \mathcal{H}^\alpha(\mathbb{B}^d)$. By \eqref{eq_error1},  there exists   $\phi_W \in \cF_{\sigma_m}(W,K)$ of the form
	\[
	\phi_W(\by) = \sum_{i=1}^W a_i \sigma_m((\by^{\top},1)\bv_i), \quad \text{with } \bv_i \in \mathbb{S}^d \text{ and } \sum_{i=1}^W|a_i| \le K,
	\]
   with $K \gtrsim \max\{W^{\frac{d+2m+1-2\alpha}{2d}},\sqrt{\log W} \}$, such that
 	\begin{align*}
		\big\| \phi_W-d^{-\alpha/2}\tilde{f}\big\|_{L^\infty(\mathbb{B}^d) } &  \lesssim W^{- \frac{\min\{d+2m+1,2\alpha\}}{2d}} \sqrt{\log W},  	\quad \text{ and } 	\\
       \sup_{\by\in \mathbb{B}^d} \big\|\nabla(\phi_W-d^{-\alpha/2}\tilde{f})(\by) \big\|_{\infty} & \lesssim    W^{ - \frac{\min\{d+2m-1,2(\alpha-1)\}}{2d}}\sqrt{\log W} .
	\end{align*}
	Define the approximation function on $\mathcal{X}$ by scaling back as $\tilde{g}(\bx) := \phi_W(\bx/\sqrt{d})$. We show that $\tilde{g}$ remains within $\cF_{\sigma_m}(W, K)$.
	Let $\bv_i = (\bw_i^\top, b_i)^\top$ with $\bw_i \in \mathbb{R}^d$, $b_i \in \mathbb{R}$, and $\|\bw_i\|_2^2+b_i^2 = 1$.  Let $\bv'_i := (\bw_i^\top/\sqrt{d}, b_i)^\top$ and let $c_i := \|\bv'_i\|_2 = \sqrt{\|\bw_i\|_2^2/d + b_i^2}$. Since $\|\bw_i\|_2^2 = 1 - b_i^2$ and $b_i^2 \in [0,1]$, we have $c_i^2 = (1-b_i^2)/d + b_i^2 \le 1$. Using the $m$-homogeneity  of $\sigma_m$ and defining the normalized vector $\bv''_i = \bv'_i / c_i \in \mathbb{S}^d$, we can write $\tilde{g}(\bx)$ as
	\[
	\tilde{g}(\bx) =  \sum_{i=1}^W a_i \sigma_m\left(\left(\frac{\bx^{\top}}{\sqrt{d}}, 1\right)\bv_i\right)=\sum_{i=1}^W (a_i c_i^m) \sigma_m((\bx^\top, 1)\bv''_i).
	\]
	Moreover, $\sum_{i=1}^W |a_i c_i^m| \le \sum_{i=1}^W |a_i| \cdot 1^m \le K$. Thus, $\tilde{g}\in \cF_{\sigma_m}(W, K)$. Let $g:=d^{\alpha/2}\tilde{g}.$   Then,
	\begin{equation}\label{eq_approfun}
	\sup_{\bx \in \mathcal{X}} |g(\bx) -  f(\bx)| \leq \sup_{\by \in \mathbb{B}^d} |d^{\alpha/2} \phi_W(\by) -  \tilde{f}(\by)| \lesssim   W^{- \frac{\min\{d+2m+1,2\alpha\}}{2d}} \sqrt{\log W}.
	\end{equation}
	For the gradient, the chain rule yields $\nabla_{\bx}(g-f)(\bx) = \frac{1}{\sqrt{d}} \nabla_{\by}(d^{\alpha/2}\phi_W - \tilde{f})(\bx/\sqrt{d})$. Taking the supremum norm gives
	\begin{equation}\label{eq_approfundei}
 	\sup_{\bx \in \mathcal{X}} \|\nabla(g-f)(\bx)\|_{\infty} \leq \frac{1}{\sqrt{d}} \sup_{\by \in \mathbb{B}^d} \|\nabla(d^{\alpha/2}\phi_W - \tilde{f})(\by)\|_{\infty} \lesssim  W^{ - \frac{\min\{d+2m-1,2(\alpha-1)\}}{2d}}\sqrt{\log W}.
	\end{equation}
	
By relabeling $W$ in \eqref{eq_approfun} and \eqref{eq_approfundei} as $N$ and combining these results with the argument in the first paragraph, we establish that for any $f \in \mathcal{H}^\alpha(\mathcal{X})$, if $W \ge N$, $\mathcal{S} \ge N(d+1)$, and $K \gtrsim \max\{N^{\frac{d+2m+1-2\alpha}{2d}}, \sqrt{\log N}\}$, there exists $g \in \mathcal{N}\mathcal{N}(W,L,\mathcal{S},K)$ such that
	\begin{align*}
	\sup_{\bx \in \mathcal{X}} |g(\bx) -  f(\bx)| &\lesssim  N^{- \frac{\min\{d+2m+1,2\alpha\}}{2d}} \sqrt{\log N},\quad  \text{ and }
\\	\sup_{\bx \in \mathcal{X}} \|\nabla(g-f)(\bx)\|_{\infty}  & \lesssim N^{ - \frac{\min\{d+2m-1,2(\alpha-1)\}}{2d}}\sqrt{\log N},
\end{align*}
which  completes the proof.
\end{proof}

	\phantomsection
	\addcontentsline{toc}{subsection}{Theorem 4.1}
	\begin{statement}{Theorem 4.1}
			Suppose  Assumption~1  holds with $\|\ell\|_{\Lip}<\infty$.  Assume $m\geq 1$, $W\ge  N,\mathcal{S}\ge N(d+1),$ $ K\!\gtrsim\max \{N^{\frac{d+2m+1-2\alpha}{2d}},\sqrt{\log N} \},$ and $M_n\asymp\log n$. Let $\beta =  \min\{(d+2m+1)/2, \alpha\}$. Then, for any $\delta>0$,  we have
   \begin{equation*}
	\begin{split}
		\Ebb\big[{\cal E}_{\Pbb,1}(\widehat{f}_{1,\delta}^{(n)};\delta) \big]  &	\lesssim   c_1+ c_4^2+ N^{-\frac{2\beta}{d}}\log N+ n^{-1}\mathcal{S}\log(Kn)\log n+ \delta  K
		\\&	\qquad \; +
		\exp\{-\mathcal{S}\log(Kn)\}\log n.
	\end{split}
\end{equation*}
	If further $\mathcal{S} \asymp W \asymp N \asymp n^{\frac{d}{d+2\beta}}$ and $K \asymp \max\big\{n^{\frac{d+2m+1-2\alpha}{2(d+2\beta)}},\sqrt{\log n}\big\}$, then
	 \begin{equation*}
		\Ebb\big[{\cal E}_{\Pbb,1}(\widehat{f}_{1,\delta}^{(n)};\delta) \big]\lesssim c_1+ c_4^2 + n^{-\frac{ 2\beta}{d+2\beta}}\{\log n\}^2+ \delta K.
	\end{equation*}
	\end{statement}

\begin{proof}[{\bf Proof of  Theorem 4.1}]
 Let $f_{1,n}\in\argmin_{f\in \cF_n} \cR_{\Pbb,1}(f;\delta)$ and $f_{1,\delta}^{\star}\in\argmin_{f\in \cH^{\alpha}} \cR_{\Pbb,1}(f;\delta)$. The excess local worst-case risk can be denoted as
 \[
 \cE_{\Pbb,1}(\widehat{f}_{1,\delta}^{(n)};\delta) =\cR_{\Pbb,1}(\widehat{f}_{1,\delta}^{(n)};\delta)- \min\big\{\cR_{\Pbb,1}(f_{1,\delta}^{\star};\delta),\cR_{\Pbb,1}(f_{1,n};\delta) \big\}.
 \]
 For notational convenience, define
 \[
 \mathcal{E}_{1,1}:=\cR_{\Pbb,1}(\widehat{f}_{1,\delta}^{(n)};\delta)-\cR_{\Pbb,1}(f_{1,\delta}^{\star};\delta),\quad  \mathcal{E}_{1,2}:=\cR_{\Pbb,1}(\widehat{f}_{1,\delta}^{(n)};\delta)-\cR_{\Pbb,1}(f_{1,n};\delta).
 \]
Under this notation, the excess local worst-case risk satisfies
 \begin{equation} \label{eq_k1error}
 	\begin{split}
 		{\cal E}_{\Pbb,1}(\widehat{f}_{1,\delta}^{(n)};\delta)   =\max\big\{  \mathcal{E}_{1,1},  \mathcal{E}_{1,2}\big \}.
 	\end{split}
 \end{equation}	
The subsequent analysis will bound $ \mathcal{E}_{1,1}$ and $  \mathcal{E}_{1,2}$ separately.

We begin with $\mathcal{E}_{1,1}$. Fix an arbitrary $f_n \in \mathcal{F}_n$ (to be specified later).  Since $\mathcal{R}_{\mathbb{P}_{n},1}(\widehat{f}_{1,\delta}^{(n)};\delta) \le \mathcal{R}_{\mathbb{P}_n,1}(f_n;\delta)$,  we can decompose $\mathcal{E}_{1,1}$ as follows.
 \begin{align*}
 	\cE_{1,1} &=		\cR_{\Pbb,1}(\widehat{f}_{1,\delta}^{(n)};\delta)-\cR_{\Pbb_n,1}(\widehat{f}_{1,\delta}^{(n)};\delta)+\cR_{\Pbb_n,1}(\widehat{f}_{1,\delta}^{(n)};\delta)-\cR_{\Pbb_n,1}(f_n;\delta)
 		\\&\qquad +\cR_{\Pbb_n,1}(f_n;\delta)-\cR_{\Pbb,1}(f_n;\delta)+\cR_{\Pbb,1}(f_n;\delta)-\cR_{\Pbb,1}(f_{1,\delta}^{\star};\delta)
 	\\&\leq  \cR_{\Pbb,1}(\widehat{f}_{1,\delta}^{(n)};\delta)-\cR_{\Pbb_n,1}(\widehat{f}_{1,\delta}^{(n)};\delta)+\cR_{\Pbb_n,1}(f_n;\delta)-\cR_{\Pbb,1}(f_n;\delta)
 	\\&\qquad+ \cR_{\Pbb,1}(f_n;\delta)- \cR_{\Pbb,1}(f_{1,\delta}^{\star};\delta).
 \end{align*}
By	Lemma 2.3 in the main text, $0\leq \cR_{\Pbb,1}(f;\delta)-\cR_{\Pbb}(f)\leq \delta \|\ell_{f}\|_{\Lip},$  which implies
\begin{align*}
&\big|	\cR_{\Pbb,1}(f;\delta)-\cR_{\Pbb_n,1}(f;\delta)-\big\{ \cR_{\Pbb}(f)-\cR_{\Pbb_n}(f)\big\}\big|\leq \delta \|\ell_{f}\|_{\Lip},
  \text{ and }\\&
 \cR_{\Pbb,1}(f_n;\delta)-\cR_{\Pbb,1}(f_{1,\delta}^{\star};\delta)\leq  \cR_{\Pbb}(f_n)-\cR_{\Pbb}(f_{1,\delta}^{\star})+ \delta \|\ell_{f_n}\|_{\Lip} .
\end{align*}
Moreover, the Lipschitz constant of the composite loss $\ell(\bz;f)$ satisfies
 \begin{align*}
 	\|\ell_{f}\|_{\Lip} \leq \|\ell\|_{\Lip} (\|f\|_{\Lip}+1).
 \end{align*}
Consequently,
  \begin{align*}
 	\cE_{1,1} &\leq  I_{1,1}+ I_{1,2}+ 3\delta\|\ell \|_{\Lip}(K+1),
 \end{align*}
 where the errors $I_{1,1} $ and $ I_{1,2}$ are defined by
  \begin{align*}
I_{1,1}&:= \cR_{\Pbb}(\widehat{f}_{1,\delta}^{(n)}) -\cR_{\Pbb_n}(\widehat{f}_{1,\delta}^{(n)})+\cR_{\Pbb_n}(f_n)-\cR_{\Pbb}(f_n) ,
\\ I_{1,2}&:=\cR_{\Pbb}(f_n)-\cR_{\Pbb}(f_{1,\delta}^{\star}).
\end{align*}
Define $H_n(f;f_0):=\Ebb[\ell(Z;f)-\ell(Z;f_0)]-\frac{1}{n}\sum_{i=1}^n\{  \ell(Z_i;f)-\ell(Z_i;f_0)\}$ for any $f \in \cF_n$, and define the event
 \begin{align*}
 	\cA_n:=\Big\{	\big|H_n(f;f_0)\big| \leq  2 \|\ell\|_{\Lip} M_n^{-1}r_n(\|f-f_0\|_2+r_n) \quad\text{ for all }  f \in \mathcal F_n\Big \},
 \end{align*}
 where $r_n$  satisfies $ \Ebb[ \overline{R}_n(r_n;\cF_n^*) ]= r_n^2/(64M_n)$ with  $\cF_n^*:=\{a(f-f_0):a\in[0,1], f\in\cF_n\}$.
 Lemma~\ref{lemma_conrate} shows that there exist constants $b_1,b_2>0$ such that
 \begin{align*}
 	P\big\{\cA_n\big\}\geq 1-2\exp\big\{-  b_1n \{M_n^{-1}r_n\}^{2}+b_2\log(\log(M_nr_n^{-1}))\big\}.
 \end{align*}
 On the event $\cA_n$, we can bound the two error terms. First, $I_{1,1}$ is upper bounded by
  \begin{align*}
 	I_{1,1}  &= H_n(\widehat{f}_{1,\delta}^{(n)};f_0) -H_n(f_n;f_0)
 	  \leq  2 \|\ell\|_{\Lip} M_n^{-1}r_n(\|\widehat{f}_{1,\delta}^{(n)}-f_0\|_2+\|f_n-f_0\|_2+2r_n)
 	  \\&\leq c_2 \|f_n-f_0\|_2^2+\frac{c_3}{2}\|\widehat{f}_{1,\delta}^{(n)}-f_0\|_{2}^2+a_{1n}M_n^{-2} r_n^2,
 \end{align*}
   where $a_{1n}=(2c_3^{-1}+c_2^{-1}) \|\ell\|_{\Lip}^2 + 4\|\ell\|_{\Lip}M_n$, and the last inequality follows from $2xy\leq  a^{-1}x^2+ay^2$ for any $a>0$.  Second, Assumption 1 in the main text implies an upper bound on the approximation error $I_{1,2}$ as follows.
     \begin{align*}
     I_{1,2}=\cR_{\Pbb}(f_n)-\cR_{\Pbb}(f_{1,\delta}^{\star}) \leq \cR_{\Pbb}(f_n)-\inf_{f\in\cH^{\alpha}}\cR_{\Pbb}(f)\leq c_1+c_2\|f_n-f_0\|_2^2.
   \end{align*}
  We now combine these bounds by applying the condition (ii) from Assumption 1 in the main text. Conditioned on the event $\cA_n$ and assuming $\|\widehat{f}_{1,\delta}^{(n)} -f_0 \|_{2}\geq c_4$, we have
 \begin{equation*}
 	\begin{split}
 		c_3\|\widehat{f}_{1,\delta}^{(n)} -f_0 \|_{2}^2	&\leq\cR_{\Pbb }(\widehat{f}_{1,\delta}^{(n)})-   \cR_{\Pbb }(f_0) \leq   \cR_{\Pbb,1}(\widehat{f}_{1,\delta}^{(n)};\delta)-\cR_{\Pbb,1}(f_{1,\delta}^{\star};\delta) +(d+1)\delta \|\ell\|_{\Lip}
 		\\&\leq I_{1,1}+I_{1,2}+(3K+d+4)\delta \|\ell\|_{\Lip}
 		\\
 		&\lesssim c_1+2c_2 \|f_n-f_0\|_2^2+\frac{c_3}{2}\|\widehat{f}_{1,\delta}^{(n)}-f_0\|_{2}^2+a_{1n}M_n^{-2} r_n^2+ \delta \|\ell\|_{\Lip}K.
 	\end{split}
 \end{equation*}
 Rearranging terms and choosing $f_n$ as the best approximation of $f_0$ in $\mathcal{F}_n$ leads that
 \begin{equation*}
 c_3	\|\widehat{f}_{1,\delta}^{(n)}-f_0\|_2^2\lesssim c_1+ c_2\inf_{f\in\cF_n}\|f-f_0\|_{\infty}^2+a_{1n}M_n^{-2}r_n^2+ \delta \|\ell\|_{\Lip}K.
 \end{equation*}
This bound holds for the case $\|\widehat{f}_{1,\delta}^{(n)} -f_0 \|_{2}\geq c_4$. To obtain a general bound valid on  $\cA_n$, we must also account for the case where the estimator is within the local region  $\|\widehat{f}_{1,\delta}^{(n)} -f_0\|_{2} < c_4$, for which the $L^2$ error is directly bounded by $c_4^2$. Combining the two cases, on the event $\cA_n$, we show
 \begin{equation*}
	\begin{split}
		\cE_{1,1}	\lesssim   c_1+ c_4^2+c_2\inf_{f \in \mathcal{F}_n}  \|f-f_0\|_{\infty}^2 +a_{1n}M_n^{-2} r_n^2+ \delta \|\ell\|_{\Lip}K.
	\end{split}
	\end{equation*}

 Next, we study  $\mathcal{E}_{1,2}.$ Following a similar analysis and on the event $\cA_n$, we obtain
  \begin{equation*}
 	\begin{split}
 		\mathcal{E}_{1,2}&\leq \cR_{\Pbb,1}(\widehat{f}_{1,\delta}^{(n)};\delta)-\cR_{\Pbb_n,1}(\widehat{f}_{1,\delta}^{(n)};\delta) +\cR_{\Pbb_n,1}(f_{1,n};\delta)-\cR_{\Pbb,1}(f_{1,n};\delta)
 		\\&\leq \cR_{\Pbb}(\widehat{f}_{1,\delta}^{(n)}) -\cR_{\Pbb_n}(\widehat{f}_{1,\delta}^{(n)})+\cR_{\Pbb_n}(f_{1,n})-\cR_{\Pbb}(f_{1,n}) +2\delta\|\ell\|_{\Lip}(K+1)
 		  \\&\leq c_2 \|f_{1,n}-f_0\|_2^2+\frac{c_3}{2}\|\widehat{f}_{1,\delta}^{(n)}-f_0\|_{2}^2+a_{1n}M_n^{-2} r_n^2+2\delta\|\ell\|_{\Lip}(K+1).
 	\end{split}
 \end{equation*}
 Consider the case $\min\{\|f_{1,n}-f_0\|_2, \|\widehat{f}_{1,\delta}^{(n)}-f_0\|_{2}\}\geq c_4$. Since $f_{1,n}\in\argmin_{f\in \cF_n} \cR_{\Pbb,1}(f;\delta)$,  Assumption~1 in the main text implies that, for any $f_n \in \cF_n$, \begin{align*}
  c_{3}\|f_{1,n}-f_0\|_2^2&\leq   \cR_{\Pbb}(f_{1,n})-\cR_{\Pbb}(f_0)
     \leq  \cR_{\Pbb,1}(f_{n};\delta)-\cR_{\Pbb}(f_0)
       \\&\leq  \cR_{\Pbb}(f_{n})-\inf_{f\in\cH^{\alpha}}\cR_{\Pbb}(f)+ \delta\|\ell\|_{\Lip}(K+1).
 \end{align*}
By choosing $f_n$ as the best approximant to $f_0$ over $\cF_n$ and applying the previously derived bound for $\|\widehat{f}_{1,\delta}^{(n)}-f_0\|_{2}$, we  obtain
   \begin{equation*}
         	\begin{split}
         		\cE_{1,2}
         		\lesssim   c_1+ c_2\inf_{f \in \mathcal{F}_n}  \|f-f_0\|_{\infty}^2 +a_{1n}M_n^{-2} r_n^2+ \delta \|\ell\|_{\Lip}K.
         	\end{split}
   \end{equation*}
 Furthermore, by considering the complementary cases where one or both   terms lie within the local region (where the $L^2$ error is bounded by $c_4^2$), we derive the following general bound for $\cE_{1,2}$ on the event $\cA_n$.
      \begin{equation*}
   	\begin{split}
   		\cE_{1,2}
   		\lesssim   c_1+ c_4^2+c_2\inf_{f \in \mathcal{F}_n}  \|f-f_0\|_{\infty}^2 +a_{1n}M_n^{-2} r_n^2+ \delta \|\ell\|_{\Lip}K.
   	\end{split}
   	\end{equation*}

From \eqref{eq_k1error}, the bound on ${\cal E}_{\Pbb,1}(\widehat{f}_{1,\delta}^{(n)};\delta)$  is determined by the bounds on $\cE_{1,1}$ and $\cE_{1,2}$, which implies that under the event $\cA_n$,
      \begin{equation*}
   	\begin{split}
      {\cal E}_{\Pbb,1}(\widehat{f}_{1,\delta}^{(n)};\delta)
   		\lesssim   c_1+c_4^2+ c_2\inf_{f \in \mathcal{F}_n}  \|f-f_0\|_{\infty}^2 +a_{1n}M_n^{-2} r_n^2+ \delta \|\ell\|_{\Lip}K.
   	\end{split}
   \end{equation*}
 By Lemma  \ref{lem_rnn} and Theorem 3.4 in the main text, for any  $N>0$, if
 \[
 W\ge  N,\;\mathcal{S}\ge N(d+1),\; \text{ and } \; K\!\gtrsim\max\big\{N^{\frac{d+2m+1-2\alpha}{2d}},\sqrt{\log N}\big\},
 \]
 then the following bounds hold.
 \[
 r_n \lesssim M_n n^{-1/2}\sqrt{\mathcal{S}\log(Kn)},
 \quad \text{and} \quad
 \inf_{f \in \mathcal{F}_n} \|f - f_0\|_{\infty}^2 \lesssim N^{-\frac{\min\{d+2m+1,2\alpha\}}{d}}\log N.
 \]
In this case, there exists positive constant $C$ such that $P \{\cA_n \}\geq 1- C\exp\{-\mathcal{S}\log(Kn)\}.$ Furthermore, it can be shown that ${\cal E}_{\Pbb,1}(\widehat{f}_{1,\delta}^{(n)};\delta)\lesssim M_n\|\ell\|_{\Lip}.$ Therefore,
   \begin{equation*}
 	\begin{split}
 		\Ebb\big[{\cal E}_{\Pbb,1}(\widehat{f}_{1,\delta}^{(n)};\delta) \big] &= 		\Ebb\big[{\cal E}_{\Pbb,1}(\widehat{f}_{1,\delta}^{(n)};\delta) 1_{\cA_n}\big] +	\Ebb\big[{\cal E}_{\Pbb,1}(\widehat{f}_{1,\delta}^{(n)};\delta) 1_{\cA_n^{c}}\big]
 		\\&	\lesssim   c_1+c_4^2+ c_2\inf_{f \in \mathcal{F}_n}  \|f-f_0\|_{\infty}^2 +a_{1n}M_n^{-2} r_n^2+ \delta \|\ell\|_{\Lip}K+ M_n\|\ell\|_{\Lip} P\big\{\cA_n^{c}\big\}
 		 		\\&	\lesssim   c_1+c_4^2+  c_2N^{-\frac{\min\{d+2m+1,2\alpha\}}{d}}\log N+(c_2^{-1}+c_{3}^{-1}+M_n)\|\ell\|_{\Lip}^2 n^{-1}\mathcal{S}\log(Kn)
 		 			\\&	\qquad\; + \delta \|\ell\|_{\Lip} K + M_n\|\ell\|_{\Lip}\exp\{-\mathcal{S}\log(Kn)\}.
 	\end{split}
 \end{equation*}
 Set $M_n \asymp \log n$. To optimize the bound, choose the network architecture and tuning parameters as follows. Let $\mathcal{S} \asymp W \asymp N$ and $K \asymp \max\big\{N^{\frac{d+2m+1-2\alpha}{2d}},\sqrt{\log N}\big\}$. Balancing the  errors yields  $N\asymp n^{\frac{d}{d+2\beta}}$, where $\beta = \min\{(d+2m+1)/2, \alpha\}$. Thus, $\mathcal{S} \asymp W \asymp n^{\frac{d}{d+2\beta}}$ and $K \asymp \max\big\{n^{\frac{d+2m+1-2\alpha}{2(d+2\beta)}},\sqrt{\log n}\big\}$. Substituting these into the bound implies
 \begin{equation*}
 \Ebb\big[{\cal E}_{\Pbb,1}(\widehat{f}_{1,\delta}^{(n)};\delta) \big]\lesssim c_1 +c_4^2+ n^{-\frac{ 2\beta}{d+2\beta}}\{\log n\}^2 + \delta K.
 \end{equation*}
 Finally, to ensure that the penalty term $\delta K$ does not dominate the   rate and vanishes asymptotically, we select $\delta$ to be sufficiently small. A specific choice $\delta  = o(n^{- \frac{d+2m+1+2\beta}{2(d+2\beta)} })$ ensures the optimal  result. This completes the proof.
	\end{proof}
	
	\phantomsection
	\addcontentsline{toc}{subsection}{Theorem 4.3}
  \begin{statement}{Theorem 4.3}
	Suppose Assumption~1 holds with  $\|\ell^{'}\|_{\infty}, \|\ell^{'}\|_{\Lip}<\infty$.   Assume  $\alpha>1$, $m\geq 2$, $W\ge  N,\mathcal{S}\ge N(d+1),$ $ K\!\gtrsim\max \{N^{\frac{d+2m+1-2\alpha}{2d}},\sqrt{\log N} \},$ and $M_n\asymp\log n$. Let  $\beta = \min\{(d+2m+1)/2, \alpha\}$. Then, for any $\delta>0$,  we have
\begin{equation*}
	\begin{split}
		\Ebb[{\cal E}_{\Pbb,k}(\widehat{f}_{k,\delta}^{(n)};\delta)  ]
		& \lesssim  c_1+c_4^2+   N^{-\frac{2\beta}{d}}\log N
		+     n^{-1}\mathcal{S}\log(Kn) \log n\nonumber\\
		&\qquad	
		+\delta \big\{1+N^{-\frac{\beta-1}{d}}\sqrt{\log N}+ \{n^{-1}\mathcal{S}\}^{1/(2k^\star)} \{L  m^L    \}^{1/k^\star}K\big\}
		\\  &\qquad   + \delta^{\min\{k,2\}}(K^2+Lm^{L}K)
		+ \exp\{-\mathcal{S}\log(Kn)\}\log n.
	\end{split}
\end{equation*}
If further $\mathcal{S} \asymp W \asymp N \asymp n^{\frac{d}{d+2\beta}}$ and $K \asymp \max\big\{n^{\frac{d+2m+1-2\alpha}{2(d+2\beta)}},\sqrt{\log n}\big\}$, then
\begin{align*}
	\Ebb\big[{\cal E}_{\Pbb,k}(\widehat{f}_{k,\delta}^{(n)};\delta) \big]&\lesssim
	c_1+c_4^2+ n^{-\frac{2\beta}{d+2\beta}} \{\log n\}^2
	+ \delta  \{1
	+ n^{-\frac{    \beta}{k^\star(d+2\beta)}} K \}
	+ \delta^{\min\{k,2\}} K^2.
\end{align*}
	\end{statement}
	\begin{proof}[{\bf Proof of Theorem 4.3}]
		We first consider the case $k \geq 2$, while the analysis for $1 < k < 2$ follows similarly.
	Let $f_{k,n}\in\argmin_{f\in \cF_n} \cR_{\Pbb,k}(f;\delta)$ and   $f_{k,\delta}^{\star}\in\argmin_{f\in \cH^{\alpha}} \cR_{\Pbb,k}(f;\delta)$. The excess local worst-case risk can be expressed as
		\[
		\cE_{\Pbb,k}(\widehat{f}_{k,\delta}^{(n)};\delta) =\cR_{\Pbb,k}(\widehat{f}_{k,\delta}^{(n)};\delta)- \min\big\{\cR_{\Pbb,k}(f_{k,\delta}^{\star};\delta),\cR_{\Pbb,k}(f_{k,n};\delta) \big\}.
		\]
   	For notational convenience, define
		\[
		\cE_{k,1}:=\cR_{\Pbb,k}(\widehat{f}_{k,\delta}^{(n)};\delta)-\cR_{\Pbb,k}(f_{k,\delta}^{\star};\delta),\quad \cE_{k,2}:=\cR_{\Pbb,k}(\widehat{f}_{k,\delta}^{(n)};\delta)-\cR_{\Pbb,k}(f_{k,n};\delta).
		\]
		Under this notation, the excess local worst-case risk satisfies
		\begin{equation*}
			\begin{split}
				{\cal E}_{\Pbb,k}(\widehat{f}_{k,\delta}^{(n)};\delta)   =\max\big\{ \cE_{k,1},\cE_{k,2}\big \}.
			\end{split}
		\end{equation*}	
	The subsequent analysis will bound $ \mathcal{E}_{k,1}$ and $  \mathcal{E}_{k,2}$ separately.

	We begin with $\mathcal{E}_{k,1}$. Fix an arbitrary $f_n \in \mathcal{F}_n$ (to be specified later).  Since $\mathcal{R}_{\mathbb{P}_{n},k}(\widehat{f}_{k,\delta}^{(n)};\delta) \le \mathcal{R}_{\mathbb{P}_n,k}(f_n;\delta)$,  we can decompose $\mathcal{E}_{k,1}$ as follows.
			\begin{align*}
			\cE_{k,1}&=		 \cR_{\Pbb,k}(\widehat{f}_{k,\delta}^{(n)};\delta)-\cR_{\Pbb_n,k}(\widehat{f}_{k,\delta}^{(n)};\delta)+\cR_{\Pbb_n,k}(\widehat{f}_{k,\delta}^{(n)};\delta)-\cR_{\Pbb_n,k}(f_n;\delta)
				\\&\qquad +\cR_{\Pbb_n,k}(f_n;\delta)-\cR_{\Pbb,k}(f_n;\delta)+\cR_{\Pbb,k}(f_n;\delta)-\cR_{\Pbb,k}(f_{k,\delta}^{\star};\delta)
				\\&\leq \big\{\cR_{\Pbb,k}(\widehat{f}_{k,\delta}^{(n)};\delta)-\cR_{\Pbb_n,k}(\widehat{f}_{k,\delta}^{(n)};\delta)+\cR_{\Pbb_n,k}(f_n;\delta)-\cR_{\Pbb,k}(f_n;\delta)	\big\}
				\\&\qquad+ \big\{\cR_{\Pbb,k}(f_n;\delta)- \cR_{\Pbb,k}(f_{k,\delta}^{\star};\delta)\big\}
				\\&:=I_{k,1}+I_{k,2}.
			\end{align*}
			To analyze $I_{k,1}$, note that for losses of the form $\ell(\bz;f)=\ell(f(\bx)-y)$ and any differentiable functions $f,\widetilde{f}$, we have
			\begin{align*}
				&  \big\| \nabla_{\bz}\ell(f(\bx)-y)-\nabla_{\bz}\ell(\widetilde f(\bx)-y)\big\|_1
				= \Big\|\ell'(f(\bx)-y)\begin{pmatrix}\nabla f(\bx)\\ -1\end{pmatrix}
				-\ell'(\widetilde f(\bx)-y)\begin{pmatrix}\nabla \widetilde f(\bx)\\ -1\end{pmatrix}\Big\|_1\\
				&\leq \big|\ell'(f(\bx)-y)-\ell'(\widetilde f(\bx)-y)\big|\;\Big\|\begin{pmatrix}\nabla f(\bx)\\ -1\end{pmatrix}\Big\|_1
				+ |\ell'(\widetilde f(\bx)-y)|\;\|\nabla f(\bx)-\nabla\widetilde f(\bx)\|_1\\
				&\leq \|\ell'\|_{\Lip}\,|f(\bx)-\widetilde f(\bx)|\;\big(\|\nabla f(\bx)\|_1+1\big)
				+ \|\ell'\|_\infty \|\nabla f(\bx)-\nabla\widetilde{f}(\bx)\|_1.
			\end{align*}
		Since $\sup_{f\in\cH^{\alpha}}\sup_{\bx\in \cX}\|\nabla f(\bx)\|_1\leq d$, it follows that for any $f \in\cH^{\alpha}$,  we have
			\begin{align}\label{eq_qqq}
				\big\| \nabla_{\bz}\ell(f(\bx)-y)-\nabla_{\bz}\ell(\widetilde f(\bx)-y)\big\|_1
				\leq C_0 \big\{|f(\bx)-\widetilde f(\bx)|+\|\nabla f(\bx)-\nabla\widetilde f(\bx)\|_1\big \},
			\end{align}
			where   $C_0=\max\{(d+1)\|\ell'\|_{\Lip},  \|\ell'\|_\infty\}$.
			By a similar argument and the bound $\|\nabla f(\bx)-\nabla f(\tilde \bx)\|_1\le d\sup_{\bx'\in\cX}\|\nabla^2 f(\bx')\|_{\infty}\|\bx-\tilde \bx\|_{\infty}$, we obtain
				\begin{align*}
				 	\big\|\nabla_{\bz}\ell(f(\bx)-y)-\nabla_{\bz}\ell(f(\tilde{\bx})-\tilde{y})\big\|_{1}
			 \leq \cD(f) (\|\bx-\tilde \bx\|_{\infty}+|y-\tilde{y}|),
			\end{align*}
			where $\cD(f)=\|\ell'\|_{\Lip}( \sup_{\bx\in\cX}\|\nabla f(\bx)\|_{1}+1)^2+ d\|\ell'\|_{\infty} \sup_{\bx\in\cX}\|\nabla^2 f(\bx)\|_{\infty}$. Define
			\[
			\mathcal {V}(f)  = \mathbb E\big[\|\nabla\ell(Z;f)\|_1^{k^\star}\big], \qquad
			\mathcal {V}_n(f)  = \frac{1}{n}\sum_{i=1}^n \|\nabla\ell(Z_i;f)\|_1^{k^\star}.
			\]
			 By  Lemma~2.5 in the main text,
			\begin{align}\label{eq:linearization}
				\big|\mathcal R_{\mathbb P,k}(f;\delta)-\mathcal R_{\mathbb P}(f)-\delta \mathcal V(f)^{1/k^\star}\big|
				&\le \delta^2 \mathcal{D}(f) , \text{ and}
				\nonumber	\\	\big|\mathcal R_{\mathbb {P}_n,k}(f;\delta)-\mathcal R_{\mathbb {P}_n}(f)-\delta \mathcal{V}_n(f)^{1/k^\star}\big|
				&\le \delta^2 \mathcal{D}(f).
			\end{align}
			Moreover, Lemma \ref{lem_gradient} ensures uniform boundedness
			\begin{align} \label{eq_upperD}
				\sup_{f\in\cH^{\alpha}}\mathcal{D}(f)\lesssim  d^2,\quad \text{ and }\quad
				\sup_{f\in\cF_n }\mathcal{D}(f)\lesssim K^2+Lm^{L}K.
			\end{align}
	Combining \eqref{eq:linearization} with Lemma \ref{lem_ab}, and noting that $1/k^{\star}<1$, we obtain
			\begin{align*}
				&\mathcal R_{\mathbb P,k}(f;\delta)-\mathcal R_{\mathbb P_n,k}(f;\delta)\nonumber\\
				&\le \mathcal R_{\mathbb P}(f)-\mathcal R_{\mathbb P_n}(f) + \delta \big\{\mathcal V(f)^{1/k^\star}-\mathcal V_n(f)^{1/k^\star}\big\}+2\delta^2\mathcal D(f)\nonumber\\
				&\le H_n(f;f_0) + \mathcal R_{\mathbb P}(f_0)-\mathcal R_{\mathbb P_n}(f_0)
				+ \delta \big|\mathcal V(f)-\mathcal V_n(f)\big|^{\,1/k^\star} + 2\delta^2\mathcal{D}(f),
			\end{align*}
			where $H_n(f;f_0): =\mathbb E[\ell(Z;f)-\ell(Z;f_0)]-\frac{1}{n}\sum_{i=1}^n\big \{\ell(Z_i;f)-\ell(Z_i;f_0)\big\}.$
			Define the event
			\begin{align*}
			 \cA_n:=\Big\{	\big|H_n(f;f_0)\big| \leq  2 \|\ell^{\prime}\|_{\infty} M_n^{-1}r_n(\|f-f_0\|_2+r_n) \;\text{ for all }  f \in \mathcal F_n\Big \},
			\end{align*}
			where $r_n$ satisfies  $ \Ebb[ \overline{R}_n(r_n;\cF_n^*) ]= r_n^2/(64M_n)$  with  $\cF_n^*:=\{a(f-f_0):a\in[0,1], f\in\cF_n\}$.   Lemma~\ref{lemma_conrate} shows that there exist constants $b_1,b_2>0$ such that
			\begin{align*}
				P\big\{\cA_n\big\}\geq 1-2\exp\big\{-  b_1n \{M_n^{-1}r_n\}^{2}+b_2\log(\log(M_nr_n^{-1}))\big\}.
			\end{align*}
		   Consequently, on the event $\cA_n$, we have
			\begin{align} \label{eq_sto_k}
				I_{k,1}&= \cR_{\Pbb,k}(\widehat{f}_{k,\delta}^{(n)};\delta)-\cR_{\Pbb_n,k}(\widehat{f}_{k,\delta}^{(n)};\delta)+\cR_{\Pbb_n,k}(f_n;\delta)-\cR_{\Pbb,k}(f_n;\delta)	\nonumber\\
				&\leq H_n(\widehat{f}_{k,\delta}^{(n)};f_0) - H_n(f_n;f_0)
				+ 2\delta \sup_{f\in\cF_n} \big|\mathcal V(f)-\mathcal V_n(f)\big|^{\,1/k^\star} + 4\delta^2 \sup_{f\in\cF_n}\mathcal D(f)
				\nonumber\\
				&\leq 2\|\ell^{\prime}\|_{\infty} M_n^{-1}r_n(\|\widehat{f}_{k,\delta}^{(n)}-f_0\|_2+\|f_n-f_0\|_2 +2r_n)
				\nonumber\\&\qquad + 2\delta \sup_{f\in\cF_n} \big|\mathcal V(f)-\mathcal V_n(f)\big|^{\,1/k^\star} +4\delta^2 \sup_{f\in\cF_n}\mathcal {D}(f).
			\end{align}
		    To bound $|\mathcal V(f)-\mathcal V_n(f) |^{ 1/k^\star}$, we define $g_f(\bz):=\|\nabla\ell(\bz;f)\|_1^{k^\star}$.
			By symmetrization argument,  it follows
			\[
			\mathbb E\sup_{f\in\mathcal F_n}\big|\mathcal V(f)-\mathcal V_n(f)\big|=	\mathbb E\sup_{f\in\mathcal F_n}\Big|\mathcal \mathbb E[g_f(Z)] - \frac{1}{n}\sum_{i=1}^n g_f(Z_i)\Big|
			\le 2\,\mathbb E\sup_{f\in\mathcal F_n}\Big|\frac{1}{n}\sum_{i=1}^n \sigma_i g_f(Z_i)\Big|,
			\]
			where $\{\sigma_i\}_{i=1}^n$ are i.i.d.  Rademacher variables independent of the data. By \eqref{eq_qqq} and the mean-value theorem on the map $t\mapsto t^{k^\star}$, for any $f,\tilde f\in\cF_n$,
			\begin{align*}
				\big|g_f(\bz)-g_{\widetilde{f}}(\bz)\big| &\leq k^\star \sup_{\bz\in\cZ}\sup_{f\in\cF_n}\|\nabla\ell(\bz;f)\|_1 ^{\,k^\star-1}
				\cdot \big\|\nabla\ell(\bz;f)-\nabla\ell(\bz;\widetilde{f})\big\|_1
				\\&
				\leq k^\star C_1^{\,k^\star-1} C_0\big\{|f(\bx)-\widetilde f(\bx)|+\|\nabla f(\bx)-\nabla\widetilde f(\bx)\|_1\big \},
			\end{align*}
			where $C_1 :=\|\ell^{'}\|_\infty (\sup_{f\in\mathcal F_n}\sup_{\bx\in\cX}\|\nabla f(\bx)\|_1+1 ) \leq \|\ell^{'}\|_\infty (K+1 ).$
			Let $\cF_n$ be equipped with the metric $\rho_\infty(f,\widetilde f)  =\sup_{\bx\in\cX}\big\{|f(\bx)-\widetilde f(\bx)|+\|\nabla f(\bx)-\nabla\widetilde f(\bx)\|_1\big\}.$ Then it follows
			\begin{align*}
				\sup_{\bz\in\cZ}\big|g_f(\bz)-g_{\widetilde{f}}(\bz)\big|
				\leq k^\star C_1^{\,k^\star-1} C_0 \rho_\infty(f,\widetilde f).
			\end{align*}
		 According to Theorem 3.3 in the main text,
		  	\begin{equation*}
		 	\log \cN(u, \cN\cN(W,L,\mathcal{S},K), \rho_\infty) \lesssim \mathcal{S}\log\left(1+ LK m^Lu^{-1} \right).
		 \end{equation*}
			Applying  Lemma \ref{lem_cover} and noting that $\sup_{f\in\cF_n}\|g_f\|_{\infty}\leq C_1^{k^*}$ , we obtain
			\begin{equation*}
				\begin{split}
					\mathbb{E}\sup_{f\in\mathcal F_n}\Big|\frac{1}{n}\sum_{i=1}^n \sigma_i g_f(Z_i)\Big|
					&\leq \inf_{\eta\geq 0}\bigg\{
					4\eta+12\int_{\eta}^{C_1^{k^*}}\sqrt{\frac{\log\mathcal{N}\big(u/(k^\star C_1^{\,k^\star-1} C_0),\mathcal F_n,  \rho_\infty\big)}{n}}du
					\bigg\}
					\\&\lesssim  \inf_{\eta\geq 0}\bigg\{
					4\eta+12\int_{\eta}^{C_1^{k^*}}\sqrt{\frac{\mathcal{S}\log \big(1+LK m^Lk^\star C_1^{\,k^\star-1} C_0u^{-1}\big)}{n}}du
					\bigg\}
					\\&\lesssim n^{-1/2}\sqrt{\mathcal{S}}\int_{0}^{C_1^{k^*}}\sqrt{\log \big(1+LK m^Lk^\star C_1^{\,k^\star-1} C_0u^{-1}\big)}du
					\\&\lesssim n^{-1/2}\sqrt{\mathcal{S}}  LK m^L k^\star  C_1^{k^*-1}C_0.
				\end{split}
			\end{equation*}
		 This implies that
			\begin{align}\label{eq_VV}
				\Ebb\big[\sup_{f\in\cF_n}\big|\mathcal V(f)-\mathcal V_n(f)\big|^{1/k^\star}\big]
				&\leq \Ebb\big[\sup_{f\in\cF_n}\big|\mathcal V(f)-\mathcal V_n(f)\big|\big]^{1/k^\star} \nonumber
			\\&	\leq 2^{1/k^\star} \Big\{\mathbb E\sup_{f\in\mathcal F_n}\big|\frac{1}{n}\sum_{i=1}^n \sigma_i g_f(Z_i)\big|\Big\}^{1/k^\star}\nonumber
				\\&\lesssim n^{-1/(2k^\star)}\mathcal{S}^{1/(2k^\star)} \{LK m^L k^\star  C_1^{k^*-1} \}^{1/k^\star}.
			\end{align}

            Next, we investigate the approximation error term $I_{k,2}$. Applying \eqref{eq:linearization} together with the triangle inequality for the norm yields
			\begin{align*}
				I_{k,2}&=\mathcal R_{\mathbb P,k}(f_n;\delta)-\mathcal R_{\mathbb P,k}(f_{k,\delta}^\star;\delta) \nonumber\\
				&\le \mathcal R_{\mathbb P}(f_n)-\mathcal R_{\mathbb P}(f_{k,\delta}^\star)
				+ \delta\big\{  \mathcal V(f_n)^{1/k^\star}-\mathcal V(f_{k,\delta}^\star)^{1/k^\star}  \big\}
				+ \delta^2\big\{\mathcal D(f_n)+\mathcal D(f_{k,\delta}^\star)\big\}\nonumber\\
				&\le \mathcal R_{\mathbb P}(f_n)- \inf_{f\in\cH^{\alpha}}\cR_{\Pbb}(f)
				+ \delta\,\mathbb E\Big[ \big|\|\nabla\ell(Z;f_n) -\nabla\ell(Z;f_{0})\|_1\big|^{k^\star}\Big]^{1/k^\star} \nonumber\\
				&\quad + \delta\,\mathbb E\Big[ \big|\|\nabla\ell(Z;f_0) - \nabla\ell(Z;f_{k,\delta}^\star)\|_1\big|^{k^\star}\Big]^{1/k^\star} + \delta^2\big\{ \mathcal D(f_n)+\mathcal D(f_{k,\delta}^\star) \big\}.
			\end{align*}
			 By \eqref{eq_qqq}, we have
			\begin{align*}
				&	\mathbb E\Big[ \big|\|\nabla\ell(Z;f_n)-\nabla\ell(Z;f_{0})\|_1\big|^{k^\star}\Big]^{1/k^\star}\nonumber
				\\&	\leq C_0  \; \Big\{ \mathbb E\big[|f_n(X)-f_{0}(X)|^{k^\star}\big]^{1/k^\star}
				+ \mathbb E\big[\|\nabla(f_n-f_{0})(X)\|_1^{k^\star}\big]^{1/k^\star}\Big\}.
			\end{align*}
			Moreover, it holds that $\sup_{f\in\cH^{\alpha}}\sup_{\bz\in\cZ}\|\nabla\ell(\bz;f)\|_1\leq \|\ell^{'}\|_{\infty}(d+1).$
			Therefore, we show
			\begin{align}\label{eq_approxm}
				I_{k,2 }	& \leq \mathcal R_{\mathbb P}(f_n)-\inf_{f\in\cH^{\alpha}}\cR_{\Pbb}(f)
				+ \delta\,  C_0 \big\{ \|f_n-f_{0}\|_{k^*}
				+ \big\|\|\nabla(f_n-f_{0})\|_1\big  \|_{k^\star} \big\}
				\nonumber\\&\qquad +   4\delta d\,\|\ell^{'}\|_{\infty} + \delta^2\big\{ \sup_{f\in\cF_n}\mathcal D(f)+\sup_{f\in\cH^{\alpha}}\mathcal D(f )\big\}.
			\end{align}

			Combining \eqref{eq_sto_k} and \eqref{eq_approxm}, and under  Assumption 1 in the main text, we obtain
			\begin{align*}
				\cE_{k,1}&=\cR_{\Pbb,k}(\widehat{f}_{k,\delta}^{(n)};\delta)-\cR_{\Pbb,k}(f_{k,\delta}^{\star};\delta)\leq I_{k,1}+I_{k,2}
				\\& \leq  c_1  +\frac{c_3}{2}\|\widehat{f}_{k,\delta}^{(n)}-f_0\|_2^2+2c_2\|f_n-f_{0}\|_{2}^2
				+ \delta  C_0  \|f_n-f_{0}\|_{k^{\star}}
				+ \delta  C_0 \big\|\|\nabla(f_n-f_{0})\|_1 \big \|_{k^{\star}}  \nonumber\\
				&\qquad \; +2\delta \sup_{f\in\cF_n} \big|\mathcal V(f)-\mathcal V_n(f)\big|^{\,1/k^\star}+ 4\delta d\|\ell^{'}\|_{\infty}+5\delta^2 \sup_{f\in\cF_n}\mathcal D(f)
					\\&\qquad \;+ \delta^2\sup_{f\in\cH^{\alpha}} \mathcal D(f)
			+a_{1n} M_n^{-2}r_n^2,
			\end{align*}
			where $a_{1n}=(2c_3^{-1}+c_2^{-1})\|\ell^{'}\|_{\infty}^2 +  4\|\ell^{'}\|_{\infty}M_n.$
	    	Assuming $\|\widehat{f}_{k,\delta}^{(n)}-f_0\|_2\geq c_4$, condition (ii) of Assumption 1 in the main text implies that
			\begin{align*}
				c_3\|\widehat{f}_{k,\delta}^{(n)}-f_0\|_2^2	&\leq \cR_{\Pbb}(\widehat{f}_{k,\delta}^{(n)}) -	\cR_{\Pbb}(f_0) \\&\leq 	\cR_{\Pbb,k}(\widehat{f}_{k,\delta}^{(n)};\delta)- \cR_{\Pbb,k}(f_0;\delta) +\sup_{f\in\cH^{\alpha}}\{ \delta\mathcal V(f)^{1/k^\star}+\delta^2\cD(f)\}\\
			   &\leq  	\cE_{k,1}+\sup_{f\in\cH^{\alpha}}\{ \delta\mathcal V(f)^{1/k^\star}+\delta^2\cD(f)\}.
			\end{align*}
		 Substituting the   bound for $\cE_{k,1}$ into this inequality leads to the following upper bound.
				\begin{align*}
				\frac{c_3}{2}\|\widehat{f}_{k,\delta}^{(n)}-f_0\|_2^2	&\leq c_1   +2c_2\|f_n-f_{0}\|_{2}^2
				+ \delta  C_0  \|f_n-f_{0}\|_{k^{\star}}
				+ \delta  C_0 \big\|\|\nabla(f_n-f_{0})\|_1 \big \|_{k^{\star}}  \nonumber\\
				&\qquad  \;+2\delta \sup_{f\in\cF_n} \big|\mathcal V(f)-\mathcal V_n(f)\big|^{\,1/k^\star}+ 6\delta d\|\ell^{'}\|_{\infty}+5\delta^2 \sup_{f\in\cF_n}\mathcal D(f)
				\\&\qquad \;+ 2\delta^2\sup_{f\in\cH^{\alpha}} \mathcal D(f) 	+a_{1n} M_n^{-2}r_n^2.
			\end{align*}
			In the complementary case, $\|\widehat{f}_{k,\delta}^{(n)}-f_0\|_2 < c_4$, its $L^2$  error is  then bounded by $c_4^2$.  Combining these results, choosing $f_n$ as the best approximation of $f_0$ in $\mathcal{F}_n$, and incorporating the $c_4^2$ term for the local case, we establish that on the event $\mathcal{A}_n$,
			\begin{align*}
		     \cE_{k,1}
				&\lesssim   c_1+c_4^2+\inf_{f\in\cF_n} \Big\{ c_2\|f-f_{0}\|_{2}^2
				+ \delta  C_0   \|f-f_{0}\|_{k^{\star}}
				+ \delta  C_0  \big\|\|\nabla(f-f_{0})\|_1 \big \|_{k^{\star}} \Big\} \nonumber\\
				&\qquad  \; + \delta \sup_{f\in\cF_n} \big|\mathcal V(f)-\mathcal V_n(f)\big|^{\,1/k^\star}+  \delta d\|\ell^{'}\|_{\infty}+ \delta^2 \sup_{f\in\cF_n}\mathcal D(f)
				\\  &\qquad \;+  \delta^2\sup_{f\in\cH^{\alpha}}\cD(f)
				+a_{1n} M_n^{-2}r_n^2.
			\end{align*}
			
		Next, we analyze $\cE_{k,2}$. Following a similar decomposition as for $\cE_{k,1}$ and applying \eqref{eq:linearization}, we obtain on the event $\mathcal{A}_n$,
			\begin{equation}
				\begin{split}\label{eq_uperEk2}			
				\cE_{k,2}&=\cR_{\Pbb,k}(\widehat{f}_{k,\delta}^{(n)};\delta)-\cR_{\Pbb,k}(f_{k,n};\delta)
				\\& \leq \cR_{\Pbb,k}(\widehat{f}_{k,\delta}^{(n)};\delta)-\cR_{\Pbb_n,k}(\widehat{f}_{k,\delta}^{(n)};\delta)+\cR_{\Pbb_n,k}(f_{k,n};\delta)-\cR_{\Pbb,k}(f_{k,n};\delta)
				 \\
				&\leq 2\|\ell^{\prime}\|_{\infty} M_n^{-1}r_n(\|\widehat{f}_{k,\delta}^{(n)}-f_0\|_2+\|f_{k,n}-f_0\|_2 +2r_n)
		  \\&\qquad+ 2\delta \sup_{f\in\cF_n} \big|\mathcal V(f)-\mathcal V_n(f)\big|^{1/k^\star}
		+4\delta^2 \sup_{f\in\cF_n}\mathcal {D}(f).
        	\end{split}
			\end{equation}
			The term $\|\widehat{f}_{k,\delta}^{(n)}-f_0\|_2$ has already been examined in the analysis of $\cE_{k,1}$. To bound $\|f_{k,n}-f_0\|_2$, we first assume $\|f_{k,n}-f_0\|_2\geq c_4$.  Since $\cR_{\Pbb,k}(f_{k,n};\delta)\leq \cR_{\Pbb,k}(f_{n};\delta)$ for any $f_n\in\cF_n$,  Assumption 1 in the main text  implies that
			\begin{align*}
		c_3	\|f_{k,n}-f_0\|_2^2&\leq \cR_{\Pbb}(f_{k,n}) -	\cR_{\Pbb}(f_0) \\&\leq 	\cR_{\Pbb,k}(f_{k,n};\delta)- \cR_{\Pbb,k}(f_0;\delta) +\sup_{f\in\cH^{\alpha}}\{ \delta\mathcal V(f)^{1/k^\star}+\delta^2\cD(f)\}
			\\&	\leq  \mathcal R_{\mathbb P,k}(f_n;\delta)-\mathcal R_{\mathbb P,k}(f_{k,\delta}^\star;\delta)+\sup_{f\in\cH^{\alpha}}\{ \delta\mathcal V(f)^{1/k^\star}+\delta^2\cD(f)\}.
			\end{align*}
Choosing $f_n$ as the best approximation of $f_0$ in $\cF_n$ and bounding $\cR_{\Pbb,k}(f_{n};\delta)- \cR_{\Pbb,k}(f_{k,\delta}^\star;\delta)$ similar to $\cE_{k,1}$ yields an upper bound on $c_3 \|f_{k,n}-f_0\|_2^2$ in this non-local regime.  In the complementary case where $\|f_{k,n}-f_0\|_2 < c_4$, its $L^2$ error is directly bounded by $c_4^2$.  Substituting the bounds for $\|\widehat{f}_{k,\delta}^{(n)}-f_0\|_2^2$ and $\|f_{k,n}-f_0\|_2^2$ into \eqref{eq_uperEk2} and applying Young’s inequality show that $\cE_{k,2}$ satisfies the same upper bound as that derived for $\cE_{k,1}$ on the event $\mathcal{A}_n$.  Consequently, on the event $\mathcal{A}_n$, we obtain
				\begin{equation*}
				\begin{split}
					{\cal E}_{\Pbb,k}(\widehat{f}_{k,\delta}^{(n)};\delta)  & =\max\big\{ \cE_{k,1},\cE_{k,2}\big \}
			\\	&\lesssim   c_1+c_4^2+\inf_{f\in\cF_n} \Big\{ c_2\|f-f_{0}\|_{2}^2
			+ \delta C_0    \|f-f_{0}\|_{k^{\star}}
			+ \delta C_0   \big\|\|\nabla(f-f_{0})\|_1 \big \|_{k^{\star}} \Big\} \nonumber\\
			&\qquad  \;  + \delta \sup_{f\in\cF_n} \big|\mathcal V(f)-\mathcal V_n(f)\big|^{\,1/k^\star}+  \delta d\|\ell^{'}\|_{\infty}+ \delta^2 \sup_{f\in\cF_n}\mathcal D(f)
			\\  &\qquad \;  +  \delta^2\sup_{f\in\cH^{\alpha}}\cD(f)
			+a_{1n}M_n^{-2}r_n^2.
				\end{split}
			\end{equation*}
			
			From Lemma \ref{lem_rnn}, we obtain  $r_n \lesssim   M_n n^{-1/2}\sqrt{\mathcal{S}\log(Kn)}$, and there exists   constant $C$ such that $P \{\cA_n \}\geq 1- C\exp\{-\mathcal{S}\log(Kn)\}.$
			Moreover,  ${\cal E}_{\Pbb,k}(\widehat{f}_{k,\delta}^{(n)};\delta)  \leq 2M_n\|\ell^{'}\|_{\infty}$. This is because
					\begin{align*}
			 \cR_{\Pbb, k}(f;\delta)- \cR_{\Pbb, k}(\widetilde{f};\delta)&=\sup_{\Qbb: W_k(\Qbb,\Pbb)\leq \delta}\Ebb_{ \Qbb} [\ell(f(X)-Y) ]-\sup_{\Qbb: W_k(\Qbb,\Pbb)\leq \delta}\Ebb_{ \Qbb} [\ell(\widetilde{f}(X)-Y) ]
			 \\&\leq \sup_{\Qbb: W_k(\Qbb,\Pbb)\leq \delta}\big|\Ebb_{ \Qbb} [\ell(f(X)-Y)-\ell(\widetilde{f}(X)-Y) ]\big|
			 \\&\leq \|\ell^{'}\|_{\infty} \|f-\widetilde{f}\|_{\infty},
					\end{align*}
			which implies that  $\cE_{k,1}, \cE_{k,2}\leq 2M_n\|\ell^{'}\|_{\infty}$.  	Therefore, we show
		\begin{equation*}
		\begin{split}
		\Ebb[{\cal E}_{\Pbb,k}(\widehat{f}_{k,\delta}^{(n)};\delta)  ]
		&= \Ebb[ {\cal E}_{\Pbb,k}(\widehat{f}_{k,\delta}^{(n)};\delta)   1_{\cA_n}] +\Ebb[ {\cal E}_{\Pbb,k}(\widehat{f}_{k,\delta}^{(n)};\delta)   1_{\cA^c_n}] \\
		&\leq \Ebb[ {\cal E}_{\Pbb,k}(\widehat{f}_{k,\delta}^{(n)};\delta)   1_{\cA_n}] +2M_n\|\ell^{'}\|_{\infty}P(\cA_n^c)\\
		& \lesssim  c_1+c_4^2+\inf_{f\in\cF_n} \Big\{ c_2 \|f-f_{0}\|_{\infty}^2
		+ \delta  C_0   \|f-f_{0}\|_{\infty}
		+ \delta    C_0 \big\|\|\nabla(f-f_{0})\|_1 \big \|_{\infty} \Big\} \nonumber\\
		&\qquad  \; +  (c_{2}^{-1}+c_3^{-1}+M_n)\|\ell^{'}\|_{\infty}^2n^{-1}\mathcal{S}\log(Kn)+\delta \Ebb[\sup_{f\in\cF_n} \big|\mathcal V(f)-\mathcal V_n(f)\big|^{\,1/k^\star}]
			\\  &\qquad \; +  \delta  \|\ell^{'}\|_{\infty}
	 + \delta^2\big\{\sup_{f\in\cF_n}\mathcal D(f)+   \sup_{f\in\cH^{\alpha}}\cD(f)\big\}
		+M_n\|\ell^{'}\|_{\infty}\exp\{-\mathcal{S}\log(Kn)\}.
		\end{split}
	\end{equation*}
 Theorem 3.4 in the main text shows that if  $\cN\cN(W,L,\mathcal{S},K)$ satisfies $W\!\ge\! N$,  $\mathcal{S}\!\ge\! N(d+1)$, and $K\gtrsim\!\max\{N^{\frac{d+2m+1-2\alpha}{2d}},\sqrt{\log N}\}$, then
		there exists $g\in\cN\cN(W,L,\mathcal{S},K)$ such that
		\begin{align*}
	&\|g-f_0\|_{\infty}\lesssim
			N^{-\frac{\min\{d+2m+1,2\alpha\}}{2d}}\sqrt{\log N},\\
		&  \| \|\nabla(g-f_0) \|_{1}\|_{\infty}\lesssim
		d	N^{-\frac{\min\{d+2m-1,2(\alpha-1)\}}{2d}}\sqrt{\log N}.
	\end{align*}
	Consequently, we derive
		\begin{equation*}
		\begin{split}
			\Ebb[{\cal E}_{\Pbb,k}(\widehat{f}_{k,\delta}^{(n)};\delta)  ]
				& \lesssim  c_1+c_4^2+  c_2N^{-\frac{\min\{d+2m+1,2\alpha\}}{d}}\log N
		 +  (c_{2}^{-1}+c_3^{-1}+M_n)\|\ell^{'}\|_{\infty}^2n^{-1}\mathcal{S}\log(Kn) \nonumber\\
			&\qquad	 \; + \delta\|\ell^{'}\|_{\infty}+\delta  C_0 N^{-\frac{\min\{d+2m-1,2(\alpha-1)\}}{2d}}\sqrt{\log N}
			\\  &\qquad   \;	+\delta\|\ell^{'}\|_{\infty}  \{n^{-1}\mathcal{S}\}^{1/(2k^\star)} \{L  m^L   \}^{1/k^\star} K
			\\  &\qquad   \;+ \delta^2(\|\ell^{'}\|_{\Lip}K^2+\|\ell^{'}\|_{\infty}Lm^{L}K)
			+M_n\|\ell^{'}\|_{\infty}\exp\{-\mathcal{S}\log(Kn)\}.
		\end{split}
	\end{equation*}
			To achieve the optimal rate, we set
			\[
			\mathcal{S} \asymp W \asymp N,\; K \asymp \max\{N^{\frac{d+2m+1-2\alpha}{2d}},\sqrt{\log N}\},  \; \text{and} \; N\asymp n^{\frac{d}{d+2\beta}},
			\]
		where $\beta = \min\{(d+2m+1)/2, \alpha\}$. Thus, $\mathcal{S} \asymp W \asymp  n^{\frac{d}{d+2\beta}}$ and $K \asymp \max\big\{n^{\frac{d+2m+1-2\alpha}{2(d+2\beta)}},\sqrt{\log n}\big\}$. Substituting these into the bound implies
			 	\begin{align*}
			 \Ebb\big[{\cal E}_{\Pbb,k}(\widehat{f}_{k,\delta}^{(n)};\delta) \big]&\lesssim
			 c_1+c_4^2+ n^{-\frac{2\beta}{d+2\beta}} \{\log n\}^2
			 	+ \delta  \{1
		+ n^{-\frac{    \beta}{k^\star(d+2\beta)}} K \}
			    + \delta^2 K^2.
			 \end{align*}
To ensure that the   terms involving $\delta$ and $\delta^2$ do not dominate the statistical rate and vanish asymptotically, we choose $\delta$ sufficiently small. In particular, setting
\[
\delta  = o\big(n^{- \frac{\max\{4\beta,  2k^{-1}\beta+d+2m+1\} }{2(d+2\beta)} }\big)
\]
 guarantees the optimal convergence rate.

For the case $1 < k < 2$, the main difference is that the $\delta^2$ term in \eqref{eq:linearization} is replaced by $\delta^k$, while the rest of the analysis remains similar. The proof is therefore complete.
			
	\end{proof}

 \phantomsection
 \addcontentsline{toc}{subsection}{Theorem 4.5}
	\begin{statement}{Theorem 4.5}
	 Assume $\|\ell\|_{\Lip}<\infty$. For any  $\Pbb^{\prime}$ satisfying $W_k(\Pbb^{\prime},\Pbb)\leq \delta$, it holds that
\begin{align*}
	{\cal E}_{\Pbb^{\prime}}(f) \lesssim  \cE_{\Pbb,k}(f;\delta)+ \delta \|\ell \|_{\Lip}.
\end{align*}
Moreover, under the conditions of Theorem~4.1 for $k=1$, it holds that
\begin{align*}
	\Ebb\big[	{\cal E}_{\Pbb^{\prime}}(\widehat{f}_{1,\delta}^{(n)})  \big]\lesssim c_1+ c_4^2 + n^{-\frac{ 2\beta}{d+2\beta}}\{\log n\}^2+ \delta K.
\end{align*}	
Alternatively, under the conditions of Theorem~4.3 for $k>1$, it holds that
\begin{align*}
	\Ebb\big[	{\cal E}_{\Pbb^{\prime}}(\widehat{f}_{k,\delta}^{(n)}) \big] \lesssim c_1+c_4^2+ n^{-\frac{2\beta}{d+2\beta}} \{\log n\}^2 + \delta  \{1 + n^{-\frac{    \beta}{k^\star(d+2\beta)}} K \} + \delta^{\min\{k,2\}} K^2.
\end{align*}	
\end{statement}

	\begin{proof}[{\bf  Proof of Theorem 4.5}]
		For any $\Pbb^{\prime}$ such that  $W_1(\Pbb^{\prime},\Pbb)\leq \delta$, the triangle inequality for the Wasserstein distance implies that the ambiguity set centered at $\Pbb$ is contained within a larger set centered at $\Pbb^{\prime}$, that is,
		\begin{align*}
			\big\{\Qbb:W_1(\Qbb, \Pbb)\leq \delta\big\}\subseteq \big\{\Qbb:W_1(\Qbb, \Pbb^{\prime})\leq 2\delta\big\}.
		\end{align*}
		Lemma 2.3 in the main text then implies
		\begin{align*}
			\cR_{\Pbb,1}(f;\delta)-\cR_{\Pbb^{\prime}}(f) \leq \sup_{\Qbb:W_1(\Qbb, \Pbb^{\prime})\leq 2\delta}\cR_{\Qbb}(f)-\cR_{\Pbb^{\prime}}(f)\leq 2\delta \|\ell_f\|_{\Lip}.
		\end{align*}
		  Moreover, noting that $W_1(\Pbb,\Qbb) \leq  W_k(\Pbb,\Qbb) $ for any $1\leq k\leq \infty $,  we have
		  \[
		  0\leq \cR_{\Pbb,k}(f;\delta)-\cR_{\Pbb^{\prime}}(f)\leq 2\delta \|\ell_{f}\|_{\Lip}.
		  \]
		 Taking the infimum over $f\in\cH^{\alpha}$ on both sides yields
		\begin{align*}
			\inf_{f\in\cH^{\alpha}} \cR_{\Pbb,k}(f;\delta)\leq\inf_{f\in\cH^{\alpha}} \big\{ \cR_{\Pbb^{\prime}}(f)+ 2\delta \|\ell_{f}\|_{\Lip}\big\}\leq  \inf_{f\in\cH^{\alpha}} \cR_{\Pbb^{\prime}}(f)+2(d+1) \delta \|\ell \|_{\Lip}.
		\end{align*}
		Therefore, we show
		\begin{equation*}
			\begin{split}
				{\cal E}_{\Pbb^{\prime}}(f)=	\cR_{\Pbb^{\prime}}(f)- \inf_{f\in\cH^{\alpha}} \cR_{\Pbb^{\prime}}(f) &\leq  \cR_{\Pbb,k}(f;\delta)-	\inf_{f\in\cH^{\alpha}} \cR_{\Pbb,k}(f;\delta)+2(d+1) \delta \|\ell \|_{\Lip}.\\
				&\lesssim \cE_{\Pbb,k}(f;\delta)+ \delta  \|\ell \|_{\Lip}.
			\end{split}
		\end{equation*}
	   This result, together with Theorems 4.1 and 4.3 completes the proof.
\end{proof}

 \phantomsection
 \addcontentsline{toc}{subsection}{Theorem 5.1}
    \begin{statement}{Theorem 5.1}
  Suppose Assumption 2 holds.  Assume $\alpha\!>\!1$, $m\!\geq \!2$, $W\!\ge \! N,\mathcal{S}\!\ge\! N(d+1),$ $ K\!\gtrsim\!\max \{N^{\frac{d+2m+1-2\alpha}{2d}},\sqrt{\log N} \},$ and $M_n\asymp\log n$. Let $\beta = \min\{(d+2m+1)/2, \alpha\}$. Then, for any $\delta>0$ and $k\geq 2$, we have
 \begin{equation*}
 	\begin{split}
 		\Ebb\big[ {\cal E}_{\Pbb,k}(\widehat{f}_{k,\delta}^{(n)};\delta) \big] & \lesssim N^{-\frac{2\beta}{d}}\log N + n^{-1}\mathcal{S}\log(Kn)\{ \log n \}^3 + n^{-1} \log n \\
 		&\qquad + \delta \big\{ 1+ N^{-\frac{\beta-1}{d}}\sqrt{\log N} + \{n^{-1}\mathcal{S}\}^{1/(2k^\star)} \{L m^L  \}^{1/k^\star}K \big\}\log n \\
 		&\qquad + \delta^2 \big\{1+K^2+Lm^{L}K\big\}\log n.
 	\end{split}
 \end{equation*}
If further $\mathcal{S} \asymp W\asymp N \asymp n^{\frac{d}{d+2\beta}}$ and $K \asymp \max\big\{n^{\frac{d+2m+1-2\alpha}{2(d+2\beta)}},\sqrt{\log n}\big\}$, then
 \begin{align*}
	\Ebb\big[{\cal E}_{\Pbb,k}(\widehat{f}_{k,\delta}^{(n)};\delta) \big]  \lesssim n^{-\frac{2\beta}{d+2\beta}}\{ \log n \}^4 + \delta \big\{1  + n^{-\frac{ \beta}{k^\star(d+2\beta)}} K \big\}\log n
	+ \delta^2 K^2\log n.
\end{align*}
 In particular, if $ \delta = o\big(n^{- \frac{c}{2(d+2\beta)} }\big)$, with $c=\max\{4\beta, 2k^{-1}\beta+d+2m+1\}$,  then
 \[
 \Ebb\big[{\cal E}_{\Pbb,k}(\widehat{f}_{k,\delta}^{(n)};\delta) \big] \lesssim n^{-\frac{2\beta}{d+2\beta}}\{ \log n \}^4.
\]
 \end{statement}

 \begin{proof}[{\bf Proof of Theorem 5.1}]
 	Let $\beta_n \ge M_n \ge 1$ be a sequence of positive numbers, where $M_n$ denotes the truncation level associated with $\mathcal{F}_n$.  We define the truncated loss function  $\ell_n(u,y):=(u-T_{\beta_n}y)^2$, where $T_{\beta_n}$ is the truncation operator at level $\beta_n$. The corresponding local worst-case   risks are given by
 	 \begin{align*}
 	\cR^{\beta_n}_{\Pbb,k}(f;\delta):= \sup_{\Qbb: W_k(\Qbb,\Pbb)\leq \delta}\Ebb_{ \Qbb}\left[\ell_n(f(X),Y)\right]\;\text{ and }	\;\cR^{\beta_n}_{\Pbb_n,k}(f;\delta):= \sup_{\Qbb: W_k(\Qbb,\Pbb_n)\leq \delta}\Ebb_{ \Qbb}\left[\ell_n(f(X),Y)\right].
 	\end{align*}
   With a slight abuse of notation, let $\ell(u,y) =(u-y)^2$. Then,
 	 \begin{align*}
 	\cR_{\Pbb,k}(f;\delta)= \sup_{\Qbb: W_k(\Qbb,\Pbb)\leq \delta}\Ebb_{ \Qbb}\left[\ell(f(X),Y)\right]\;\text{ and }	\;\cR_{\Pbb_n,k}(f;\delta)= \sup_{\Qbb: W_k(\Qbb,\Pbb_n)\leq \delta}\Ebb_{ \Qbb}\left[\ell(f(X),Y)\right].
 	\end{align*}
 	Under  Assumption 2 in the main text, we can establish upper bounds for $\sup_{f\in\cF_n} |\cR_{\Pbb,k}(f;\delta) - \cR^{\beta_n}_{\Pbb,k}(f;\delta)|$ and $\sup_{f\in\cF_n} |\cR_{\Pbb_n,k}(f;\delta) - \cR^{\beta_n}_{\Pbb_n,k}(f;\delta)|.$  Specifically, for any $\bz^{\prime}=(\bx^{\prime},y^{\prime})\in\cZ$,
 \begin{align*}
  \big|\{f(\bx^{\prime})-y^{\prime}\}^2-\{f(\bx^{\prime})-T_{\beta_n}y^{\prime}\}^2\big|
 \leq 2\big|f(\bx^{\prime})-T_{\beta_n}y^{\prime}\big|\big|y^{\prime}-T_{\beta_n}y^{\prime}\big|+\big|y^{\prime}-T_{\beta_n}y^{\prime}\big|^2.
 	\end{align*}
 	Moreover,
 	\begin{align*}
 	\big|y^{\prime}-T_{\beta_n}y^{\prime}\big|\leq \big|y^{\prime}-y\big|+\big|y-T_{\beta_n}y\big|+\big|T_{\beta_n}y-T_{\beta_n}y^{\prime}\big|\leq 2\big|y-y^{\prime}\big|+\big|y-T_{\beta_n}y\big|,
 	\end{align*}
 which further implies
 	\begin{align*}
 	\big|y^{\prime}-T_{\beta_n}y^{\prime}\big|^2\leq   8\big|y-y^{\prime}\big|^2+2\big|y-T_{\beta_n}y\big|^2.
 	\end{align*}
Let $\Pbb'$ be any distribution within  $\{\Qbb: W_k(\Qbb,\Pbb)\leq \delta\}$. There exist coupled random variables $(X,Y)\sim\Pbb$ and $(X',Y')\sim\Pbb'$ such that   $\Ebb[\max\{\|X^{\prime}-X\|_{\infty}, |Y^{\prime}-Y|\}^k]^{1/k}=W_k(\Pbb^{\prime},\Pbb).$ Setting $\Delta = Y'-Y$, we have  $\Ebb[|\Delta|^2]\leq \Ebb[|\Delta|^k]^{2/k}\leq \delta^2$ and $\Ebb[|\Delta|]\leq \delta$. Combining the above results, for any $f\in\cF_n$,
		\begin{align*}
		&\Ebb_{\Pbb^{\prime}}\big[\big|\ell(f(X^{\prime}),Y^{\prime})- \ell_n(f(X^{\prime}),Y^{\prime})\big|\big]
		\\&\leq \Ebb_{\Pbb^{\prime}}\big[2(M_n+\beta_n)\big|Y^{\prime}-T_{\beta_n}Y^{\prime}\big|+\big|Y^{\prime}-T_{\beta_n}Y^{\prime}\big|^2\big]
		\\&\leq \Ebb\big[2(M_n+\beta_n) \big(2 |\Delta |+\big|Y-T_{\beta_n}Y\big| \big)\big]+\Ebb\big[8 |\Delta |^2+2\big|Y-T_{\beta_n}Y\big|^2 \big]
		\\&\leq 8\beta_n\delta+8\delta^2 	+4\beta_n \Ebb_{\Pbb}[|Y|1\{|Y|> \beta_n\}]+2\Ebb_{\Pbb}[|Y|^21\{|Y|> \beta_n\}].
		\end{align*}
Moreover, we have
 	  	 \begin{align*}
 	  	   &2\Ebb_{\Pbb} \big[ |Y|^21\{|Y|> \beta_n\}\big]+4\beta_n \Ebb_{\Pbb}\big[ |Y|1\{|Y|> \beta_n\}\big]
 	  	 \\&\leq \tfrac{16}{\sigma_{Y}^2} \Ebb_{\Pbb} \big[ \tfrac{\sigma_{Y}^2}{8}|Y|^2 \exp\big\{\tfrac{\sigma_{Y}}{2}(|Y|-\beta_n)\big\}\big]+
 	  	 \tfrac{8\beta_n}{\sigma_{Y}}
 	  	 \Ebb_{\Pbb}\big[ \tfrac{\sigma_{Y}}{2}|Y|\exp\big\{\tfrac{\sigma_{Y}}{2}(|Y|-\beta_n)\big\}\big]
 	  	 \\&\leq \tfrac{16}{\sigma_{Y}^2} \Ebb_{\Pbb} \big[  \exp\{ \sigma_{Y} |Y|\}\exp\{-\tfrac{\sigma_{Y}}{2}\beta_n\}\big]
 	  	 + \tfrac{8\beta_n}{\sigma_{Y}}
 	  	 \Ebb_{\Pbb}\big[  \exp\{ \sigma_{Y} |Y|\}\exp\{-\tfrac{\sigma_{Y}}{2}\beta_n\}\big]
 	  	 \\&\leq \big(\tfrac{16}{\sigma_{Y}^2} +\tfrac{8\beta_n}{\sigma_{Y}}\big)\Ebb_{\Pbb}\big[  \exp\{ \sigma_{Y} |Y|\}\big]\exp\big\{-\tfrac{\sigma_{Y}}{2}\beta_n\big\}
 	  	 \\&\leq \Big\{\big(\tfrac{16}{\sigma_{Y}^2} +\tfrac{8}{\sigma_{Y}}\big)\Ebb_{\Pbb}\big[  \exp\{ \sigma_{Y} |Y|\}\big]  \Big\}\beta_n\exp\big\{-\tfrac{\sigma_{Y}}{2}\beta_n\big\},
 	  	 \end{align*} 	
where the first inequality follows from $\exp\{ \sigma_{Y} (|Y|-\beta_n)/2\}\geq 1$ when $|Y|> \beta_n$, and the second inequality follows from $\exp\{x\}\geq \max\{x^2/2,x\}$ for any $x\geq 0.$  Hence,
	\begin{equation} \label{eq_uppp}
	\begin{split}
	\Ebb_{\Pbb^{\prime}}\big[\big|\ell(f(X^{\prime}),Y^{\prime})- \ell_n(f(X^{\prime}),Y^{\prime})\big|\big]
	\leq 8\beta_n (\delta+\delta^2)	+C\beta_n\exp\big\{-\tfrac{\sigma_{Y}}{2}\beta_n\big\},
	\end{split}
	\end{equation}	  	
 where $C=16(\sigma_{Y}^{-2}+\sigma_{Y}^{-1})\Ebb_{\Pbb}[\exp\{ \sigma_{Y} |Y|\}]$.
  Since \eqref{eq_uppp} holds for any distribution in the set $\{\Qbb: W_k(\Qbb,\Pbb)\leq \delta\}$,  for any uniformly bounded function class $\cG_n$, we obtain
	\begin{equation}\label{eq_221}
	\begin{split}
	\Delta_k(\Pbb;\cG_n)&:=\sup_{f\in\cG_n}	\big| \cR_{\Pbb,k}(f;\delta)- \cR_{\Pbb,k}^{\beta_n}(f;\delta) \big|
	\\&=\sup_{f\in\cG_n}\big|\sup_{\Qbb: W_k(\Qbb,\Pbb)\leq \delta}\Ebb_{ \Qbb}\left[\ell(f(X),Y)\right]-\sup_{\Qbb: W_k(\Qbb,\Pbb)\leq \delta}\Ebb_{ \Qbb}\left[\ell_n(f(X),Y)\right]\big|
	\\&\leq \sup_{f\in\cG_n}\sup_{\Qbb: W_k(\Qbb,\Pbb)\leq \delta}\Ebb_{ \Qbb}\left[\big|\ell(f(X),Y)- \ell_n(f(X),Y)\big|\right]
	\\& \leq 8\beta_n (\delta+\delta^2)	+ C\beta_n\exp\big\{-\tfrac{\sigma_{Y}}{2}\beta_n\big\}.
	\end{split}
	\end{equation}
Similarly,
	\begin{align*}
	\Delta_k(\Pbb_n;\cG_n)& :=\sup_{f\in\cG_n}\big|\cR_{\Pbb_n,k}(f;\delta)- \cR_{\Pbb_n,k}^{\beta_n}(f;\delta)\big|
	\\&=\sup_{f\in\cG_n}\big|	 \sup_{\Qbb: W_k(\Qbb,\Pbb_n)\leq \delta}\Ebb_{ \Qbb}\left[\ell(f(X),Y)\right]-\sup_{\Qbb: W_k(\Qbb,\Pbb_n)\leq \delta}\Ebb_{ \Qbb}\left[\ell_n(f(X),Y)\right]\big|
	\\&\leq\sup_{f\in\cG_n} \sup_{\Qbb: W_k(\Qbb,\Pbb_n)\leq \delta}\Ebb_{ \Qbb}\left[\big|\ell(f(X),Y)- \ell_n(f(X),Y)\big|\right]
	\\&\leq 8\beta_n   (\delta+ \delta^2)	+4\beta_n \Ebb_{\Pbb_n}[|Y|1\{|Y|> \beta_n\}]+2\Ebb_{\Pbb_n}[|Y|^21\{|Y|> \beta_n\}].
	\end{align*}
By taking the expectation on both sides, we show
	\begin{align}\label{eq_222}
	\Ebb[	\Delta_k(\Pbb_n;\cG_n)]  \leq   8\beta_n(\delta+\delta^2)	+C\beta_n\exp\big\{-\tfrac{\sigma_{Y}}{2}\beta_n\big\}.
	\end{align}

Using the bounds in \eqref{eq_221} and \eqref{eq_222}, we now show that the analysis of ${\cal E}_{\Pbb,k}(\widehat{f}_{k,\delta}^{(n)};\delta)$ can be reduced to studying the truncated loss $\ell_n$. Following the notation in the proof of Theorem 4.3, we decompose
  \begin{equation*}
	\begin{split}
		{\cal E}_{\Pbb,k}(\widehat{f}_{k,\delta}^{(n)};\delta)  \leq  \max\big\{  I_{k,1}+I_{k,2}, I_{k,3} \big \},
	\end{split}
\end{equation*}	
where  $I_{k,1} = \cR_{\Pbb,k}(\widehat{f}_{k,\delta}^{(n)};\delta)-\cR_{\Pbb_n,k}(\widehat{f}_{k,\delta}^{(n)};\delta)+\cR_{\Pbb_n,k}(f_n;\delta)-\cR_{\Pbb,k}(f_n;\delta),$   $I_{k,2}= \cR_{\Pbb,k}(f_n;\delta)- \cR_{\Pbb,k}(f_{k,\delta}^{\star};\delta)$, and $I_{k,3}=\cR_{\Pbb,k}(\widehat{f}_{k,\delta}^{(n)};\delta)-\cR_{\Pbb_n,k}(\widehat{f}_{k,\delta}^{(n)};\delta)+\cR_{\Pbb_n,k}(f_{k,n};\delta)-\cR_{\Pbb,k}(f_{k,n};\delta)$. We define their truncated analogues $I_{k,1}^{\beta_n}$, $I_{k,2}^{\beta_n}$, and $I_{k,3}^{\beta_n}$ by replacing $\ell$ with the truncated loss $\ell_n$. Specifically,
  	\begin{align*}
 	I_{k,1}^{\beta_n}&:=  \cR_{\Pbb,k}^{\beta_n}(\widehat{f}_{k,\delta}^{(n)};\delta)-\cR_{\Pbb_n,k}^{\beta_n}(\widehat{f}_{k,\delta}^{(n)};\delta)+\cR_{\Pbb_n,k}^{\beta_n}(f_n;\delta)-\cR_{\Pbb,k}^{\beta_n}(f_n;\delta),\\
 	I_{k,2}^{\beta_n}&:= \cR_{\Pbb,k}^{\beta_n}(f_n;\delta)- \cR_{\Pbb,k}^{\beta_n}(f_{k,\delta}^{\beta_n};\delta),\\
 	I_{k,3}^{\beta_n}&:=\cR_{\Pbb,k}^{\beta_n}(\widehat{f}_{k,\delta}^{(n)};\delta)-\cR_{\Pbb_n,k}^{\beta_n}(\widehat{f}_{k,\delta}^{(n)};\delta)+\cR_{\Pbb_n,k}^{\beta_n}(f_{k,n};\delta)-\cR_{\Pbb,k}^{\beta_n}(f_{k,n};\delta),
 \end{align*}
 where $f_{k,\delta}^{\beta_n}\in\argmin_{f\in \cH^{\alpha}} \cR_{\Pbb,k}^{\beta_n}(f;\delta)$.  For any $f_n\in\cF_n$, we have
	\begin{align*}
	\big|I_{k,1}-I_{k,1}^{\beta_n} \big|&\leq \big|\cR_{\Pbb,k}(\widehat{f}_{k,\delta}^{(n)};\delta)-\cR_{\Pbb,k}^{\beta_n}(\widehat{f}_{k,\delta}^{(n)};\delta)\big|+\big|
	\cR_{\Pbb_n,k}(\widehat{f}_{k,\delta}^{(n)};\delta)-\cR_{\Pbb_n,k}^{\beta_n}(\widehat{f}_{k,\delta}^{(n)};\delta)\big|
	\\&\qquad +\big|
	\cR_{\Pbb_n,k}(f_n;\delta)-\cR_{\Pbb_n,k}^{\beta_n}(f_n;\delta)\big|+\big|\cR_{\Pbb,k}(f_n;\delta) -\cR_{\Pbb,k}^{\beta_n}(f_n;\delta)\big|
	\\&\leq 2\Delta_k(\Pbb;\cF_n)+2\Delta_k(\Pbb_n;\cF_n).
	\end{align*}
Similarly,
	\begin{align*}
	\big|	I_{k,2}-I_{k,2}^{\beta_n} \big|&\leq \big| \cR_{\Pbb,k}(f_n;\delta)-\cR_{\Pbb,k}^{\beta_n}(f_n;\delta)\big|+\big|   \cR_{\Pbb,k}^{\beta_n}(f_{k,\delta}^{\beta_n};\delta)-\cR_{\Pbb,k}(f_{k,\delta}^{\star};\delta)\big|\\
	&\leq \Delta_k(\Pbb;\cF_n)+\Delta_k(\Pbb;\cH^{\alpha}),
	\end{align*}
where the last inequality follows from
	\begin{align*}
	\big|	\cR_{\Pbb,k}^{\beta_n}(f_{k,\delta}^{\beta_n};\delta)-\cR_{\Pbb,k}(f_{k,\delta}^{\star};\delta)\big|&=\big|\inf_{f\in \cH^{\alpha}} \cR_{\Pbb,k}^{\beta_n}(f;\delta)-	\inf_{f\in \cH^{\alpha}} \cR_{\Pbb,k}(f;\delta)\big|
	\\&\leq \sup_{f\in\cH^{\alpha}}| \cR_{\Pbb,k}^{\beta_n}(f;\delta)-\cR_{\Pbb,k}(f;\delta)| = \Delta_k(\Pbb;\cH^{\alpha}).
	\end{align*}
A similar argument yields that for any $f_n\in\cF_n$,
 \begin{align*}
 	\big|	I_{k,3}-I_{k,3}^{\beta_n} \big|&\leq \big| \cR_{\Pbb,k}(\widehat{f}_{k,\delta}^{(n)};\delta)- \cR_{\Pbb,k}^{\beta_n}(\widehat{f}_{k,\delta}^{(n)};\delta)\big|+\big|\cR_{\Pbb_n,k}(\widehat{f}_{k,\delta}^{(n)};\delta)-\cR_{\Pbb_n,k}^{\beta_n}(\widehat{f}_{k,\delta}^{(n)};\delta)\big|
 	\\&\qquad +\big|\cR_{\Pbb_n,k}(f_{k,n};\delta)-\cR_{\Pbb_n,k}^{\beta_n}(f_{k,n};\delta)\big|+\big|\cR_{\Pbb,k}(f_{k,n};\delta)-\cR_{\Pbb,k}^{\beta_n}(f_{k,n};\delta)\big|
 	\\&\leq  2\Delta_k(\Pbb;\cF_n)+2 \Delta_k(\Pbb_n;\cF_n).
 \end{align*}
Moreover, for the truncated risk, $f_{k,n}$ satisfies $ \cR_{\Pbb,k}^{\beta_n}(f_{k,n};\delta)\leq \inf_{f\in \cF_n} \cR_{\Pbb,k}^{\beta_n}(f;\delta)+2 \Delta_k(\Pbb;\cF_n)$. This is because
  \begin{align*}
  &	\big|\cR_{\Pbb,k}^{\beta_n}(f_{k,n};\delta)-\inf_{f\in \cF_n} \cR_{\Pbb,k}^{\beta_n}(f;\delta)\big|
  \\&\leq	\big|\cR_{\Pbb,k}^{\beta_n}(f_{k,n};\delta)-\cR_{\Pbb,k}(f_{k,n};\delta)\big|+
  \big|\inf_{f\in \cF_n} \cR_{\Pbb,k}^{\beta_n}(f;\delta)-
 	\inf_{f\in \cF_n} \cR_{\Pbb,k}(f;\delta)\big|
 \\&	\leq \Delta_k(\Pbb;\cF_n)+\sup_{f\in\cF_n}| \cR_{\Pbb,k}^{\beta_n}(f;\delta)-\cR_{\Pbb,k}(f;\delta)|
 	\leq 2 \Delta_k(\Pbb;\cF_n)	.
 \end{align*}
Combining the above bounds, we obtain
  \begin{equation*}
	\begin{split}
		{\cal E}_{\Pbb,k}(\widehat{f}_{k,\delta}^{(n)};\delta)  & \leq \max\big\{  I_{k,1}^{\beta_n}+I_{k,2}^{\beta_n}+\big| I_{k,1}- I_{k,1}^{\beta_n}+I_{k,2}-I_{k,2}^{\beta_n}\big|, I_{k,3} ^{\beta_n}+\big|I_{k,3}-I_{k,3}^{\beta_n}\big|\big \}
		\\&\leq \max\big\{  I_{k,1}^{\beta_n}+I_{k,2}^{\beta_n}, I_{k,3} ^{\beta_n}\big\}+3\Delta_k(\Pbb;\cF_n)+2\Delta_k(\Pbb_n;\cF_n)+\Delta_k(\Pbb;\cH^{\alpha}).
	\end{split}
\end{equation*}	
Therefore, it suffices to analyze $I_{k,1}^{\beta_n}$, $I_{k,2}^{\beta_n}$, and $I_{k,3}^{\beta_n}$, which are associated with the truncated loss $\ell_n$.  The truncation ensures $\sup_{f\in\cF_n}|f(\bx)-T_{\beta_n}y|\leq 2\beta_n$, implying $\|\ell^{'}_n\|_{\infty}\leq 4\beta_n$ and $\|\ell^{'}_n\|_{\Lip}=2.$ Let  $f_0^{\beta_n}(\bx):=\Ebb[T_{\beta_n}Y|X=\bx]$. Then,
\begin{align*}
\Ebb_{\Pbb}\big[\{f(X)-T_{\beta_n}Y\}^2\big]-\Ebb_{\Pbb}\big[\{f_0^{\beta_n}(X)-T_{\beta_n}Y\}^2\big]&=\|f-f_0^{\beta_n}\|_2^2,
\end{align*}
which verifies Assumption 1 in the main text for the truncated loss with $c_1=c_4=0$ and $c_2=c_3=1$. By arguments similar to those in the proof of Theorem 4.3,   if  $\cN\cN(W,L,\mathcal{S},K)$ satisfies $W\!\ge\! N$,  $\mathcal{S}\!\ge\! N(d+1)$, and $K\gtrsim \max\{N^{\frac{d+2m+1-2\alpha}{2d}},\sqrt{\log N}\}$,  we obtain
\begin{align*}
	&\Ebb\big[ \max\big\{  I_{k,1}^{\beta_n}+I_{k,2}^{\beta_n}, I_{k,3} ^{\beta_n}\big\}\big]
	\\& \lesssim   N^{-\frac{\min\{d+2m+1,2\alpha\}}{d}}\log N
	+ M_n  \beta_n ^2n^{-1}\mathcal{S}\log(Kn)
+ \delta\beta_n \nonumber\\
&\qquad	 \; +\delta  \beta_n N^{-\frac{\min\{d+2m-1,2(\alpha-1)\}}{2d}}\sqrt{\log N} 	+\delta \beta_n \{n^{-1}\mathcal{S}\}^{1/(2k^\star)} \{L  m^L    \}^{1/k^\star} K
	\\  &\qquad   \;+ \delta^2(K^2+Lm^{L}K)\beta_n
	+M_n\beta_n\exp\{-\mathcal{S}\log(Kn)\}+\Ebb[\Delta_k(\Pbb;\cF_n)].
\end{align*}
Setting $  \max\{2\log n/\sigma_{Y},M_n\}\leq \beta_n\leq c\log n$ for some positive constant $c$ and letting $ M_n\asymp \log n$, \eqref{eq_221} and \eqref{eq_222} imply
\[
\Ebb\big[4\Delta_k(\Pbb;\cF_n)+2 \Delta_k(\Pbb_n;\cF_n)+\Delta_k(\Pbb;\cH^{\alpha})\big]\lesssim     (\delta+\delta^2)\log n	+n^{-1} \log n .
\]
Consequently, we  derive
   \begin{equation*}
 	\begin{split}
  	 \Ebb\big[	{\cal E}_{\Pbb,k}(\widehat{f}_{k,\delta}^{(n)};\delta) \big] & \lesssim   N^{-\frac{\min\{d+2m+1,2\alpha\}}{d}}\log N
 	+n^{-1}\mathcal{S}\log(Kn)\{ \log n \}^3 +     n^{-1} \log n
 	\nonumber\\
 	& 	 + \delta\big\{1 + N^{-\frac{\min\{d+2m-1,2(\alpha-1)\}}{2d}}\sqrt{\log N}
 	+  \{n^{-1}\mathcal{S}\}^{1/(2k^\star)} \{L  m^L    \}^{1/k^\star}K\big\} \log n
 	\\  &    + \delta^2\{1+K^2+Lm^{L}K\}\log n .
 	\end{split}
 \end{equation*}	
 	To achieve the optimal rate, let  $\beta = \min\{(d+2m+1)/2, \alpha\}$, and set $\mathcal{S} \asymp W \asymp N\asymp n^{\frac{d}{d+2\beta}}$ and $K \asymp \max\big\{n^{\frac{d+2m+1-2\alpha}{2(d+2\beta)}},\sqrt{\log n}\big\}$. Substituting these into the above  bound implies
 \begin{align*}
 	\Ebb\big[{\cal E}_{\Pbb,k}(\widehat{f}_{k,\delta}^{(n)};\delta) \big]&\lesssim
 	 n^{-\frac{2\beta}{d+2\beta}} \{\log n\}^4
 	 + \delta \{1+  n^{-\frac{    \beta}{k^\star(d+2\beta)}} K \}\log n
 	+ \delta^2 K^2\log n.
 \end{align*}
 To ensure that the terms involving $\delta$ and $\delta^2$ do not dominate the statistical rate, we choose $\delta$ sufficiently small. In particular, setting  $ \delta  = o\big(n^{- \frac{\max\{4\beta,  2k^{-1}\beta+d+2m+1\} }{2(d+2\beta)} }\big)$ ensures the optimal convergence rate. This completes the proof.
 \end{proof}

\phantomsection
\addcontentsline{toc}{subsection}{Theorem 5.2}
\begin{statement}{Theorem  5.2}
	Suppose the conditions of Theorem 5.1 hold and $ \delta = o\big(n^{- \frac{c}{2(d+2\beta)} }\big)$, with $c=\max\{4\beta, 2k^{-1}\beta+d+2m+1\}$. For any    $\Pbb^{\prime}$ satisfying  $W_k(\Pbb^{\prime}, \Pbb)\leq \delta$, and assuming $f_0^{\prime}\in\cH^{\alpha}$, we have
 	\begin{align*}
	\Ebb\big[	\|\widehat{f}_{k,\delta}^{(n)}-f_0^{\prime}\|_{L^2(\Pbb^{\prime}_X)}^2\big] \lesssim n^{-\frac{2\beta}{d+2\beta}}\{ \log n \}^4.
\end{align*}
\end{statement}

\begin{proof}[{\bf Proof of Theorem  5.2}]
From the proof of Theorem 5.1, we have established that
	\begin{align*}
	\big|\inf_{f\in \cH^{\alpha}} \cR_{\Pbb,k}^{\beta_n}(f;\delta)-	\inf_{f\in \cH^{\alpha}} \cR_{\Pbb,k}(f;\delta)\big|
&	\leq \sup_{f\in\cH^{\alpha}}| \cR_{\Pbb,k}^{\beta_n}(f;\delta)-\cR_{\Pbb,k}(f;\delta)| = \Delta_k(\Pbb;\cH^{\alpha})
\\& \leq 8\beta_n (\delta+\delta^2)+C\beta_n\exp\big\{-\tfrac{\sigma_{Y}}{2}\beta_n\big\},
	\end{align*}
where $C=16(\sigma_{Y}^{-2}+\sigma_{Y}^{-1})\Ebb_{\Pbb}[\exp\{ \sigma_{Y} |Y|\}]$. For any distribution $\Pbb^{\prime}$ such that  $W_k(\Pbb^{\prime}, \Pbb)\leq \delta$ and $(X^{\prime}, Y^{\prime}) \sim \Pbb^{\prime}$, define the truncated natural risk $\cR^{\beta_n}_{\Pbb^{\prime}}(f):=\Ebb_{\Pbb^{\prime}}[\ell_n(f(X^{\prime}),Y^{\prime})]$. We can bound the difference between the original and truncated risks under $\Pbb^{\prime}$ as follows.
	 	\begin{align*}
	  \big| \inf_{f\in\cH^{\alpha}}\cR_{\Pbb^{\prime}}(f)-\inf_{f\in\cH^{\alpha}}\cR^{\beta_n}_{\Pbb^{\prime}}(f)\big|
	 	&\leq \sup_{f\in\cH^{\alpha}}|\cR_{\Pbb^{\prime}}(f)-\cR^{\beta_n}_{\Pbb^{\prime}}(f)\big|
	\\&  \leq  8\beta_n (\delta+\delta^2)+C\beta_n\exp\big\{-\tfrac{\sigma_{Y}}{2}\beta_n\big\}.
	 \end{align*}
From the proof of Theorem 4.5, for the truncated loss we also have
\begin{align*}
	\inf_{f\in\cH^{\alpha}}  \cR_{\Pbb,k}^{\beta_n}(f;\delta)  \leq 	  \inf_{f\in\cH^{\alpha}}  \cR^{\beta_n}_{\Pbb^{\prime}}(f)+8 (d+1)\beta_n \delta.
\end{align*}
Combining these inequalities yields
 \begin{align*}
	   -\inf_{f\in\cH^{\alpha}} \cR_{\Pbb^{\prime}}(f)
	   &\leq       -\inf_{f\in\cH^{\alpha}} \cR_{\Pbb^{\prime}}^{\beta_n}(f)+8\beta_n (\delta+\delta^2)+C\beta_n\exp\big\{-\tfrac{\sigma_{Y}}{2}\beta_n\big\}
	\\& \leq -	\inf_{f\in\cH^{\alpha}}  \cR_{\Pbb,k}^{\beta_n}(f;\delta) +8\beta_n (\delta+\delta^2)+C\beta_n\exp\big\{-\tfrac{\sigma_{Y}}{2}\beta_n\big\}+8 (d+1)\beta_n \delta
	   \\&  \leq - \inf_{f\in\cH^{\alpha}}  \cR_{\Pbb,k}(f;\delta)+16\beta_n (\delta+\delta^2)+2C\beta_n\exp\big\{-\tfrac{\sigma_{Y}}{2}\beta_n\big\}+8 (d+1)\beta_n \delta.
	   	  \end{align*}
	Let $f_0^{\prime}(\bx):=\Ebb_{\Pbb^{\prime}}[Y^{\prime}|X^{\prime}=\bx]$. The risk $\cR_{\Pbb^{\prime}}(f)$ decomposes as
	  \begin{align*}
	  \cR_{\Pbb^{\prime}}(f)&= \Ebb_{\Pbb^{\prime}}[|Y^{\prime}-f_0^{\prime}(X^{\prime})+f_0^{\prime}(X^{\prime})-f(X^{\prime})|^2]
	  \\&=\Ebb_{\Pbb^{\prime}}[|Y^{\prime}-f_0^{\prime}(X^{\prime})|^2+2\{Y^{\prime}-f_0^{\prime}(X^{\prime})\}\{f_0^{\prime}(X^{\prime})-f(X^{\prime})\}
	  +|f_0^{\prime}(X^{\prime})-f(X^{\prime})|^2]
	  	  \\&=\Ebb_{\Pbb^{\prime}}[|Y^{\prime}-f_0^{\prime}(X^{\prime})|^2 ]+\|f_0^{\prime}-f\|_{L^2(\Pbb^{\prime}_X)}^2 .
	  \end{align*}
 Therefore,
	 	\begin{align*}
	  \Ebb	\|\widehat{f}_{k,\delta}^{(n)}-f_0^{\prime}\|_{L^2(\Pbb^{\prime}_X)}^2& =\Ebb \big[\cR_{\Pbb^{\prime}}(\widehat{f}_{k,\delta}^{(n)})-\inf_{f\in\cH^{\alpha}} \cR_{\Pbb^{\prime}}(f)\big]
		\\&\leq \Ebb[\cE_{\Pbb,k}(\widehat{f}_{k,\delta}^{(n)};\delta)]+16\beta_n (\delta+\delta^2)	+2 C\beta_n\exp\big\{-\tfrac{\sigma_{Y}}{2}\beta_n\big\}+8 (d+1)\beta_n \delta.
	 \end{align*}
	 Setting $  \max\{2\log n/\sigma_{Y},M_n\}\leq \beta_n\leq c\log n$ for some positive constant $c$ with  $ M_n\asymp \log n$, and combining with  Theorem 5.1, we obtain the  convergence rate. Specifically, let $\mathcal{S} \asymp W \asymp N\asymp n^{\frac{d}{d+2\beta}}$, $K \asymp \max\big\{n^{\frac{d+2m+1-2\alpha}{2(d+2\beta)}},\sqrt{\log n}\big\}$, and $ \delta  = o\big(n^{- \frac{\max\{4\beta,  2k^{-1}\beta+d+2m+1\} }{2(d+2\beta)} }\big)$, where $\beta = \min\{(d+2m+1)/2, \alpha\}$, then
 	\begin{align*}
 	\Ebb	\|\widehat{f}_{k,\delta}^{(n)}-f_0^{\prime}\|_{L^2(\Pbb^{\prime}_X)}^2 \lesssim n^{-\frac{2\beta}{d+2\beta}}\{ \log n \}^4.
 	\end{align*}
This completes the proof.
\end{proof}


\section{Proof of auxiliary lemmas}\label{sec_prof_aux}
\phantomsection
\addcontentsline{toc}{subsection}{Lemma \ref{lem_covermatrix}}
\begin{proof} [{\bf Proof of Lemma \ref{lem_covermatrix}}]
	Let  $\bw_i=(w_{i1}, \dots, w_{in})^{\top}$ denote the $i$-th row of $\bW$. The    infinity matrix norm  can   be denoted by
	\[
	\|\bW\|_{\infty}=\max _{1 \leq i \leq m} \sum_{j=1}^n\left|w_{i j}\right|= \max _{1 \leq i \leq m} \|\bw_i\|_1.
	\]
	According to Lemma \ref{lem_covercall}, there exists a set
	\[
	\big\{\bu_j: j=1,\dots, (2r/u+1)^n\big\}
	\]
	satisfying $\|\bu_j\|_{\infty}\leq 1$ for $ j=1,\dots, (2r/u+1)^n$ such that for any $\bw\in\Rbb^n$ satisfying $\|\bw\|_1\leq r$, one has
	\[
	\big\| \bw-\tilde{\bw}\big\|_1\leq u, \quad  \tilde{\bw}= r \bu_j,
	\]
	for some $j\in\{1,\dots, (2r/u+1)^n\}$. For each $\bw_i$,   let $\widetilde{\bw}_i$ denote the associated approximation constructed above, and let $\widetilde{\bW}$ denote the matrix with $i$-th row being $\widetilde{\bw}_i$.  Then,
	\[
	\big  \|\bW-\widetilde{\bW}\big\|_{\infty}=  \max _{1 \leq i \leq m} \big\|\bw_i-\widetilde{\bw}_i\big\|_1\leq u.
	\]
	Therefore,   the $u$-covering number  associated with the  infinity matrix norm for the class  $\mathcal{M}(m,n,r)$ satisfies
	\[
	\mathcal{N}\left(u, \mathcal{M}(m,n,r),\|\cdot\|_{\infty}\right)\leq (2r/u+1)^{mn},
	\]
	which completes the proof.
\end{proof}

\phantomsection
\addcontentsline{toc}{subsection}{Lemma \ref{lem_normalize}}
\begin{proof}[{\bf Proof of Lemma \ref{lem_normalize}}]
	
	We first define the scaling factors for each layer  $0\le \ell \le L-1$ as $r_{\ell}=\max \{\| (\bA_{\ell},\bb_{\ell})\|_{q},1\}$. We then define the new parameters for the rescaled FNN as  $\tilde{\bA}_\ell = \bA_\ell/r_\ell$, $\tilde{\bb}_\ell = \bb_\ell/(\prod_{s=0}^{\ell} r_s^{m^{\ell-s}})$, $\tilde{\bA}_L=\bA_L (    \prod_{s=0}^{L-1} r_s^{m^{L-s}}  ) $ and set the new parameterization of $g$ as follows:
	\begin{align*}
	&g =\tilde{g}_{L}\circ \tilde{g}_{L-1}\circ\cdots\circ \tilde{g}_{0}, \quad   \text{ where }
	 \\& \tilde{g}_{\ell}(\bx) =\sigma_{m}(\tilde{\bA}_{\ell}\bx+\tilde{\bb}_{\ell})  \text{ for }0\leq \ell\leq L-1, \text{ and }
	\\ &\tilde{g}_{L}(\bx) =\tilde{\bA}_{L}\bx.
	\end{align*}
	For any $0\leq\ell\leq L-1$, due to $r_s \ge 1$ for $ s =0,\dots,  L-1$, we have
	\[
	\| (\tilde{\bA}_\ell, \tilde{\bb}_\ell) \|_{q} = \frac{1}{r_\ell} \left\| \left(\bA_\ell, \frac{\bb_\ell}{\prod_{s=0}^{\ell-1} r_s^{m^{\ell-s}}} \right) \right\|_{q} \le \frac{1}{r_\ell} \| (\bA_\ell, \bb_\ell) \| _{q}\le 1.
	\]

Next, define $G_\ell = g_\ell \circ \dots \circ g_0$ and $\tilde{G}_\ell = \tilde{g}_\ell \circ \dots \circ \tilde{g}_0$ for   $0\leq\ell\leq L-1$.	We show by induction that  $G_{\ell}(\bx) =  ( \prod_{s=0}^{\ell} r_s^{m^{\ell-s+1}}  ) \tilde{G}_{\ell}(\bx)$. First, for $\ell =0$,   since $\sigma_m$ is $m$-homogeneous, i.e. $\sigma_m(c\bx)=c^m\sigma_m(\bx)$, then
	\[
	G_0(\bx)=g_0(\bx) = \sigma_m(\bA_0 \bx+\bb_0) = r_0^m \sigma_m(\tilde{\bA}_0 \bx+\tilde{\bb}_0) = r_0^m \tilde{g}_0(\bx)=r_0^m \tilde{G}_0(\bx).
	\]
	Suppose the claim holds for $\ell-1$, i.e. $G_{\ell-1}(\bx) = \left( \prod_{s=0}^{\ell-1} r_s^{m^{\ell-s}} \right) \widetilde{G}_{\ell-1}(\bx)$, then
	\begin{align*}
		G_{\ell}(\bx) &= \sigma_m(\bA_{\ell} G_{\ell-1}(\bx)+\bb_\ell) = \left( r_{\ell}  \prod_{s=0}^{\ell-1} r_s^{m^{\ell-s}} \right)^m \sigma_m \left(
		\frac{\bA_{\ell}}{r_{\ell}} \frac{G_{\ell-1}(\bx)}{\prod_{s=0}^{\ell-1} r_s^{m^{\ell-s}}} + \frac{\bb_\ell}{r_{\ell}  \prod_{s=0}^{\ell-1} r_s^{m^{\ell-s}}} \right) \\
		&=\left(    \prod_{s=0}^{\ell} r_s^{m^{\ell-s+1}} \right) \sigma_m\left( \tilde{\bA}_\ell \tilde{G}_{\ell-1}(\bx) + \tilde{\bb}_\ell \right) =\left(    \prod_{s=0}^{\ell} r_s^{m^{\ell-s+1}} \right)  \tilde{G}_{\ell}(\bx).
	\end{align*}
	Thus, the claim holds for all $0\leq \ell\leq L-1$. Then, it follows
	\[
	g(\bx) = \bA_L G_{L-1}(\bx) = \bA_L\left(    \prod_{s=0}^{L-1} r_s^{m^{L-s}} \right)  \tilde{G}_{L-1}(\bx) = \tilde{\bA}_L \tilde{G}_{L-1 }(\bx).
	\]
	Therefore,  $g$ admits the constructed reparameterization, which completes the proof.
\end{proof}

\phantomsection
\addcontentsline{toc}{subsection}{Lemma \ref{lem_gradient}}
\begin{proof}[{\bf Proof of Lemma \ref{lem_gradient}}]
	We define the pre-activation vectors   $\bz_0 = \bA_0 \bx+ \bb_0$ and $\bz_\ell = \bA_\ell (g_{\ell-1} \circ \dots \circ g_0)(\bx)+ \bb_\ell$ for $1\leq \ell \leq  L-1$, and let the $i$-th component of $\bz_\ell $ be denoted by $z_{\ell,i}$.
	The gradient of $g$ satisfies
	\[
	\nabla g(\bx) = \bA_0^{\top} \mathbf{D}_0  \bA_1^{\top} \mathbf{D}_1 \cdots  \bA_{L-1}^{\top} \mathbf{D}_{L-1}  \bA_L^{\top}.
	\]
	Here, $\mathbf{D}_\ell = \text{diag}(\sigma_m'(z_{\ell,1}), \dots, \sigma_m'(z_{\ell,d_{\ell+1}}))$ is the diagonal matrix of activation derivatives at layer $\ell$, where $\sigma_m'(z) = m(\max\{z,0\})^{m-1}$.
	By the sub-multiplicative property of the matrix   norm, we have
	\begin{align}\label{eq_profuct}
		\|\nabla g(\bx)\|_1 = \|\nabla g(\bx)^{\top}\|_{\infty} &= \|\bA_L \mathbf{D}_{L-1} \bA_{L-1} \cdots \mathbf{D}_0 \bA_0\|_{\infty}\nonumber
		\\&\leq
		\left\| \bA_L \right\|_{\infty} \prod_{\ell=0}^{L-1} \left( \left\| \mathbf{D}_\ell \right\|_{\infty} \left\| \bA_{\ell} \right\|_{\infty} \right).
	\end{align}
	
	We now bound the norm of each term in the above product.  Under the parameter norm constraint $\|(\bA_\ell, \bb_\ell)\|_{\infty} \le 1$, it follows that
	\[
	\| \bA_{\ell}\|_{\infty}\leq\|(\bA_{\ell}, \bb_{\ell})\|_{\infty} \leq 1, \qquad \text{for }\;   0\leq \ell \leq L-1.
	\]
	Next, we show by induction that $\|\bz_\ell\|_{\infty} \le 1$ for   $0\leq \ell \leq L-1$.
	For $\ell=0$, since $\|\bx\|_{\infty}\leq 1$, we have $\|\bz_0\|_{\infty} =\|\bA_0 \bx+ \bb_0\|_{\infty} \le \|(\bA_0, \bb_0)\|_{\infty} \le 1$. Suppose $\|\bz_{\ell-1}\|_{\infty} \le 1$. This implies $\|\sigma_m(\bz_{\ell-1})\|_{\infty} \le 1$. Then,   $\|\bz_{\ell}\|_{\infty} =\|\bA_{\ell}\sigma_m(\bz_{\ell-1})+\bb_{\ell}\|_{\infty} \le 1$. Thus, the property holds for all $0\leq \ell\leq L-1.$  Therefore,
	\begin{align*}
		\|\mathbf{D}_\ell\|_{\infty}  = \max_{i} |m(\max\{z_{\ell,i},0\})^{m-1}| \le m \cdot 1^{m-1} = m.
	\end{align*}
	Substituting these bounds into \eqref{eq_profuct} yields
	\begin{align*}
		\|\nabla g(\bx)\|_{1} & \leq   m^L \|\bA_L \|_{\infty}.
	\end{align*}

Next, we analyze the Hessian of $g(\bx)$, denoted $\nabla^2 g(\bx)$.
The $i$-th column of the Hessian corresponds to the gradient of the partial derivative $\partial_{x_i} g(\bx)$.
For any $i \in [d]$, applying the product rule yields a sum of $L$ terms:
\[
\nabla \{\partial_{x_i} g(\bx) \} = \sum_{j=0}^{L-1}
\left( \bA_0^\top \mathbf{D}_0 \cdots \mathbf{D}_{j-1} \bA_j^\top \right)
\left( \frac{\partial \mathbf{D}_j}{\partial x_i} \right)
\left( \bA_{j+1}^\top \mathbf{D}_{j+1} \cdots \mathbf{D}_{L-1} \bA_L^\top \right),
\]
where $\frac{\partial \mathbf{D}_j}{\partial x_i}$ is a diagonal matrix whose $k$-th diagonal entry is
\[
\sigma_m''(z_{j,k}) \cdot \frac{\partial z_{j,k}}{\partial x_i} = \sigma_m''(z_{j,k}) \cdot \left( \bA_j \mathbf{D}_{j-1} \cdots \mathbf{D}_0 \bA_0 \right)_{k,i}.
\]
Here,   $\sigma_m''(z) = m(m-1)(\max\{z,0\})^{m-2}$ and $(\bA)_{k,i}$ denotes the $(k,i)$-th entry of $\bA$.
By  the triangle inequality and the sub-multiplicative property of the operator norm, it follows
\begin{align*}
	\big\|	\nabla \{\partial_{x_i} g(\bx) \} \big\|_1\leq  \sum_{j=0}^{L-1}
	\big\|	  \bA_0^\top \mathbf{D}_0 \cdots \mathbf{D}_{j-1} \bA_j^\top   \big\|_1
	\Big\|  \frac{\partial \mathbf{D}_j}{\partial x_i} \Big\|_1
	\big\|	  \bA_{j+1}^\top \mathbf{D}_{j+1} \cdots \mathbf{D}_{L-1} \bA_L^\top \big\|_1.
\end{align*}
Then, we bound each part in the product for a given term $j$.  From the established bounds  $	\| \bA_{\ell}\|_{\infty} \leq 1$ and $	\|\mathbf{D}_\ell\|_{\infty}\leq m$, for $ 0\leq \ell \leq L-1$, we obtain
\begin{align*}
	&\big\|	  \bA_0^\top \mathbf{D}_0 \cdots \mathbf{D}_{j-1} \bA_j^\top   \big\|_1\cdot \big\|	  \bA_{j+1}^\top \mathbf{D}_{j+1} \cdots \mathbf{D}_{L-1} \bA_L^\top \big\|_1
	\\&= \big\|	 \bA_j \mathbf{D}_{j-1}  \cdots  \mathbf{D}_0   \bA_0  \big\|_{\infty}\cdot \big\|	 \bA_L \mathbf{D}_{L-1} \cdots \mathbf{D}_{j+1}  \bA_{j+1}  \big\|_{\infty}
	\\&\leq \Big(\prod_{s=0}^{j-1} \| \bA_s\|_{\infty}\|\mathbf{D}_{s} \|_{\infty}\Big) \| \bA_j \|_{\infty} \Big(\prod_{s=j+1}^{L-1} \| \bA_s\|_{\infty}\|\mathbf{D}_{s} \|_{\infty}\Big) \|\bA_L\|_{\infty}
	\\&\leq  m^{L-1} \|\bA_L\|_{\infty}.
\end{align*}
For the diagonal matrix, since $\|\bz_\ell\|_{\infty} \le 1$ for   $0\leq \ell \leq L-1$, then
\begin{align*}
	\Big\|  \frac{\partial \mathbf{D}_j}{\partial x_i} \Big\|_1&=\max_{k}\big|\sigma_m''(z_{j,k}) \cdot \left( \bA_j \mathbf{D}_{j-1} \cdots \mathbf{D}_0 \bA_0 \right)_{k,i}\big|
	\\&\leq \max_{k} \big| m(m-1)(\max\{z_{j,k},0\})^{m-2}\big|\Big(\prod_{s=0}^{j-1} \| \bA_s\|_{\infty}\|\mathbf{D}_{s} \|_{\infty} \Big)\| \bA_j\|_{\infty}
	\\&\leq  (m-1)m^{j+1}.
\end{align*}

Combining these bounds leads that
\begin{align*}
	\big\|	\nabla \{\partial_{x_i} g(\bx) \} \big\|_1\leq  \sum_{j=0}^{L-1} m^{L-1} \|\bA_L\|_{\infty} (m-1)m^{j+1}\leq Lm^{2L}\|\bA_L\|_{\infty}.
\end{align*}
Therefore, we show
\begin{align*}
	\big\|\nabla^2 g(\bx) \big\|_{\infty}=\max_{i}\big\|	\nabla \{\partial_{x_i} g(\bx) \} \big\|_1\leq   Lm^{2L}\|\bA_L\|_{\infty}.
\end{align*}
This completes the proof.
\end{proof}

\phantomsection
\addcontentsline{toc}{subsection}{Lemma \ref{lem_Sk_property}}
\begin{proof}[{\bf Proof of Lemma \ref{lem_Sk_property}}]
	By the defintion,
	\begin{equation*}
		\sup_{\bx\in\mathbb{B}^d}|	S_m(h)(\bx) |\leq 2^{m/2}\sup_{\bx\in\mathbb{B}^d} |h\big(\bu(\bx)\big)|\leq  2^{m/2}\|h\|_{L^{\infty}(\mathbb{S}^d)}.
	\end{equation*}
	Set $ r=\|\bx\|_2$.  A direct chain-rule calculation gives
	\begin{equation}\label{eq_chainrule}
		\nabla_{\bx} S_m(h)(\bx)
		= m(1+r^2)^{m/2-1} \bx h(\bu(\bx))
		+ (1+r^2)^{m/2} J_{\bu}(\bx)^{\top} \nabla_{\bu} h(\bu(\bx)),
	\end{equation}
	where $J_{\bu}(\bx)\in \mathbb{R}^{(d+1) \times d}$ is the Jacobian matrix of $\bu(\bx)$, which maps vectors from $\Rbb^d$ to $\Rbb^{d+1}$.  Let $u_i(\bx)$ denote the $i$-th component of $\bu(\bx)$.  For any $\ell \in[d]$, we have
	\[
	\frac{\partial u_i}{\partial x_\ell} = \frac{\delta_{i\ell}(1+r^2)-x_i x_\ell}{(1+r^2)^{3/2}}
	\quad \text{ for } i \in[d],
	\qquad
	\frac{\partial u_{d+1}}{\partial x_\ell} = - \frac{x_\ell}{(1+r^2)^{3/2}}.
	\]
    This implies that
	\[
	J_{\bu}(\bx) = \frac{1}{(1+r^2)^{3/2}}
	\begin{pmatrix}
		(1+r^2) \bI_d - \bx \bx^\top \\[4pt]
		- \bx^\top
	\end{pmatrix}.
	\]
	Let $\|J_{\bu}(\bx)\|_{1}$ denote the $1$-norm of $J_{\bu}(\bx)$.
	Then $\|J_{\bu}(\bx)\|_{1} \leq 2+\sqrt{d}$ for any $\bx\in\mathbb{B}^d$.
	
 Next, we analyze $\nabla_{\bu} h(\bu)$ on the sphere $\mathbb{S}^d$.  We will show the following identity holds.
	\begin{equation}\label{eq_identity}
		\nabla_{\bu}h(\bu)
		= -\sum_{1\le i<j\le d+1}\big(u_j\mathbf e_i-u_i\mathbf e_j\big)\,D_{i,j}h(\bu) \quad \text{ for }\bu\in\mathbb{S}^d.
	\end{equation}
	Since $\|u_j\mathbf e_i-u_i\mathbf e_j\|_{\infty}\leq |u_j|+|u_i|\leq \sqrt{2}$, it follows that
	\[
	\|\nabla_{\bu} h(\bu)\|_{\infty}
	\leq  \sum_{1\le i<j\le d+1}\|u_j\mathbf e_i-u_i\mathbf e_j\|_{\infty} |D_{i,j}h(\bu)|
	\leq  \sqrt{2} \sum_{1\le i<j\le d+1}  \|D_{i,j}h\|_{L^{\infty}(\mathbb{S}^d)}.
	\]
	Combining this with \eqref{eq_chainrule}, we obtain
	\begin{align*}
		\|\nabla_{\bx} S_m(h)(\bx)\|_{\infty}
		& \leq  m(1+r^2)^{m/2-1} |h(\bu(\bx))| \|\bx\|_{\infty}
		+ (1+r^2)^{m/2}\, \|J_{\bu}(\bx)^{\top}\nabla_{\bu} h(\bu(\bx))\|_{\infty} \\
		& \leq  m2^{m/2-1} |h(\bu(\bx))|
		+ 2^{m/2} \|J_{\bu}(\bx)\|_{1}\|\nabla_{\bu} h(\bu(\bx))\|_{\infty} \\
		& \leq  m2^{m/2-1}\|h \|_{L^{\infty}(\mathbb{S}^d)}
		+ 2^{(m+1)/2}(2+\sqrt{d}) \sum_{1\le i<j\le d+1} \|D_{i,j}h\|_{L^{\infty}(\mathbb{S}^d)}
		\\&\leq C_{m,d}\Big\{ \|h \|_{L^{\infty}(\mathbb{S}^d)}+\sum_{1\le i<j\le d+1} \|D_{i,j}h\|_{L^{\infty}(\mathbb{S}^d)}\Big\},
	\end{align*}
	where $C_{m,d}=\max\{m2^{m/2-1}, 2^{(m+1)/2}(2+\sqrt{d})\}$ is a constant depending only on $m$ and $d$.
	
	It remains to prove the identity \eqref{eq_identity}.   To this end, we define $H(\bu)=h\big(\tfrac{\bu}{\|\bu\|_2}\big)$ for any non-zero $\bu\in\Rbb^{d+1}$ . We then study   $\nabla_{\bu} H(\bu)$ and restrict the gradient  to $ \mathbb{S}^d$. Fix $k\in[d+1]$. We first show that
	\begin{equation}\label{eq_ukk}
		\partial_{u_k}H(\bu) = -\sum_{j=1}^{d+1} u_j D_{k,j}H(\bu)\quad \text{ for }\bu\in \mathbb{S}^d.
	\end{equation}
	By the definition of $D_{k,j}$,  we have
	\begin{align}\label{eq_111}
		\sum_{j=1}^{d+1} u_j\,D_{k,j}H
		= \sum_{j=1}^{d+1} u_j\big(u_k\partial_{u_j}H - u_j\partial_{u_k}H\big)
		= u_k\sum_{j=1}^{d+1} u_j\partial_{u_j}H- \Big(\sum_{j=1}^{d+1} u_j^2\Big)\partial_{u_k}H.
	\end{align}
	For any $j\in[d+1]$, we have
	\begin{align*}
	\partial_{u_j}H=\sum_{s=1}^{d+1}\partial_{s}h\big(\tfrac{\bu}{\|\bu\|_2}\big) \frac{\delta_{sj}\|\bu\|_2^2-u_su_j}{\|\bu\|_2^3}.
	\end{align*}
	This implies   $\bu^{\top}\nabla_{\bu}H(\bu)=0$ on $\mathbb{S}^d$. Since $\|\bu\|_2^2 = 1$, the right-hand side of \eqref{eq_111} reduces to $- \partial_{u_k}H$, which verifies \eqref{eq_ukk}. Next, define
	\[
	\bV(\bu)  := \sum_{1\le i<j\le d+1}\big(u_j\mathbf e_i-u_i\mathbf e_j\big) D_{i,j}H(\bu).
	\]
	Its $k$-th component $V_k(\bu)$ satisfies
	\[
	\begin{aligned}
		V_k(\bu)
		&= \sum_{1\le i<j\le d+1}\big(u_j\delta_{ik}-u_i\delta_{jk}\big)\,D_{i,j}H(\bu)\\
		&= \sum_{j>k} u_j\,D_{k,j}H(\bu)- \sum_{i<k} u_i\,D_{i,k}H(\bu).
	\end{aligned}
	\]
	Since $D_{k,i}=-D_{i,k}$ for $i<k$ and  $D_{k,k}=0$, the right-hand side simplifies to
	$\sum_{j=1}^{d+1} u_j\,D_{k,j}H$. By \eqref{eq_ukk}, it follows $V_k(\bu) = -\partial_{u_k}H(\bu)$.   Since this holds for every   $k\in[d+1]$, we obtain $\bV(\bu) = -\nabla_{\bu}H(\bu)$. Rearranging the sign and noting that $\nabla_{\bu}h(\bu)=\nabla_{\bu}H(\bu)|_{\mathbb{S}^d} $ yield
	\[
	\nabla_{\bu}h(\bu) = -\sum_{1\le i<j\le d+1}\big(u_j\mathbf e_i-u_i\mathbf e_j\big)\,D_{i,j}h(\bu).
	\]
	This completes the proof.
\end{proof}

\phantomsection
\addcontentsline{toc}{subsection}{Lemma \ref{lem_star}}
 \begin{proof}[{\bf Proof of Lemma \ref{lem_star}}]
     The star-shaped property of the class implies that for any $r_1\leq r_2$,
\[
\Big\{f:\frac{r_2}{r_1}f\in\cF,\|f\|_2\leq r_1\Big\}\subseteq \Big\{f: f\in\cF,\|f\|_2\leq r_1\Big\}.
\]
From this, it follows that
\begin{align*}
\sup_{\substack{g \in \mathcal{F} \\ \|g\|_{2} \leq r_2}}\Big|\frac{1}{n}\sum_{i=1}^n\sigma_ig(\bx_i)\Big| &= \sup_{\substack{g \in \mathcal{F} \\ \frac{r_1}{r_2}\|g\|_{2} \leq r_1}}\Big|\frac{1}{n}\sum_{i=1}^n\sigma_i g(\bx_i)\Big|
\\&=\sup_{\substack{ \frac{r_2}{r_1}f \in \mathcal{F} \\  \|f\|_{2} \leq r_1}}\Big|\frac{1}{n}\sum_{i=1}^n\sigma_i \frac{r_2}{r_1}f(\bx_i)\Big| \leq  \frac{r_2}{r_1}\sup_{\substack{f\in \mathcal{F} \\  \|f\|_{2} \leq r_1}}\Big|\frac{1}{n}\sum_{i=1}^n\sigma_i f(\bx_i)\Big|.
\end{align*}
Thus, the proof is completed.
 \end{proof}

\phantomsection
\addcontentsline{toc}{subsection}{Lemma \ref{lemma_conrate}}
\begin{proof}[{\bf Proof of Lemma \ref{lemma_conrate}}]

We define a series of random variables indexed by  $r$:
\begin{align*}
	 \mathbb{H}_n(r): =\sup_{\substack{f \in \mathcal{F}_n \\ \|f-f_0\|_{2} \leq r}}\big|H_n(f;f_0)\big|
	  =\sup_{\substack{f \in \mathcal{F}_n \\ \|f-f_0\|_{2} \leq r}}\Big|\frac{1}{n}\sum_{i=1}^n\big\{  \ell(Z_i;f)-\ell(Z_i;f_0)\big\}-\Ebb\big[\ell(Z;f)-\ell(Z;f_0)\big]\Big|.
\end{align*}
For any  $r>0$,   define the event
\begin{align*}
    \mathcal{I}_{1,n}(r):=\left\{
   \sup_{f\in\cF_n} \frac{\big|H_n(f;f_0)\big|}{\|f-f_0\|_2+r}> 2 \|\ell\|_{\Lip} M_n^{-1}r
    \right\}.
\end{align*}
To analyze $\mathcal{I}_{1,n}(r)$, we further  define two auxiliary events
\begin{align*}
    \cA_{1n}(r):=\Big\{\mathbb{H}_n(r)\geq 2 \|\ell\|_{\Lip} M_n^{-1}r^2\Big\},
\end{align*}
and
 \begin{align*}
    \cA_{2n}(r):=\Big\{\exists f\in\cF_n \text{ such that } \|f-f_0\|_2\geq r\text{ and } |H_n(f;f_0)|\geq 2 \|\ell\|_{\Lip} M_n^{-1}r\|f-f_0\|_2\Big\}.
\end{align*}
Clearly, $\mathcal{I}_{1,n}(r)\subseteq   \cA_{1n}(r)\cup\cA_{2n}(r)$. To see this, we consider two cases.
If there exists $f\in\cF_n$ such that $\|f-f_0\|_2\leq r$ and $\frac{|H_n(f;f_0)|}{\|f-f_0\|_2+r}> 2\|\ell\|_{\Lip}M_n^{-1}r$, then $\sup_{f\in\cF_n, \|f-f_0\|_2\leq r}|H_n(f;f_0)|\geq 2\|\ell\|_{\Lip}M_n^{-1}r^2$, thus $ \cA_{1n}(r)$ holds. If there exists $f\in\cF_n$ such that $\|f-f_0\|_2> r$ and $\frac{|H_n(f;f_0)|}{\|f-f_0\|_2+r}> 2\|\ell\|_{\Lip}M_n^{-1}r$, then $|H_n(f;f_0)|>2\|\ell\|_{\Lip}M_n^{-1}r\|f-f_0\|_2 $, thus $ \cA_{2n}(r)$ holds. This confirms $\mathcal{I}_{1,n}(r)\subseteq \cA_{1n}(r)\cup\cA_{2n}(r)$.

We first bound the probability of  $\cA_{1n}(r)$. Define $h(\bz;f):=\ell(\bz;f)-\ell(\bz;f_0).$ Then,   $H_n(f;f_0)=\frac{1}{n}\sum_{i=1}^n\{\Ebb[h(Z;f)]-h(Z_i;f)\}$. From the Lipschitz property of  $\ell$ and the uniform boundedness $\sup_{f\in\cF_n}\|f\|_{\infty}\leq M_n$, we have
\[
\sup_{\bz\in\cZ}|h(\bz;f)|=\sup_{\bz\in\cZ}|\ell(\bz;f)-\ell(\bz;f_0)| \leq \|\ell\|_{\Lip}\sup_{\bx\in\cX}|f(\bx)-f_0(\bx)|\leq 2\|\ell\|_{\Lip}  M_n,
\]
which implies $\sup_{f\in\cF_n}\|h(\cdot;f)-\Ebb[h(Z;f)]\|_{\infty}\leq 4\|\ell\|_{\Lip} M_n.$
Furthermore, for any $f\in\cF_n$ satisfying $\|f-f_0\|_2\leq r$, it holds that
\[
 \Ebb\big[\big\{h(Z;f)-\Ebb[h(Z;f)]\big\}^2\big]\leq \Ebb\big[\{h(Z;f)\}^2\big]\leq \|\ell\|_{\Lip}^2 \| f -f_0\|_2^2\leq \|\ell\|_{\Lip}^2r^2.
\]
Additionally, we derive an upper bound for $\Ebb[\mathbb{H}_n(r)]$ as follows.
\begin{align*}
\Ebb[\mathbb{H}_n(r)]\leq    2\Ebb\Big[\sup_{ \substack{f \in \mathcal{F}_n \\ \|f-f_0\|_{2} \leq r} }\Big|\frac{1}{n}\sum_{i=1}^n\sigma_i   h(Z_{i};f)\Big| \Big]&\leq 4\|\ell\|_{\Lip}\cdot \Ebb\Big[\sup_{\substack{f \in \mathcal{F}_n \\ \|f-f_0\|_{2} \leq r}}\Big|\frac{1}{n}\sum_{i=1}^n\sigma_i   (f-f_0)(X_i)\Big| \Big]\\&\leq 4\|\ell\|_{\Lip}\cdot \Ebb\Big[ \overline{R}_n(r;\cF_n^*) \Big],
\end{align*}
where the first inequality follows from the symmetrization argument, and the second inequality follows from the Ledoux–Talagrand contraction
inequality in Lemma \ref{lem_Contraction}.
Since $r_n$  is selected such that
$ \Ebb[ \overline{R}_n(r_n;\cF_n^*)]= r_n^2/(64M_n)$, then
 for any $s, r\geq r_n,$ it follows from Lemmas \ref{lem_star}  and \ref{lemma_talagrand}  that
 $\Ebb[\mathbb{H}_n(r)]\leq \|\ell\|_{\Lip}rr_n/(16M_n) $  and
\begin{align*}
P\Big\{\mathbb{H}_n(r)\geq \frac{\|\ell\|_{\Lip}}{16M_n} rr_n+s^2\Big\}  \le 2\exp\Big\{-\frac{  ns^4}{8e(\|\ell\|_{\Lip}^2r^2+   \|\ell\|_{\Lip}^2rr_n)+16\|\ell\|_{\Lip}M_ns^2}\Big\} .
\end{align*}
By selecting $s^2=  15\|\ell\|_{\Lip} rr_n/(16M_n)$,   there exists a constant $b_1$ such that for any $r\geq r_n$, we have
\begin{align}\label{eq_HH}
P\Big\{\mathbb{H}_n(r)\geq  \|\ell\|_{\Lip} M_n^{-1} rr_n \Big\}  \le 2\exp\Big\{- \frac{nb_1r^2r_n^2}{M_n^2(r^2+rr_n)}\Big\}\le 2\exp\Big\{- \frac{nb_1 r_n^2}{2M_n^2}\Big\}.
\end{align}
Furthermore, by setting $r=2r_n$, we obtain
\begin{align*}
    P\Big\{\cA_{1n}(r_n)\Big\}=P\Big\{  \mathbb{H}_n(r_n)\geq 2 \|\ell\|_{\Lip}M_n^{-1}r_n^2\Big\}\leq 2\exp\Big\{- \frac{2nb_1 r_n^2}{3M_n^2 }\Big\}.
\end{align*}

Next, we derive an upper bound for the probability of $\cA_{2n}(r),$ which is defined by
\begin{align*}
    \cA_{2n}(r):=\Big\{\exists f\in\cF_n \text{ such that } \|f-f_0\|_2\geq r\text{ and } |H_n(f;f_0)|\geq 2\|\ell\|_{\Lip}M_n^{-1}r\|f-f_0\|_2\Big\}.
\end{align*}
For integers $m\geq 1$, we define a series of events
\begin{align*}
    \mathcal{S}_{m}(r):=\big\{f\in\cF_n:2^{m-1}r\leq \|f-f_0\|_2< 2^{m}r\big\}.
\end{align*}
Setting $r=r_n$, we have
$\|f-f_0\|_2\leq 2M_n< 2^{\bar{m}_n}r_n, $ where $\bar{m}_n= \lceil \log(M_nr_n^{-1})/\log 2+1\rceil $. Then, $P\{ \cA_{2n}(r_n)\}=\sum_{m=1}^{\bar{m}_n}P\{\cA_{2n}(r_n)\cap \mathcal{S}_{m}(r_n)\}$. Now, if the event $\cA_{2n}(r_n)\cap \mathcal{S}_{m}(r_n)$ occurs,   there exists   $f\in\cF_n$ with $\|f-f_0\|_2< 2^{m}r_n:=\tilde{r}_m$
 such that
 \[
 |H_n(f;f_0)|\geq 2\|\ell\|_{\Lip}M_n^{-1}r_n\|f-f_0\|_2\geq \|\ell\|_{\Lip}M_n^{-1}  \tilde{r}_mr_n.
 \]
From \eqref{eq_HH}, we obtain
 \[
 P\Big\{\cA_{2n}(r_n)\cap \mathcal{S}_{m}(r_n)\Big\}\leq  P\Big\{\mathbb{H}_n(\tilde{r}_m)\geq\|\ell\|_{\Lip}M_n^{-1} \tilde{r}_mr_n \Big\} \leq  2\exp\Big\{- \frac{nb_1 r_n^2}{2M_n^2}\Big\}.
 \]
Hence, for some constant $b_2>0$, it follows
\begin{align*}
P\Big\{\cA_{2n}(r_n)\Big\}&=\sum_{m=1}^{\bar{m}_n}P\Big\{\cA_{2n}(r_n)\cap \mathcal{S}_{m}(r_n)\Big\}\leq 2\bar{m}_n \exp\Big\{- \frac{nb_1 r_n^2}{2M_n^2}\Big\}\\&\leq \exp\Big\{- \frac{nb_1 r_n^2}{2M_n^2}+b_2\log(\log(M_nr_n^{-1}))\Big\}.
\end{align*}

Combining these results, we show that  the probability of $\mathcal{I}_{1,n}(r_n)$ can be bounded as
\begin{align*}
   P\Big\{\mathcal{I}_{1,n}(r_n)\Big\}&\leq P\Big\{\cA_{1n}(r_n)\Big\}+P\Big\{\cA_{2n}(r_n) \Big\}\\&\leq 2\exp\Big\{- \frac{2nb_1 r_n^2}{3M_n^2 }\Big\}+\exp\Big\{- \frac{nb_1 r_n^2}{2M_n^2}+b_2\log(\log(M_nr_n^{-1}))\Big\}
  \\& \leq 2\exp\Big\{- \frac{nb_1 r_n^2}{2M_n^2}+b_2\log(\log(M_nr_n^{-1}))\Big\}.
\end{align*}
Therefore, with probability at least  $1-2\exp\{- \frac{nb_1 r_n^2}{2M_n^2}+b_2\log(\log(M_nr_n^{-1}))\}$, we derive
 \[
    \big|H_n(f;f_0)\big| \leq  2 \|\ell\|_{\Lip} M_n^{-1}r_n(\|f-f_0\|_2+r_n)~~~~\text{ for any $f\in\cF_n$},
 \]
which completes the proof.
\end{proof}

\phantomsection
\addcontentsline{toc}{subsection}{Lemma \ref{lem_rr}}
\begin{proof}[{\bf Proof of Lemma \ref{lem_rr}}]
We first show the relationship between $\mathbb{E}[\overline{\mathcal{R}}_n(r;\mathcal{F}_n^*)]$ and $\widehat{\mathcal{R}}_n(r;\mathcal{F}_n^*)$. For any $r>0$, define
\begin{align*}
    \mathcal{A}_{1,n}(r):=\left\{
   \sup_{g\in\cF_n^*} \frac{\big|\|g\|_2^2-\|g\|_{2,n}^2\big|}{\|g\|_{2}^2 +r^2}\geq \frac{1}{2}
    \right\}.
\end{align*}
On the complement of $\mathcal{A}_{1,n}(r)$, for every $g\in\mathcal{F}_n^*$, it holds that
\[
\|g\|_{2,n}\le 2\|g\|_2 + r,
\quad \text{ and }\quad
\|g\|_{2}\le 2\|g\|_{2,n} + r.
\]
In this scenario, the following bounds for $\widehat{\cR}_n(r;\cF_n^*)$ and $\overline{\cR}_n(r;\cF_n^*)$ can be derived.
\begin{align*}
\overline{\cR}_n(r;\cF_n^*) =\Ebb_{\bm{\sigma}}\Big[\sup_{\substack{g \in \mathcal{F}_n^* \\ \|g\|_{2} \leq  r}}\Big|\frac{1}{n}\sum_{i=1}^n\sigma_ig(X_i)\Big|\Big]  \leq \Ebb_{\bm{\sigma}}\Big[\sup_{\substack{g \in \mathcal{F}_n^* \\ \|g\|_{2,n} \leq 3r}}\Big|\frac{1}{n}\sum_{i=1}^n\sigma_ig(X_i)\Big|\Big]=\widehat{\cR}_n(3r;\cF_n^*).
 \end{align*}
For any $r,s>0$,   we define the event
 \[
 \mathcal{A}_{2,n}(r,s):=\Big\{
 \big|\overline{\cR}_n(r;\cF_n^*)-\Ebb\big[\overline{\cR}_n(r;\cF_n^*)\big]\big|\geq s
 \Big\}.
 \]
 On the complement of $\mathcal{A}_{1,n}(r) \cup \mathcal{A}_{2,n}(r,s)$,  we have
\begin{equation}\label{eq_completI23}
\begin{split}
 \Ebb\big[\overline{\cR}_n( r;\cF_n^*)]\leq  \widehat{\cR}_n(3r;\cF_n^*)+s.
   \end{split}
 \end{equation}
Subsequent analysis  focuses on bounding the probabilities of $\mathcal{A}_{1,n}(r)$ and $\mathcal{A}_{2,n}(r,s)$.

First, we consider the event $\mathcal{A}_{2,n}(r,s)$. Let
\begin{align*}
m(X_1,\dots,X_n): = \overline{\mathcal{R}}_n(r;\mathcal{F}_n^*)=\mathbb{E}_{\boldsymbol{\sigma}}\Big[\sup_{\substack{g \in \mathcal{F}_n^* \\ \|g\|_{2} \leq  r}}\Big|\frac{1}{n}\sum_{i=1}^n\sigma_ig(X_i)\Big|\Big].
\end{align*}
Since $\mathcal{F}_n^*$ is uniformly bounded by $2M_n$,  $m(\cdot)$ satisfies the bounded differences property. Specifically, if  two samples $\{X_1,\dots, X_n\}$ and $\{\tilde{X}_1,\dots, \tilde{X}_n\}$ differ in at most one element,
\begin{align*}
&\Big|m(X_1, \dots,X_n)-m(\tilde{X}_1,\dots,\tilde {X}_n) \Big|\\&=\bigg|\mathbb{E}_{\boldsymbol{\sigma}}\Big[\sup_{\substack{g \in \mathcal{F}_n^* \\ \|g\|_{2} \leq r}}\Big|\frac{1}{n}\sum_{i=1}^n\sigma_ig(X_i)\Big|\Big]-\mathbb{E}_{\boldsymbol{\sigma}}\Big[\sup_{\substack{g \in \mathcal{F}_n^* \\ \|g\|_{2} \leq r}}\Big|\frac{1}{n}\sum_{i=1}^n\sigma_ig(\tilde{X}_i)\Big|\Big]\bigg|
 \\&\leq \mathbb{E}_{\boldsymbol{\sigma}}\Big[\sup_{\substack{g \in \mathcal{F}_n^* \\ \|g\|_{2} \leq r}}\Big|\frac{1}{n}\sum_{i=1}^n\sigma_i\big\{g(X_i)- g(\tilde{X}_i)\big\}\Big|\Big]\\&\leq \sup_{\substack{g \in \mathcal{F}_n^* \\ \|g\|_{2} \leq  r}}\frac{1}{n}\sum_{i=1}^n\big|g(X_i)- g(\tilde{X}_i)\big|\leq \frac{4M_n}{n}.
\end{align*}
Applying McDiarmid's inequality from Lemma~\ref{lem_MC}, we obtain
\begin{align*}
  P\Big\{\Big|\overline{\mathcal{R}}_n(r;\mathcal{F}_n^*)-\mathbb{E}[\overline{\mathcal{R}}_n(r;\mathcal{F}_n^*)]\Big|\geq s \Big\}\leq 2\exp\Big\{-\frac{ns^2}{8M_n^2}\Big\}.
\end{align*}
Therefore, for any $r,s>0$, there holds
\begin{align}\label{eq_event3}
    P\Big\{ \mathcal{A}_{2,n}(r,s)\Big\}\leq 2\exp\Big\{-\frac{ns^2}{8M_n^2}\Big\}.
\end{align}

Next, to analyze the event $\mathcal{A}_{1,n}(r)=\big\{
   \sup_{g\in\cF_n^*}   |\|g\|_2^2-\|g\|_{2,n}^2  | (\|g\|_{2}^2 +r^2)^{-1} \geq \frac{1}{2}
    \big\},$
we define a series of random variables   $\mathbb{G}_n(r)$  and   events $\mathcal{B}_{n}(r)$,  given as
\begin{align*}  \mathbb{G}_n(r)=\sup_{\substack{g \in \mathcal{F}_n^* \\ \|g\|_{2} \leq r}}\Big|\|g\|_2^2- \|g\|_{2,n}^2\Big|\quad \text{ and }\quad \mathcal{B}_{n}(r)=\Big\{\mathbb{G}_n(r)\geq  r^2/2\Big\}.
\end{align*}
We first establish that $\mathcal{A}_{1,n}(r)\subseteq \mathcal{B}_{n}(r)$. Specifically, consider any $g\in\cF_n^*$ satisfying $\|g\|_2\leq r$ and $|\|g\|_2^2-\|g\|_{2,n}^2|\geq \frac{1}{2}\|g\|_2^2+\frac{1}{2}r^2$. In this case, $\mathcal{B}_{n}(r)$ clearly holds. Now, consider  $g\in\cF_n^*$ such that $\|g\|_2> r$ and $|\|g\|_2^2-\|g\|_{2,n}^2|\geq \frac{1}{2}\|g\|_2^2+\frac{1}{2}r^2$. Let $\tilde{g}=\frac{r}{\|g\|_2}g$. Since  $\cF_n^*$  is star-shaped,   $\tilde{g}\in\cF_n^*$. Furthermore, we have $\|\tilde{g}\|_2=r$ and $|\|\tilde{g}\|_2^2-\|\tilde{g}\|_{2,n}^2|\geq  \frac{1}{2}r^2$, which demonstrates that $\mathcal{B}_{n}(r)$ also holds in this case. Thus, $\mathcal{A}_{1,n}(r)\subseteq \mathcal{B}_{n}(r)$ is verified. Consequently, it suffices to analyze the event  $\mathcal{B}_{n}(r)$.

To bound the probability of $\mathcal{B}_{n}(r)$, note that $\sup_{g\in\cF_n^*}\|g^2-\Ebb[g^2(X)]\|_{\infty} \leq 8M_n^2$ and
\[
 \sup_{\substack{g \in \mathcal{F}_n^* \\ \|g\|_{2} \leq r}}\Ebb\big[\big|g^2(X)-\Ebb[g^2(X)]\big|^2\big] \leq  \sup_{\substack{g \in \mathcal{F}_n^* \\ \|g\|_{2} \leq r}}\Ebb\big[|g(X)|^4\big]\leq 4M_n^2r^2.
\]
Moreover, we have
\begin{align*}
\Ebb[\mathbb{G}_n(r)]\leq    2\Ebb\Big[\sup_{\substack{g \in \mathcal{F}_n^* \\ \|g\|_{2} \leq r}}\Big|\frac{1}{n}\sum_{i=1}^n\sigma_i   g^2(X_i)\Big| \Big]&\leq 16M_n\cdot  \Ebb\Big[\sup_{\substack{g \in \mathcal{F}_n^* \\ \|g\|_{2} \leq r}}\Big|\frac{1}{n}\sum_{i=1}^n\sigma_i   g(X_i)\Big| \Big]\\&= 16M_n\cdot \Ebb\Big[ \overline{R}_n(r;\cF_n^*) \Big],
\end{align*}
where the first inequality follows from the symmetrization argument, and the second inequality follows from the Ledoux–Talagrand contraction
inequality in Lemma \ref{lem_Contraction}.
Since $\Ebb[ \overline{R}_n(r;\cF_n^*) ]/r$ is a non-increasing function for $r\in(0,\infty)$, as established in Lemma \ref{lem_star}, then for any $r\geq r_n$, we have
\[
\Ebb\Big[ \overline{R}_n(r;\cF_n^*) \Big]\leq \frac{r}{r_n}\Ebb\Big[ \overline{R}_n(r_n;\cF_n^*) \Big].
\]
From Talagrand concentration in Lemma  \ref{lemma_talagrand}, for any $r\geq r_n$ and any $x>0$, we have
\begin{align}\label{eq_firsG1}
    P\Big\{\mathbb{G}_n(r) \ge  \mathbb{E} \big[\mathbb{G}_n(r)\big]+x\Big\}\leq 2\exp\bigg\{\frac{-nx^2}{8e(4M_n^2r^2+16^2M_n^3rr_n^{-1}\Ebb[ \overline{R}_n(r_n;\cF_n^*) ])+32M_n^2x}\bigg\}.
\end{align}
As $r_n$ is chosen to satisfy
\begin{equation} \label{eq_deltan}
\Ebb\Big[ \overline{R}_n(r_n;\cF_n^*) \Big]= \frac{r_n^2}{64M_n},
\end{equation}
it follows that $\Ebb[\mathbb{G}_n(r)]\leq rr_n/4$, and the selection $x=r^2/4$ further leads \eqref{eq_firsG1}  to
\begin{align*}
    P\Big\{\mathbb{G}_n(r) \ge   \frac{rr_n}{4}+\frac{r^2}{4}\Big\}\leq 2\exp\Big\{\frac{-nr^4}{a_2M_n^2( r^2+ rr_n) }\Big\},
\end{align*}
where $a_2$ is a positive constant. Therefore, for any $r\geq r_n$, we have
\begin{align}\label{eq_I2}
P\Big\{\mathcal{A}_{1,n}(r)\Big\}\leq    P\Big\{\mathcal{B}_n(r)\Big\}\leq 2\exp\Big\{\frac{-nr^4}{a_2M_n^2( r^2+ rr_n) }\Big\}.
\end{align}

Combining \eqref{eq_completI23}, \eqref{eq_event3}, and \eqref{eq_I2}, for any $r\geq r_n$ and  any $s\geq 0$, with $r_n$ satisfying \eqref{eq_deltan}, then
\begin{align*}
   \Ebb\big[\overline{\cR}_n( r;\cF_n^*)]\leq  \widehat{\cR}_n(3r;\cF_n^*)+s,
 \end{align*}
with probability at least $1- 2\exp\{-\frac{n s^2}{8M_n^2}\}-2\exp\{-\frac{nr^4}{a_2M_n^2( r^2+ rr_n) }\}$. Set $r=r_n$ and $s= r_n^2/(128M_n)$. If $ r_n \geq \hat{r}_n$, then by a similar argument as in the proof of Lemma~\ref{lem_star},
\begin{align*}
    \frac{r_n^2}{64M_n}=\Ebb\big[\overline{\cR}_n( r_n;\cF_n^*)]\leq \widehat{\cR}_n(3r_n;\cF_n^*)+ \frac{r_n^2}{128M_n}&\leq 3r_n \hat{r}_n^{-1}\widehat{\cR}_n(\hat{r}_n;\cF_n^*)+ \frac{r_n^2}{128M_n}
    \\&\leq 3cr_n \hat{r}_n+\frac{r_n^2}{128M_n}.
\end{align*}
This implies that $\hat{r}_n\geq a_3  r_n/(cM_n)$ for some constant $a_3 > 0$, with probability at least $1-2\exp\{-\frac{n r_n^4}{a_4M_n^4}\}-2\exp\{-\frac{nr_n^2}{2a_2M_n^2  }\}$ for some constant $a_4>0$. If additionally $r_n/M_n=o(1)$, then for large  $n$, the bound holds with probability at least $1- \exp\{-\frac{n r_n^4}{a_5M_n^4}\} $ for some constant $a_5>0.$ By appropriately adjusting constants, the proof is completed.
\end{proof}

\phantomsection
\addcontentsline{toc}{subsection}{Lemma \ref{lem_rnn}}
\begin{proof}[{\bf Proof of Lemma \ref{lem_rnn}}]
We first derive a bound on the covering number of $\cF_n^*$.
For any $u>0$, let $\{f_1,\dots,f_q\}$ be an $(u/2)$-cover of $\cF_n$, and let $\{a_1,\dots,a_s\}$ be an $(u/4M_n)$-cover of $[0,1]$. Then, for any $f\in\cF_n$ and $a\in[0,1]$,
\[
\|a(f-f_0)-a_j(f_i-f_0)\|_\infty
\leq |a-a_j|\|f-f_0\|_\infty + a_j\|f-f_i\|_\infty.
\]
Thus $\{a_j(f_i-f_0):j=1,\dots, s, i=1,\dots, q\}$ forms a $u$-cover of $\cF_n^*$. This implies
\begin{align*}
\log \cN(u,\cF_n^*,L^2(\Pbb_n))
&\leq \log \cN(u,\cF_n^*,\|\cdot\|_\infty)  \\
&\leq \log \cN(u/2,\cF_n,\|\cdot\|_\infty) + \log \cN(u/(4M_n),[0,1],|\cdot|\;) \\
&\lesssim \mathcal{S} \log(1+K u^{-1})+\log(1+M_nu^{-1}) \\
&\lesssim  \mathcal{S} \log(1+K u^{-1}),
\end{align*}
where the third inequality follows from Theorem 3.3 in the main text and Lemma \ref{lem_covercall}, and the last inequality follows from the condition $M_n\lesssim K$. By Dudley’s integral covering number bound from Lemma \ref{lem_cover}, we derive
\begin{equation*}
\begin{split}	 \widehat{\cR}_n(r;\cF_n^*)
& \leq \inf_{\eta\geq 0}\left\{	4\eta+12\int_{\eta}^{r}\sqrt{\frac{\log\mathcal{N}(u,\cF_n^*,L^2(\Pbb_n))}{n}}du
	\right\}
    \\&\leq \inf_{\eta\geq 0}\left\{	4\eta+12r \sqrt{\frac{\log\mathcal{N}(\eta, \cF_n^*,L^2(\Pbb_n))}{n}}
 	\right\}
\\&\lesssim \inf_{\eta\geq 0}\left\{	4\eta+12 r\sqrt{\frac{\mathcal{S}\log(1+ K \eta^{-1})}{n}}
	\right\}.
\end{split}
\end{equation*}
 Choosing $\eta = r n^{-1/2}\sqrt{\mathcal{S}\log(K)}$ yields
\[
\widehat{\cR}_n(r;\cF_n^*) \;\lesssim\; r n^{-1/2}\sqrt{\mathcal{S}\big(\log(Kn)+\log(r^{-1})\big)}.
\]
Define  $\hat{r}_n  = n^{-1/2}\sqrt{\mathcal{S}\log(Kn)}.$ Then there exists a constant $b_1>0$ such that
\[
\widehat{\cR}_n(\hat r_n;\cF_n^*) \;\leq\; b_1 \hat r_n^2.
\]
Further, Lemma~\ref{lem_rr} implies that
\[
r_n \leq   b_1 s_1^{-1} M_n n^{-1/2}\sqrt{\mathcal{S}\log(Kn)}.
\]
 Setting $c=b_1s_1^{-1}$ completes the proof.
\end{proof}

\phantomsection
\addcontentsline{toc}{subsection}{Lemma \ref{lem_ab}}
\begin{proof}[{\bf Proof of Lemma \ref{lem_ab}}]
	It suffices to show the subadditivity property
	\[
	(x+y)^r \le x^r + y^r \quad \text{for any } x,y>0, \ r\in(0,1].
	\]
	Without loss of generality, assume $a \ge b$. Then
	\[
	a^{\,r} - b^{\,r} = (b + (a-b))^{\,r} - b^{\,r} \le (a-b)^{\,r}.
	\]
	
	To verify the subadditivity, define $g(t) = (1+t)^r - 1 - t^r$ for $t \ge 0.$
	We have $g(0) = 0$, and $g'(t) = r((1+t)^{\,r-1} - t^{\,r-1}).$
	Since $r-1<0$, the function $s\mapsto s^{\,r-1}$ is strictly decreasing, and $1+t>t$ for $t>0$, we have $g'(t)<0$.
	Thus $g(t) \leq 0$ for all $t\geq 0$, which implies
	\[
	(1+t)^r \le 1 + t^r.
	\]
	Let $t=y/x$  and multiply both sides by $x^r$ to obtain
	\[
	(x+y)^r \le x^r + y^r.
	\]
	This completes the proof.
	
\end{proof}

  \section{Algorithm}\label{sec_algorithm}

  Algorithm~\ref{alg:wdro_concise} implements WDRO using a dual perturbation strategy that systematically generates adversarial examples by perturbing both covariates and responses. At each iteration, it applies the Frank-Wolfe method~\cite{frank1956algorithm} to identify the optimal covariate perturbation that maximizes the loss, followed by a response perturbation based on the gradient information with respect to the updated covariates. This procedure follows the framework proposed by Staib et al.~\cite{staib2017distributionally}, where the Frank-Wolfe method is employed to efficiently approximate the worst-case distribution of covariates for solving the distributionally robust optimization problem.
\begin{algorithm}[H]
	\caption{WDRO with perturbations on both covariates and responses}
	\label{alg:wdro_concise}
	\begin{algorithmic}[1]
		\REQUIRE Dataset $(X,Y)$, model $f_\theta$, uncertainty level $\delta$, loss $L$, step size $\gamma_t = \frac{2}{t+2}$
		
		\FOR{each iteration $t = 0,1,\ldots,T$}
		\STATE Sample  mini-batch $(x, y) \sim \mathbb{P}_{t}$
		\STATE Initialize $x_{\text{adv}} \gets x$
		\FOR{each inner step $k = 0,1,\ldots,K$}
		\STATE Compute loss $\ell = \mathbb{E} [L(f_\theta(x_{\text{adv}}), y)]$
		\STATE Compute gradients $\nabla_x \ell$
		\STATE Find $\delta_x^* = \argmax_{\delta_x: \|\delta_x\|_{t,p} \leq \delta/2} \delta_x^{\top} \nabla_x \ell$
		\STATE Covariate perturbation: $x_{\text{adv}} \gets (1-\gamma_k)x_{\text{adv}} + \gamma_k(x_{\text{adv}} + \delta_x^*)$
		\ENDFOR
		\STATE Compute loss $\ell = \mathbb{E}[L(f_\theta(x_{\text{adv}}), y)]$
		\STATE Compute gradients $\nabla_y \ell$
		\STATE Find $\delta_y^* = \argmax_{\delta_y: \|\delta_y\|_{t,p}  \leq \delta/2} \delta_y^{\top} \nabla_y \ell$
		\STATE Response perturbation: $y_{\text{adv}} \gets y + \delta_y^*$
		\STATE Update $\theta$ using adversarial loss $\mathbb{E}[L(f_\theta(x_{\text{adv}}), y_{\text{adv}})]$
		\ENDFOR
	\end{algorithmic}
\end{algorithm}

\end{appendix}

\bibliography{WDROarXiv2.bib}
\bibliographystyle{imsart-number}
\end{document}